# Large language models as oncology world models

Julien Vibert[1,2,3,4], Jakob Nikolas Kather[1,5,*]

[1] Institute for Medical AI (IMAI), Heidelberg University Hospital and Medical Faculty of Heidelberg University, Heidelberg, Germany.

[2] Medical Oncology, National Center for Tumor Diseases (NCT), University Hospital Heidelberg, Heidelberg, Germany.

[3] Drug Development Department, Gustave Roussy, Villejuif, France.

[4] Université Paris-Saclay, Gustave Roussy, INSERM U1361, IHU PRISM, Villejuif, France.

[5] Else Kroener Fresenius Center for Digital Health (EKFZ), Faculty of Medicine Carl Gustav Carus, TUD Dresden University of Technology, Dresden, Germany.

[*] Correspondence: contact@kather.ai.

## Abstract

Large language models (LLMs) can synthesize cancer patient cases, but their internal representation of oncology, or world model, has not been explored systematically. By only specifying tumor type, we show that LLMs tend to generate similar patients per cancer, with on average 81.0% to 93.2% of a cancer's patients sharing the most frequent value of each baseline feature. By systematically varying patient features in 12 clinical scenarios, we show that they follow biomarker and treatment logic in 29 of 36 model–scenario assessments, against 1 of 12 when previous treatment or frailty matters. Continuing real patient trajectories, an LLM predicted 89.9% 12-month survival, and 79.2% with similar patients as examples, against 72.2% observed. To represent these trajectories, we developed the Clinical State Space, a dynamic representation framework that we also applied to real patients. Despite the limits we show, improved world models may support digital twins, which our framework can evaluate.

Large language models (LLMs) are increasingly used for clinical questions in oncology, such as question answering[1], tumor board assistance[2] and patient-facing chatbots[3]. Pretrained on very large datasets that include trials, studies and public patient cases, they may internalize knowledge of oncology and use it to generate content. Accordingly, researchers have started to use LLMs to generate synthetic oncology data, such as simulated patients for medical education[4] and longitudinal records for benchmarking AI methods[5], reviewed in[6]. Unlike synthetic data generators that must be trained on specific cohorts, such as models trained on large electronic health records[7], LLMs can be used “off-the-shelf” and are thus easily and widely used[4,5]. However, since the pretraining data of the most powerful proprietary models are not known, the reliability, biases and limits of these synthetic data are difficult to judge unless audited carefully, and models can perform well on clinical tasks and still hold an inconsistent model of the world[8]. The concept of world models describes the ability of LLMs to represent their knowledge of the world, notably to generate different scenarios according to different inputs[9,10]. In analogy to how clinicians reason on patient cases, we propose to study the oncology world models of LLMs to assess their capacity and audit their limits.

For this, we develop a systematic, model-agnostic way to generate structured patient trajectories from a varied selection of LLMs, in three experiments (Fig. 1a): unconstrained generation from the cancer type alone, probing with varying inputs of context, history, treatment and patient characteristics, and prediction of real patients, with or without patient examples. Clinical outcomes are usually summarized by endpoints such as survival, which do not describe the disease control, treatment burden and physical condition that patients experience along the way. Because generated trajectories can be sampled densely in time and in clinical characteristics, they also led us to propose the Clinical State Space (CSS), a new dynamic representation of a patient's current state, history and possible future courses, which we apply to the observed trajectories of real patients.

## Different LLMs generate similar cancer patients with model-specific characteristics

To assess how LLMs generate cancer patients when used off-the-shelf, we built a pipeline that generates complete structured trajectories from diagnosis to death, with the same fields for all models: host characteristics (such as age, sex, performance status (PS), weight), disease (such as histology, biomarkers, molecular clones, metastatic sites and burden), treatment and toxicity, with a new entry each time one of these features changes (Methods). We obtained 4,434 trajectories for 148 tumor types from six LLMs: Claude Sonnet 5, Gemini 3.1 Flash-Lite, Kimi K3, GLM-5.2, Qwen 3.6 Flash and Gemma 4 (Methods). Models tended to emit model- and tumor type-specific “archetype” patients: within a cancer type, the most frequent value of each of five baseline features was shared on average by 81.0% to 93.2% of generated patients (five per cancer type and model), depending on the model (Fig. 1b; Extended Data Fig. 1a). One generated trajectory is shown in Fig. 1c (one per model in Supplementary Fig. 1; the full record of another trajectory in Supplementary Fig. 2a,b), and the treatment sequences and adverse events of all trajectories in Supplementary Fig. 3.

Dimension reduction of all timepoints of all trajectories, without cancer type in the input features (PCA, UMAP[11], diffusion map[12] and PHATE[13]; Supplementary Fig. 4a), showed a coherent global structure: in every model, trajectories of different cancers differed more than those of the same cancer (Extended Data Fig. 1b,c). Yet heterogeneity between models was high for the same cancer type, and models disagreed on which cancer types resemble each other (Spearman correlation 0.26 to 0.55 between models; Supplementary Fig. 5). Survival also depended on the model: median survival in lung cancer differed by 11.2 months between models, and Kimi K3 generated the longest survival in every cancer family (Fig. 1d), also among patients with metastatic disease at diagnosis (Supplementary Note 1).

## The Clinical State Space as a dynamic representation of oncology

These complete trajectories led us to define formally the CSS, a common set of coordinates to describe the state of a patient at any time and its evolution. With only two states, alive and dead, such a representation reduces to survival over time, which a Kaplan-Meier curve estimates. In our CSS, both axes describe the patient, and time is shown along the path of each patient (Fig. 2a) or as successive frames for a population. Its features describe the patient (such as PS, weight, toxicity) and the disease (such as tumor sites and burden, molecular clones); they should vary in time and describe the dynamics of disease (unlike age, sex or tumor type, which are rather static context), and ideally be common to all cancer patients so that patients can be compared across cancer types. In this study we chose 20 variables, 10 describing the tumor and 10 the host; the map is formed by their first two principal components (25.6% and 13.9% of the variance), fitted once on the generated atlas (Fig. 2a; Supplementary Fig. 4b,c), or, for interpretability, by the first principal component of each block as separate disease and host axes.

As for state spaces in physics or trajectories in single-cell biology, a patient becomes a curve connecting all states from diagnosis to death, with varying speeds and residence times (Fig. 2a; another course in Supplementary Fig. 2c), and the entire course is important: two patients with the same survival can differ in disease control, treatment burden, physical condition and toxicity (Fig. 2b). Patients in the same region of the map today can also evolve very differently depending on their context, treatment and previous history (Extended Data Fig. 2a), and in the generated atlas, at similar tumor burden and host condition, the share of patients who died within 12 months was higher in all six models after earlier progression (95% intervals excluding zero in four; Extended Data Fig. 2b; an association in generated data, groups not matched on time or cancer type).

A world model can then be asked for a patient's possible future courses under each treatment strategy (Extended Data Fig. 2c; Supplementary Table 1). Averaging the movements of many trajectories over regions of the CSS gives a vector field: in each region, an arrow is the mean displacement over a fixed interval among courses still alive at its end, death being counted as a separate outcome. Conceptually, disease attracts patients towards more tumor burden and poorer condition, and an active treatment pulls them towards a controlled state (Supplementary Fig. 6, top). In generated trajectories, average movements in the same regions differed between phases of the course such as treatment start, response and progression (Supplementary Fig. 6, bottom), and were shorter where cancer families moved in different directions (Supplementary Figs 7 and 8). In addition to what the classical survival curves of four cancer families depict, the CSS shows where the living patients are in disease burden and host condition and how they move over the next 12 months (Fig. 2c), and survival and mean movements can be mapped by starting position (Extended Data Fig. 3). These fields summarize trajectories and are not physical forces, and patients shown at the same place on the map are not necessarily similar (Methods). Nevertheless, the CSS allows us to follow patients through disease burden and host condition over time, and to map treatments as modifiers of their trajectories.

## Evaluating the clinical adequacy of LLM oncology world models

We then assessed whether generated trajectories change appropriately with clinically meaningful inputs. We model generation as an LLM responding to four types of input: context (C; such as age, sex, comorbidities, tumor type, date and site of care); the current state (X) in the CSS; history (H), the course from diagnosis to the start of generation; and action (A), a local or systemic treatment. Common clinical questions then become world model scenarios, such as comparing treatments by varying A for the same patient (Supplementary Table 1). We built 12 scenarios covering variations of C, X, H and A for four models (Supplementary Table 2), and controlled experiments in EGFR-mutant and EGFR-wild-type lung cancer, and checked whether courses changed in the direction expected clinically (Fig. 3; Extended Data Fig. 4).

Across the 12 scenarios, all generated comparisons were correct in 29 of 36 model–scenario results when a biomarker or treatment class decided the answer (e.g., longer progression-free survival (PFS) with EGFR-targeted treatment than with chemotherapy in EGFR-mutant lung cancer), but only in 1 of 12 when previous treatment or physical condition mattered (Extended Data Fig. 4a,b; Supplementary Fig. 9). For instance, in relapsed ovarian cancer, none of the four models fully separated platinum-sensitive from platinum-resistant relapse[14].

In a factorial experiment with Claude Sonnet 5 (120 combinations of EGFR status, current condition, growth before treatment and treatment), the time before first-line progression depended on treatment and EGFR status, whereas the pace of untreated growth before the decision (progression 1 or 4 months after diagnosis, or none) had almost no effect (Extended Data Fig. 4c). Over the first 24 months in EGFR-mutant disease, futures generated with osimertinib spent on average 6.8 more months without progression than with chemotherapy (95% interval 4.9 to 8.9) and 6.8 fewer months after progression (Fig. 3a; Extended Data Fig. 4d), and the CSS shows in which clinical states these months were spent (Fig. 3b-d; Extended Data Fig. 4e). EGFR-mutant futures started on chemotherapy lived longer than wild-type ones (Fig. 3a,d), possibly because Claude switched all 45 of them to osimertinib by month 12 (23 before first-line progression), and none in wild-type disease; without an early switch, PFS with chemotherapy was similar in both (median 9.3 and 9.4 months; Methods). In a three-model EGFR experiment, Claude reversed the osimertinib advantage when EGFR was absent, whereas Kimi and GLM kept osimertinib ahead of chemotherapy in EGFR-wild-type patients (osimertinib minus chemotherapy, median time before first-line progression: Claude -7.1, Kimi +3.8 and GLM +22.3 months; Fig. 3e,f), as if these two models attributed longer disease control to a targeted therapy irrespective of its target. Thus, LLM world models seem to follow general clinical knowledge, with a consistency that depends on the model and scenario, and auditing them systematically with such probing can reveal their inconsistencies.

## Predicting the continuation of trajectories of real patients

We then tested whether LLMs can forecast the future of real patients, using GENIE-BPC[15,16], a densely documented registry from North American academic centers. To limit risks to confidentiality, we used one open-weight model (DeepSeek V4 Flash) through providers configured to deny data collection, with zero data retention requested. We masked each record after the start of a recorded treatment and asked the LLM to generate the rest of the trajectory until death from the patient's context, state, history and treatment. We compared four settings: the record alone; plus the care setting; plus summaries of example patients from the same cohort chosen at random (generic examples); or chosen to be similar (matched examples), i.e., few-shot in-context learning (Methods). We applied the four settings to 450 patients (150 each with lung, breast and colorectal cancer), with ten generated futures per patient and setting (Fig. 4a). With the record alone, the LLM was overoptimistic: generated survival was above observed survival in the first months and below it at longer follow-up in colorectal and lung cancer (Fig. 4b), and early deaths were under-represented (16.2% of patients died within 6 months, against 1.2% to 3.6% of generated futures; Fig. 4c). At 12 months, context and examples modestly reduced the overestimation of survival: predicted survival was 89.9% with the record alone, 89.3% with the care setting, 86.4% with generic examples and 79.2% with matched examples, against 72.2% observed (Extended Data Fig. 5a,b). At 24 months, predictions with generic or matched examples fell below observed survival (43.8% and 42.9%, against 56.8%), whereas the record alone gave 56.3% (Extended Data Fig. 5a). Among the 260 patients who died, the observed survival lay between the 10th and 90th percentiles of the patient's generated survival times in only 32% to 34% of cases (Extended Data Fig. 5c). Matched examples lowered the Brier score compared with the record alone (0.193 against 0.219; mean paired difference -0.026, 95% interval -0.048 to -0.002; unadjusted Wilcoxon P = 0.06), and generic and matched examples had numerically similar scores (0.192 and 0.193; P = 0.79; Fig. 4d). A Cox model cross-validated on the other patients of the cohort, and the survival estimated directly from the matched example patients, had better Brier scores than the LLM (0.160 and 0.162; Fig. 4d), but had numerically similar discrimination (paired differences inconclusive; Extended Data Fig. 5e), although the Cox model is fitted on the same cohort and the LLM sees zero or few examples.

## Assessing the richness of generated patient trajectories beyond point outcomes

While we can assess the generated courses on single endpoints like survival, the LLM generates whole courses, so we also compared them with observed courses. Month by month, generated futures were in response far more often than the registry patients (40% to 48% between months 6 and 12 with the record alone, against 13% to 19%) and almost never in stable disease (0% to 2%, against 10% to 12%; Fig. 5a), partly because generated courses had no event for stable disease (Methods). To visualize these courses, we placed on the CSS the 109 of 150 patients with lung cancer whose liver and thoracic status was known from recent imaging at the start of the forecast treatment (Fig. 5b,c; all groups in Supplementary Fig. 10; colorectal cancer in Extended Data Fig. 6). Generated and recorded courses disagreed for more than half of the evaluable time at 12 months (57% with matched examples, 59% with the record alone; Extended Data Fig. 7a,b). The LLM also over-predicted documented systemic treatment (about 10 months in the first year, against 6.7 months; Fig. 5b; Extended Data Fig. 7c), and started a new regimen before any new progression far more often (averaged over each patient's futures, 70% to 72% of generated courses within 12 months, against 22% of documented courses; 233 patients; Extended Data Fig. 7e). A score combining 17 clinical summaries of the trajectory (Methods) improved with matched examples over the record alone at 12 months (233 patients; Holm-adjusted P = 0.0003) and 24 months (181 patients; P = 0.04), and over generic examples at 24 months only (P = 0.02; Fig. 5d; Extended Data Fig. 7d; scores by block and cancer in Supplementary Fig. 11b,c; individual courses in Supplementary Figs 11a, 12 and 13). Generated courses thus differed largely from observed courses, but they were clinically rich and came closer to them on some measures with similar patients as examples.

## Applying the Clinical State Space to an observed randomized trial

We built the CSS to map the world models of LLMs, but it could also map the observed courses of real patients, the empirical reference that a world model should ideally reproduce, if the data are dense enough. It could then show together the survival, tumor burden, host condition and toxicity that clinicians weigh in oncology studies. Randomized trials usually collect the densest data, so we applied the CSS to PRIME, a trial comparing panitumumab plus FOLFOX with FOLFOX alone in first-line metastatic colorectal cancer[17,18]; we chose it because its shared dataset[19] contains 935 randomized patients and, unlike GENIE-BPC, serial lesion, laboratory and weight measurements, so that its patients could be placed on a CSS map using 14 trial features (Methods). The estimates were consistent in direction with the published findings: restricted mean progression-free time over 24 months[20] was +1.43 months with panitumumab plus FOLFOX in KRAS exon 2 wild-type disease (95% interval 0.09 to 2.74) and -1.14 months in mutant disease (-2.60 to 0.24; Fig. 6a; Extended Data Fig. 8), and in mutant disease, patients assigned panitumumab plus FOLFOX spent 1.82 more of the first 24 months dead (95% interval 0.26 to 3.36; Fig. 6a). Beside Kaplan-Meier curves and forest plots, which show one endpoint at a time, the CSS showed for each arm how patients moved in tumor burden and condition, where they were last measured before progression or death, and how response, severe adverse events and low albumin (a laboratory marker of condition) evolved together with survival (439 patients assessed by day 14; Fig. 6b,c and Supplementary Fig. 14, wild-type disease; mutant disease in Extended Data Fig. 9). Beyond mapping LLM world models, the CSS can thus organize densely documented clinical data and describe treatment effects as differences between whole courses, in complement to classical plots of single endpoints.

### The ladder of precision oncology and the support of the Clinical State Space

The ultimate goal of a world model would be to predict the future of an individual patient and to compare treatments for this patient. On the CSS, this would require estimating the distribution of courses expected with this patient's context, state, history and treatment, not only the mean vector field, and probably learning from many comparable patients, which we call the support. Examples of similar patients, or even of patients from the same cohort, improved some forecasts, but they require comparable patients in the first place, to inform the model and to validate its predictions. Generative models may therefore struggle most where support is low, since their forecasts may rest on the comparable patients represented in their pretraining data. We could not test whether the failures of LLM world models come from low support, but we quantified how support decreases as a clinical question becomes more precise. In the generated atlas, the median survival of comparators matched on more clinical criteria predicted a patient's survival better (median error from 7.2 to 5.1 months, against 8.9 months with random comparators of the same number; Methods), but only while comparators remained (Supplementary Fig. 15). We measured this “ladder of precision oncology” in GENIE-BPC: at each treatment decision, we counted the other patients sharing the same cancer, stage, driver, regimen depth, treatment class, latest disease status, previous regimen (best status and duration), and age band and sex, in this order (Methods). The median number of comparable patients fell from 778 to 1,020 for the cancer type alone to 0 or 1 once the previous regimen was added, and with age and sex no comparable patient remained for 77.8% of decisions in breast cancer, 79.9% in colorectal cancer and 83.3% in lung cancer (Fig. 6d). Missing documentation also reduced support: when two patients with the same undocumented field were no longer counted as comparable, 83% to 96% of decisions had no comparable patient. A precisely described patient with a common cancer can thus have as few comparable records as a patient with a rare cancer. In-context learning or fine-tuning on specific populations may improve LLM world models, but regions of the CSS without relevant support may remain blind spots, where local evidence may be insufficient to learn or directly validate predictions for these patients.

## Discussion

LLMs are already used “off-the-shelf” to generate synthetic patient cases, but their validation has trailed their use: unlike classical generators such as CART[21] or CTGAN[22,23], they are not trained on the target population, and realistic-looking data can hide large failures. Our audit of the oncology world models of LLMs gave three main results. First, unconstrained generation showed model-specific archetypes, a possible form of mode collapse that may reflect pretraining data, so that a generated cohort depends on the model chosen. Second, models largely followed biomarker and treatment logic, but most showed some clinically inappropriate behavior, such as long disease control with a targeted therapy without its target, and almost all failed when history or physical condition should change the answer. Third, on real patients, the tested LLM under-represented early deaths and overestimated survival in the first months, possibly because it learned more from typical “textbook” courses, a hypothesis we could not test. Similar patients as examples improved some forecasting measures without beating fitted baselines, and the added benefit of matched over generic examples depended on the outcome and horizon. Whole courses showed further discrepancies: excess response, scarce stable disease and over-predicted treatment.

Independently of their clinical consistency, dense synthetic trajectories allowed us to propose the CSS, which represents a patient by current state, history and possible future courses. It illustrated how history can matter at the same current state (Extended Data Fig. 2a, with an association in generated data in Extended Data Fig. 2b), where a simulated treatment difference is spent (Fig. 3), discrepancies of generated courses (Fig. 5) and how trial arms differ as whole courses (Fig. 6). Time without and after progression can be derived from survival curves alone, as in partitioned and multi-state survival analyses[24,25]; the CSS adds the clinical states in which this time is spent (Extended Data Fig. 4e), within a common framework of host and disease features, since clinicians aim not only for longer survival but also for more time in favorable states. Coordinates were adapted to each dataset, and these summaries do not depend on the two-dimensional map. Predicting the direction of the difference between treatments, which LLMs may reproduce even when they misestimate its magnitude[26], could be a first target for future world models, although a correct ordering of treatments is not a calibrated clinical benefit. We envision that the CSS could remap completed trials and real-world data to describe dynamic phenotypes of real patients and to look for treatment effects that classical plots of single endpoints may mask, as hypotheses to be confirmed. Denser collection of clinical data[27] could also refine how patients are grouped in studies and trials, and the framework could extend to other specialties in which patient features change with disease and treatment.

Taken together, the oncology world models of the LLMs we tested are already rich and largely follow clinical logic, but they are model-specific, inconsistent in some situations and not adapted as such to a given population. The failure of the tested LLM to beat fitted baselines recalls the debate on foundation models against linear baselines in single-cell biology[28]. Outcomes of similar patients also vary for irreducible reasons. This is the difficulty of rare cancers, where few or no comparable patients exist, either as real patients from whom outcomes can be learned or as examples in the pretraining data of a generative model. While features irrelevant to an outcome reduce support artificially, the clinical challenge is to find the features and granularity at which groups genuinely differ. Since LLMs learn associations, counterfactual questions may need causal approaches and diverse longitudinal data. Generating more trajectories does not supply evidence in unsupported regions, where predictions would need to transport known effects to related cancers and treatments[29] and be validated externally.

Our study has several limitations. First, our pipeline may have constrained the generation; for instance, requiring generation until death may have favored some outcomes. Second, we tested few scenarios and models, and for cost and privacy reasons only one model (DeepSeek, reasoning turned off) on GENIE-BPC patients, whose records come from academic centers, depend on the timing of scans and lack dynamic host features such as PS and weight; the models may also have seen related data during pretraining. Third, representing patients as trajectories requires a density of features and timepoints rarely available in routine care, although trials and densely documented registries can already be represented. Fourth, the state space features, the grouping of patients and most scoring rules were chosen by the authors, some after exploratory analyses. Finally, we did not show improved clinical decisions or patient outcomes.

LLMs can already generate rich clinical courses, but the model and conditioning strategies we tested did not forecast them reliably for individual patients, and our results do not support their use as predictive models for now. Training domain knowledge into generalist models, or building lightweight domain-specific models, may yield better oncology world models, which should be validated prospectively and could one day support digital twins comparing treatment scenarios for an individual patient[30]. We provide a framework to evaluate these models, as well as a new dynamic interpretable representation for generated and observed patient trajectories within the Clinical State Space framework.

# Methods

## Data sources and ethics

We analyzed existing, de-identified records from the AACR Project GENIE Biopharma Collaborative and the PRIME trial, accessed through Synapse and Project Data Sphere, respectively. No patients were recruited or contacted, and no clinical interventions were performed. The characteristics of the patients analyzed are given in Supplementary Table 3.

## Generation of patient trajectories

**Generative model.** We evaluated LLMs as oncology world models through the trajectories they generate[10]. A trajectory $\omega$ is drawn from a distribution that depends on the inputs given to the model:

$$\omega \sim Q_\theta(\cdot \mid C, X, H, A, E)$$

where *C* is the context (such as age, sex, tumor type, date and site of care), *X* the current state of the patient, *H* the history from diagnosis to the start of generation, *A* the treatment and *E* optional reference evidence, here summaries of example patients. $\theta$ denotes the model with its prompt, output schema, sampling settings and acceptance rules (Supplementary Table 4). We used the models as provided, without training or fine-tuning.

**Generated record.** Each trajectory is a structured JSON record (schema and prompts, Supplementary Table 5). It contains a baseline description of the patient and a chronological list of dated events. The prompt asked for a new event whenever the host characteristics, the disease, the treatment or toxicity changed, until death. When generation started from a supplied history, the model returned only the events after the start of generation.

**Reconstruction of the patient state.** We reconstructed the state of the patient at any time by starting from the baseline and applying the dated events in chronological order. A disease site or a molecular clone remained present until an event changed it, even when later events did not mention it. Death ended the trajectory. Supplementary Fig. 2 shows one generated trajectory as its record, its summaries and its path on the state map.

**Acceptance of generated records.** We accepted a record when it contained every required field, used only allowed values, had a consistent chronology and could be reconstructed. When generation started from a supplied history, the events also had to be dated after the start and could not repeat the supplied history. A rejected record was requested again, up to three attempts per requested trajectory in the atlas and the registry experiment, and four in the controlled experiments. We kept every raw response and rejected attempt, and never corrected the clinical content of a response. Accepted counts per run are in Supplementary Table 4.

**Models and settings.** The atlas models were Gemma 4 (google/gemma-4-31b-it-20260402), Claude Sonnet 5 (anthropic/claude-sonnet-5-20260630), Gemini 3.1 Flash-Lite (google/gemini-3.1-flash-lite-20260507), Kimi K3 (moonshotai/kimi-k3), GLM-5.2 (GLM-5.2-FP8) and Qwen 3.6 Flash (qwen/qwen3.6-flash), with the identifiers returned with each response. We called them between 3 June and 27 July 2026 through OpenRouter, except GLM-5.2, which ran on an institutional OpenAI-compatible endpoint. All calls requested temperature 1.0, reasoning off, a JSON-object response format, at most 16,384 output tokens and no sampling seed. Top-p was left at the provider default, except for GLM-5.2 (1.0). The controlled experiments used the same prompt template and settings, with the models given in each experiment below. The registry model is described with the registry experiment.

## Cancer atlas

**Cancer types and requests.** The 148 cancer types are the OncoTree codes that the clinician first author classified as primary entities, by default leaf codes, among the 897 codes of the release of 3 October 2025. Each request fixed only the OncoTree code, its name and a diagnosis date (15 January 2024), and the model generated every other baseline field and the course. We requested trajectories until five were accepted for each cancer type and model, with at most 24 requests per cancer type and model.

**Baseline concentration.** For Fig. 1b, we computed, for each cancer type and model, the fraction of generated patients with the most frequent value of each of five baseline features (driver, sex, stage, smoking and frailty). We averaged these fractions over the five features and then over cancer types, with 95% bootstrap intervals over cancer types. With five patients, a feature split evenly between two values gives a mean modal fraction of 68.8%.

**Survival.** Every accepted atlas trajectory ends in death, so generated survival is the time from diagnosis to death, without censoring. For Fig. 1d, we grouped 56 of the 148 cancer types into eight cancer families and compared generated survival between the six models within each family with a Kruskal-Wallis test, Holm-adjusted across the eight families. Survival by family included all stages. The same comparison restricted to patients with metastatic disease at diagnosis is given in Supplementary Note 1. Supplementary Fig. 3 summarizes the treatment sequences and adverse events, with the categories assigned by each model from the 13 categories of the output schema.

**Differences between cancers and between models.** To compare trajectories of different cancers (Extended Data Fig. 1b,c), we summarized the first 24 months of each trajectory with the five measurement families described below (clock, occupancy, richness, temporal grammar and treatment). The scaling of these summaries was fitted on two trajectories per cancer type and model, selected by the smallest SHA-256 hash of their identifier, and the comparisons used the three other trajectories. For each model, we divided the mean Euclidean distance between trajectories of different cancers by the mean distance between trajectories of the same cancer. A ratio of one means that the cancers are not distinguished. The analysis used the 146 cancer types with five accepted trajectories in every model, with 95% intervals from 399 bootstrap resamples of cancer types (seed 20260820). To compare which cancers each model placed close to each other (Supplementary Fig. 5c), we computed the Spearman correlation between models of their distances between cancer centroids.

**Selected patients.** The patient of Fig. 1c was chosen in two steps. We first kept the generated EGFR-mutant lung adenocarcinomas with at most 30 events, a progression, a second treatment line and death. Among them, we took the record with the smallest SHA-256 hash of its identifier, which gives a reproducible choice. Fig. 2a shows the same patient on the state map. Supplementary Fig. 1 shows, for each model, the lung adenocarcinoma record with the smallest hash among its five. Supplementary Fig. 2 shows the longest-followed of the five invasive ductal carcinomas generated by Kimi K3.

## Clinical State Space

**State variables.** We described the state of a patient at any time by 20 variables, each on a fixed range from 0 to 1: 10 for the tumor and 10 for the host. The tumor variables are the burden in nine anatomical classes (hepatic, thoracic, nodal, osseous, marrow, central nervous system, peritoneal, soft tissue or cutaneous, and other) and the overall anatomical extent. Burden in each class is the highest burden band of its sites, from 0 (absent) to 5 (diffuse), divided by 5. Extent is the number of involved classes, capped at 5 and divided by 5. The host variables are ECOG performance status, symptoms, renal, hepatic, cardiac, marrow and pulmonary function, relative weight loss, and the maximum grade and number of active adverse events. ECOG, symptoms and organ function are divided by their maximum grade. Weight loss from the first recorded weight is clipped to -20% to +30% and rescaled to 0 to 1. The maximum grade and the number of active adverse events, capped at 5, are divided by 5.

**State map.** To display the 20 variables in two dimensions, we used PCA, a linear projection better suited than nonlinear embeddings to represent movements and trajectories. We centred the variables without scaling and fitted the principal components on all living atlas states, giving equal weight to each model, to each trajectory within a model and to each state within a trajectory. The state map is formed by the first two components (25.6% and 13.9% of the variance; Fig. 2a; Supplementary Fig. 4b,c). PC1 increases with overall encoded burden and PC2 with host burden. All atlas and controlled-experiment displays labelled as the state map use this fixed projection. The projection contracts the space, so patients shown at the same place are not necessarily similar. Comparisons of patients used the variables specified for each analysis, not positions on the map. UMAP[11], PHATE[13] and a diffusion map[12] were computed on the same states for comparison (Supplementary Fig. 4a).

**Disease and host axes.** For interpretability, several displays use two separate axes: the first principal component of the 10 tumor variables (the disease axis) and of the 10 host variables (the host axis), fitted on all atlas positions (30.6% and 43.6% of the variance of each block). Higher values mean more disease or a more impaired host. The axes of GENIE-BPC and PRIME were fitted on their own data and are described with these datasets.

**Encoding check.** To check the encoding and the reconstruction, we compared the mean change of the tumor, host and toxicity coordinates at each generated event type with the direction expected (Supplementary Fig. 4d). Tumor burden was expected to decrease with response and local intervention and to increase with progression and clonal evolution. The host coordinate was expected to increase with a patient-state update and toxicity with a toxicity event. The start and the switch of a systemic treatment, which do not change the state by themselves, served as negative controls.

**Measurement families.** To summarize a trajectory over a given period, we used five families of measures: elapsed time and endpoints (clock), time in clinical states (occupancy), event counts and diversity (richness), the sequence of events (temporal grammar), and treatment exposure and changes. In GENIE-BPC, we used only the measures that the registry records. Supplementary Table 1 restates common clinical questions with these measures.

**Population displays.** Each patient has one starting position: diagnosis in the atlas, the treatment decision in the controlled experiments, the start of the forecast treatment (the landmark) in GENIE-BPC and randomization in PRIME. Maps of the starting position show survival at a fixed time according to this position, as a Kaplan-Meier estimate weighted by a Gaussian kernel around each point (standard deviation, 14% of each axis range, 18% in GENIE-BPC) on a 40 × 40 grid. We left the map blank where the kernel weights of patients with known status summed to less than 12 (10 in GENIE-BPC) or to less than 80% of the local weights. In the controlled experiments, which start from five fixed states, survival is the Kaplan-Meier estimate of the futures from each starting state, without smoothing. Arrows show the mean move from the start to a given month among the patients alive and located at that month. In Fig. 2c, arrows show instead the mean move over the next 12 months of patients still alive 12 months later, grouped by their position in the frame on a 6 × 6 grid, in regions with at least 8 such patients. Follow-up frames show the same patients at successive months. In the frames, shading is relative to the peak density at the start, so it fades as patients die or are no longer assessed, and contours enclose 50% and 90% of the positions located in that frame. Events are bright when new since the previous frame and pale when earlier. In the clinical course tables, survival is the Kaplan-Meier estimate from the start, and the other rows are percentages among living patients with a known value, so denominators differ by row and month. Survival curves are Kaplan-Meier estimates, with 95% intervals from a bootstrap of patients, or of starting records in the controlled experiments.

**Vector fields.** A vector field summarizes the mean displacement of trajectories over a given interval among those still followed at its end. For Supplementary Fig. 7, every atlas trajectory alive at 3, 6, 12, 18 and 24 months contributed one start visit at each of these times. We cut the state map into square cells of 0.12 map units. Each arrow runs from the mean start position of a cell to the mean position 6 months later of the start visits still alive then. We drew a cell only when it had at least 30 distinct trajectories starting in it and at least 30 alive at the end of the interval (69 cells). Arrow color is the share of the cell's start visits that died within 6 months, so death is counted as an outcome and not as a position. Supplementary Fig. 8 decomposes one arrow into the clinical changes behind it, using the only lung adenocarcinoma course of at most 10 events in which response, toxicity onset and progression each changed the state.

**Fields by phase of the course.** For Supplementary Fig. 6, we assigned monthly start visits of every atlas trajectory to six overlapping phases, defined from the trajectory's own events: before the first systemic treatment; the first 3 months after the first systemic start, before any progression; after a first response, before the next progression; 3 months or more after a first response, with no progression recorded since; the first 6 months after the first progression; and the first 3 months after the next treatment start following a progression. On the grid of Supplementary Fig. 7, each arrow is the mean 3-month displacement of the start visits followed for 3 months, drawn when at least 30 distinct trajectories started in the cell and at least 30 contributed an endpoint. Intervals come from 2,000 bootstrap resamples of whole trajectories (seed 20261028). Regions where generated patients accumulate are the 50% highest-density regions of time-weighted positions (120 × 120 grid, Gaussian smoothing of 1.5 bins). Deaths within 3 months were counted as distinct trajectories.

**History at similar states.** To compare patients at similar states with or without a previous progression (Extended Data Fig. 2b), we used a PCA of 16 tumor and host variables, scaled by their 10th to 90th percentile range and fitted without weighting on all living atlas visits. Its first component was oriented toward a fixed severity index and standardized by its median and interquartile range. Within each model and cancer type, we kept the living visits at least 6 months after diagnosis that lay between the 35th and 65th percentiles of both components and between the 10th and 90th percentiles of time since diagnosis, and used the earliest such visit of each trajectory. A previous progression was a progression more than 3 months before the visit. We pooled cancers within each model and compared the share of patients who died within 12 months in the two groups. The groups were not matched on time since diagnosis or on cancer type.

**Constructed illustrations.** Fig. 2b, Extended Data Fig. 2a,c and Supplementary Fig. 8c show constructed patients, computed from drawn intervals. In these patients, disease state follows thresholds adapted from RECIST 1.1[31]: response starts when the burden falls at least 30% below its value at the start of the current treatment and lasts until progression, when the burden rises at least 20% above its lowest value. The state is unknown before the first assessment.

## Controlled clinical experiments

**Scenario benchmark.** We built 12 constructed clinical scenarios, each with one or two comparisons between treatments whose expected direction is given in Supplementary Table 2. Gemini 3.1 Flash-Lite, Qwen 3.6 Flash, Gemma 4 and Claude Sonnet 5 generated three futures per arm and model on 13 July 2026. Treatments were compared on the median generated progression-free time of the assigned line (line 1, or line 2 when the scenario included previous treatment). A comparison was correct when the expected ordering held, and the platinum-sensitivity comparison required a difference of more than 1.5 months. Comparisons with supportive care, which has no assigned line, were not scored. The history-dependent comparisons were platinum control after platinum-sensitive versus platinum-resistant relapse of ovarian cancer, and switching treatment versus continuing osimertinib after progression. The comparison depending on the patient's condition was the ranking of intensive chemotherapy in fit versus frail patients with pancreatic cancer. The score of a scenario is the share of its comparisons that were correct (Extended Data Fig. 4a,b). In a sensitivity analysis, we varied the platinum threshold and used overall survival instead of progression-free survival (Supplementary Note 1).

**Factorial experiment.** Claude Sonnet 5 generated three futures for each of 120 combinations of four inputs (360 futures, 23 July 2026): EGFR status (mutant or wild-type), five current states, three histories (decision at diagnosis, or untreated progression 1 or 4 months after diagnosis) and four treatments. The readouts of Extended Data Fig. 4c over the 3 months after the decision were the change along state-map PC1, the change in anatomical extent, the change in ECOG performance status and the maximum active toxicity grade. The readouts from the decision were the time before first-line progression, death by 12 months and overall survival. Time before first-line progression ran from the start of the first line to the first progression or death, whatever treatment changes occurred in between. For each input and readout, we grouped the futures by the levels of the three other inputs. Within each group, we computed the mean readout at each level of the input and took the difference between the highest and lowest means. We averaged this difference over groups and divided it by the median standard deviation among futures with identical inputs. We then subtracted the mean of the same statistic over 400 permutations of the input's levels within groups. A value of 0 thus means that the observed spread equals the mean spread after permutation. For the history input, we also report the proportion of permutations with a spread at least as large as observed, for the change along PC1.

**EGFR treatment comparisons.** Claude Sonnet 5, Kimi K3 and GLM-5.2 generated futures from the same 15 starting records (five states by three histories), with three futures per record and arm, in EGFR-mutant and EGFR-wild-type disease (23 to 28 July 2026). This gives 45 futures per arm. In GLM-5.2, rejected futures were not replaced, which left 35 EGFR-mutant and 42 EGFR-wild-type chemotherapy futures (Supplementary Table 4). For Fig. 3f, we computed for each model and EGFR status the median time before first-line progression with osimertinib minus that with carboplatin-pemetrexed chemotherapy. Intervals are 95% bootstrap percentile intervals over starting records (4,000 resamples, seed 20261028), with both arms of a record resampled together.

We also scored five orderings expected clinically: osimertinib longer than chemotherapy in mutant disease; osimertinib longer than supportive care in mutant disease; osimertinib not first among the three active treatments in wild-type disease; supportive care shorter than the best active treatment in both contexts; and a larger osimertinib-versus-chemotherapy difference in mutant than in wild-type disease. These checks assess the direction of the effects and not their duration. Their results are given in Supplementary Note 1.

**Time without progression by state.** For Extended Data Fig. 4d,e, we used the EGFR-mutant Claude Sonnet 5 futures and computed the restricted mean time before first-line progression over 24 months. We allocated each future's time before first-line progression to the states it passed through, holding each generated state until the next generated assessment, on a 28 × 28 grid of the state map. Extended Data Fig. 4e shows months per future in each cell, osimertinib minus chemotherapy, smoothed with a Gaussian kernel of 0.85 cells for display and cropped to the visited states. The three histories of each starting state share one position. The sums by region use the unsmoothed grid. In Extended Data Fig. 4d, each pair joins the k-th futures of the two arms from the same starting record, with 95% intervals from 5,000 bootstrap resamples of pairs.

**Switch to osimertinib.** To compare EGFR-mutant and wild-type chemotherapy futures without the effect of a later targeted treatment, we separated the EGFR-mutant chemotherapy futures that switched to osimertinib before first-line progression from the others, and compared the median time before first-line progression of the latter with that of the wild-type chemotherapy futures.

## Forecasting registry patients

**Data.** GENIE-BPC links curated clinical and genomic information[15,16]. We used the public releases NSCLC v2.0-public[32], CRC v2.0-public[33] and BrCa v1.0-public[34]. The SHA-256 checksums of the release files are recorded in the code repository. For each field used, we recorded its clinical meaning, the date from which it is known and its missingness. We kept stage at diagnosis separate from the extent of disease documented by the landmark. Metastatic sites came from the curated metastasis flags and their dates, and the regions imaged were not taken as disease sites. We distinguished the oncologist's assessment of the presence of cancer from response or progression, and a sequencing panel with no relevant alteration from a panel not yet reported. Performance status, weight, organ function and toxicity are not recorded in these files and were stated as unknown. The absence of a systemic-treatment record was not read as absence of surgery or radiotherapy.

**Landmarks and patients.** A landmark was the start of a named systemic regimen. We did not interpret the order of regimens as clinical lines of therapy. Eligible patients were alive at the landmark and either followed for at least 12 months after it or dead within 12 months. A recent disease assessment was not required. We selected one landmark per patient, stratified by regimen depth (first, second, third or later). For each landmark, comparable patients had the same cancer, the same class of treatment and a landmark year within 3 years, and we required at least 40 of them. Within each stratum of regimen depth, we took patients in random order (seed 20260824) until 50 had enough comparable patients, which gave 450 patients (150 each with lung, breast and colorectal cancer). Cohorts, exclusions and analysis populations are listed in Supplementary Table 6.

**Patient record supplied to the model.** The model received the patient record up to and including the landmark. It contained the regimen started at the landmark and the year of the landmark, and no information on the patient after it. We tested this cut by comparing records derived from complete files with records derived from files cut at the landmark, and by checking that future information introduced on purpose was detected.

**Four settings.** The four settings had identical patient records and differed only in added information. The first setting gave the record alone. The second added a sentence stating the care setting (an academic North American center and the year of the landmark). The third and fourth added summaries of 20 example patients from the same cohort, chosen at random among patients with the same cancer and era (generic examples) or as the most similar patients (matched examples). Matched examples shared cancer, set of treatment modalities and era with the patient. We then ranked them by agreement on clinical criteria that a clinician can check: presence of disease, stage at diagnosis, metastatic extent, driver, current status, regimen depth, sex, age at the landmark, time from diagnosis to the landmark, metastatic sites where recorded, and histology. Each criterion scored +1 for agreement, -1 for disagreement and 0 when unknown, and we averaged the scores over the criteria informative in each cohort (11 in lung, 10 in colorectal and 8 in breast cancer). Age agreed within 10 years and time from diagnosis to the landmark within a factor of 2. A patient could not be their own example. Each example patient was described at their own landmark, by their history before it and their observed course after it. We checked that all matching variables were present in the patient information rendered in the prompts. Generic and matched examples used the same format, without labels or similarity scores. Generic examples excluded the matched ones.

**Generation.** The model was DeepSeek V4 Flash (deepseek/deepseek-v4-flash-0731, the identifier returned for every attempt), called on 26 August 2026 through OpenRouter, restricted to three inference providers (Novita, SiliconFlow and DeepInfra), with data collection denied and zero data retention requested. Calls requested temperature 1.0, top-p at the provider default, reasoning off and at most 24,576 output tokens. We requested ten futures per patient and setting, each until death, with at most three attempts per future. Of the 18,000 requested futures, 17,679 were accepted (Supplementary Table 4).

**Reading generated and recorded courses on the same dates.** We compared each generated future with the patient's recorded course on the dates of that patient's assessments. We first reconstructed the generated course and expressed it in the GENIE-BPC vocabulary. We kept the presence of cancer and its response status as separate variables. In the main analysis, a documented state remained in effect until a later assessment updated it, the patient died or follow-up ended. The first assessment after the landmark was not read backward, and pre-landmark information that had not been supplied to the model was not reused. A later assessment updated only the information it reported, and incompatible records at the same date were counted as conflicts. Clearing one lesion did not count as complete response while another known lesion remained active. Generated time without a resolved state, during periods documented for the patient, counted as disagreement. Generated courses recorded response and progression as events but had no event for stable disease, so a generated course was stable only when it started stable, and a response lasted until the next progression. In a sensitivity analysis, an assessment remained valid for at most four months. Recorded times were converted to months by dividing days by 30.4, and generated times by 30.4375.

**Survival forecasts.** For each patient and setting, the LLM probability of being alive at a horizon is the proportion of accepted futures alive at that horizon, computed when at least seven accepted futures were available and all resolved survival at that horizon. A future ending without death before the horizon was treated as truncated and not extended. We compared the LLM with two types of forecast. The first was a cancer-stratified Cox model[35] fitted on the 450 patients from the landmark, with 14 covariates known at the landmark: age, sex, stage IV at diagnosis, metastases documented by the landmark, regimen number, months from diagnosis to the landmark, disease state at the landmark (four indicators), progression in the previous 3 months, squamous histology, panel reported and driver present. We used a ridge penalty of 0.1 (lifelines 0.30.3) and ten-fold cross-validation stratified by cancer and death (seed 20260908), so that each patient's prediction came from the model fitted without that patient's fold. Missing values were replaced by the median of the training folds, with an indicator for each variable with missing values. The second was the Kaplan-Meier estimate[36] of the example patients themselves, computed separately for the 20 generic and the 20 matched example patients of each patient, from their own follow-up and not extended beyond it. This gives seven forecasts per patient: the four LLM settings, the generic and matched example patients, and the Cox model.

We evaluated forecasts at 3, 6, 12 and 24 months, each horizon on its own population. Survival status is unknown after earlier censoring. Brier scores use inverse probability of censoring weights[37] estimated within cancer on the 450 patients, with left limits and without capping. The comparison population holds the patients for whom all seven forecasts were available (449 at 3 to 12 months, 447 at 24 months). Calibration error uses at most five bins that never split identical predictions. We computed Harrell's C-index[38] on each forecast's restricted mean survival[20] to 24 months. Absolute errors in survival time were computed among patients with an observed death (Extended Data Fig. 5f). Coverage was the share of these patients whose observed survival lay between the 10th and 90th percentiles of their generated survival times (Extended Data Fig. 5c). In a sensitivity analysis, we corrected for the finite number of futures per patient. The survival estimated directly from the generic example patients was compared with that from the matched example patients in the same way (Supplementary Note 1).

**Disease state over time.** For Fig. 5a, we computed the share of patients in each disease state (responding, stable, progressing, no active disease, unresolved, dead) at every month from 0 to 24 after the landmark, for the recorded course and for each setting. We used the 437 patients of the 24-month disagreement analysis (Extended Data Fig. 7a), the same in every setting. At each month, a death on or before that month counted as dead. A patient whose follow-up had ended without death was excluded from that month for every source, so censored time was never counted as unresolved. We averaged generated shares over each patient's futures first (7 to 10 accepted futures per setting), then over patients. The shares are distributions among the patients retained at each month and not population estimates corrected for censoring. In a sensitivity analysis, we read the generated futures as stated by the model, without the assessment dates of the patient.

**Disagreement between courses.** On the documented follow-up within a horizon, including known time after death, we compared each future with the recorded course by two measures, both as fractions of the window. Time-aligned disagreement is the share of time in a different state (Extended Data Fig. 7a,b). Occupancy total variation is the difference in time spent in each state, regardless of order. We averaged these measures over a patient's futures, then over patients, so each patient has equal weight. Treatment exposure was the documented exposure, and an absent or unresolved record was not read as absence of treatment. Treatment duration used unions of intervals, so overlapping regimens were not counted twice. Drug classes followed a fixed table, and an unrecognized agent left its modality unresolved. Treatment summaries (Fig. 5b; Extended Data Fig. 7c) compare each forecast's mean with the documented value in clinical units. Generic and matched example patients contributed only those followed through the horizon, at least seven per patient. New sites were not scored where site-level data were not collected, and exact drug sequences were not compared.

**Joint score.** To score whole courses (Fig. 5d; Extended Data Fig. 7d; Supplementary Fig. 11b,c), we described each course at a horizon by 17 summaries in four blocks of equal weight. The survival block contained being alive at the horizon and months alive. The disease block contained months with no evidence of disease, responding, stable and progressing. The event block contained documented entries into progression and into response. The treatment block contained months on any systemic therapy and on each of four modality classes, subsequent regimens, modality changes, whether a change occurred and when. Durations were divided by the horizon and counts n by 1 + n. We scored a patient's futures against the observed summaries with the energy score, a proper score for a vector of summaries[39]. We computed it on equal numbers of futures across the four settings (the smallest accepted number, with 200 fixed random subsets). The score requires the observed patient and every accepted future to resolve all 17 summaries, and no value was imputed, so each horizon has its own population. To test whether the relations between blocks mattered, we permuted blocks across futures, which keeps each block's distribution. Sensitivity analyses repeated the score with all futures as delivered, with an estimator for independent draws, with only the futures accepted at the first attempt, and with recorded states valid for at most four months.

**Comparison with the example patients.** We compared the courses generated with examples with the courses of the example patients themselves, separately for generic and matched examples. For each patient and example patient, we defined the window on which both were documented on the patient's assessment dates (at least one month). On each window, we compared the mean disagreement of the patient's futures in the setting that received these examples with the disagreement of the example patient's own course. We then averaged over usable example patients (at least seven) and over patients (Extended Data Fig. 5d). In a secondary version, unresolved generated time was excluded from both sides of each pair.

**Gain from matched examples by agreement of the matched patients.** For Supplementary Fig. 16, we tested whether the gain in the 12-month joint score from matched over generic examples depended on how closely the matched patients agreed with the patient. We measured agreement by the clinical agreement score of the weakest of the 20 matched patients, ranked within cancer. We computed the Spearman correlation between the gain and this rank, with a 95% interval from 2,000 bootstrap resamples stratified by cancer and a two-sided P value from 2,000 permutations within cancer. We repeated the analysis for the gain of matched examples over the record alone.

**Display in the Clinical State Space.** GENIE-BPC does not record host variables, so we placed patients from their imaging. For each of eight tumor classes, we read the last scan of the previous four months that covered this class. A class without such a scan stayed unknown, so that an unscanned region never counted as recovery. We located a patient when the hepatic and thoracic classes were known (109 of 150 patients with lung cancer and 116 of 150 with colorectal cancer at the landmark). Other unknown classes took the observed mean, on the map only. The axes are the first two principal components of these classes and of disease extent, fitted on the observed positions with equal weight per patient. We read generated courses on the patient's assessment dates and projected them on the same axes. The GENIE-BPC population displays show aggregates only. Counts of 1 to 4 appear as <5 and the matching cells are blank.

**Individual courses.** Supplementary Figs 11a, 12 and 13 show each future of one patient as a separate timeline of disease state, systemic treatment class and death. We chose each patient in two steps from the observed course only: ten accepted futures in every setting, at least 18 months of observed follow-up and at least two changes of documented disease state within 24 months, one of them a progression. Among these patients, we took the one with the smallest SHA-256 hash of its identifier. Futures with the same index in different settings are not paired.

## PRIME trial

**Patients and endpoints.** The PRIME dataset shared through Project Data Sphere[19] contains 935 of the 1,183 randomized patients[17] (469 panitumumab plus FOLFOX, 466 FOLFOX alone). KRAS exon 2 (codons 12 and 13) was wild-type in 514, mutant in 352 and unknown or failed in 69. Survival analyses included all 935 patients as randomized, with progression-free survival by central review (progression or death). The protocol defines overall survival as the time from randomization to death, censored at the last contact for patients alive at the data cut-off (protocol amendment 2, section 10.2.1.2), and we used the dataset's death-day field as this time. Endpoints were used as supplied. No language model was used in the PRIME analyses.

**Survival and time in clinical states.** We compared restricted mean progression-free survival over 24 months between arms in each predefined baseline subgroup (Extended Data Fig. 8a). The tumor-burden subgroup used the 488 patients with lesion measurements. Hazard ratios (Extended Data Fig. 8b) came from separate Cox models, without an interaction test. For Fig. 6a, we divided the first 24 months after randomization of each arm, within each KRAS exon 2 group, into time alive without progression (restricted mean progression-free survival), time alive after progression (restricted mean overall survival minus restricted mean progression-free survival) and time dead (24 months minus restricted mean overall survival), from Kaplan-Meier curves. Patients with unknown or failed KRAS tests were excluded. Intervals for Fig. 6a and Extended Data Fig. 8a come from one shared set of 2,000 bootstrap resamples of patients within arm and KRAS group (seed 20261028), so the progression-free values of the two panels agree. Months are days divided by 30.4375.

**Display in the Clinical State Space.** The PRIME axes are the first two principal components of 14 trial features, fitted on the trial (16.5% and 10.7% of the variance): target-lesion size and its change, lesion size in liver, lung, lymph nodes and other sites, new lesions, low albumin, low hemoglobin, lactate dehydrogenase, alkaline phosphatase, creatinine, weight loss and maximum adverse-event grade. Laboratory values and weight were taken from the last measurement within 35 days before each assessment. Values still missing were replaced by the median of that feature over all assessments, and features were standardized before the components were fitted. Of the 866 patients with known KRAS exon 2 status, 439 had a lesion assessment by day 14 and form the map population. A progression-free survival event on the date of a death is shown only as that death, and unevaluable responses stay missing. We compared arms only from randomization.

## Availability of comparable patients

**Registry.** In GENIE-BPC (Fig. 6d), we counted the other patients who shared a decision's description at eight cumulative steps: cancer, stage at diagnosis, driver, regimen depth, treatment class started, latest disease status, previous regimen, and age band with sex. A decision was the start of a recorded systemic regimen after metastatic disease was documented, in a living patient with at least one assessment between metastatic onset and the decision. Only information dated on or before the decision was used. Drivers were, for lung cancer, selected EGFR alterations (L858R, exon 19 in-frame deletion, exon 20 insertion, G719X, L861Q, S768I), ALK fusion or KRAS G12, G13 or Q61, in that order of precedence. For colorectal cancer, they were mismatch-repair deficiency or high microsatellite instability, BRAF V600E or KRAS hotspot mutations, and for breast cancer, the index receptor subtype. A sequencing result counted from its panel report date. Regimen depth was the order of the regimen among those started after metastatic onset. The treatment class combined the agents of the regimen known at the decision, classified with the drug table above, and an unrecognized agent was kept as unclassified. The latest status was the last documented disease status, with no active disease taken from the record's statement of no evidence of cancer. The previous regimen was described by its best documented status and its duration (<3, 3-6, 6-12 or ≥12 months). When one of its drugs had no stop date, the regimen was taken to run until the start of the next regimen. We drew one decision per patient (seed 20261028), and the comparators of a decision were the other patients' drawn decisions with the same description. A missing value matched only the same missing value. In sensitivity analyses, missing values matched nothing, and we used all decisions and counted distinct other patients.

**Atlas.** In the generated atlas (Supplementary Fig. 15), we matched each trajectory to the other trajectories on baseline characteristics added in turn: primary-site class, stage, molecular driver, baseline condition (performance status and disease extent), age and sex, and exact histology (OncoTree code). We report the median number of other trajectories sharing all criteria. For the 3,503 trajectories with a comparator at every step, we computed the median absolute difference between a trajectory's survival and the median survival of its comparators. We compared it with the same number of comparators drawn at random from all other trajectories, over 10 draws.

## Statistical analysis

The patient is the unit of analysis in GENIE-BPC and PRIME, and patient-level resampling keeps each patient's settings, futures and example patients together. In the atlas, we resampled cancer types for the baseline concentration and the differences between cancers, generated patients within each model and family for survival by family, and whole trajectories for the fields. In the controlled experiments, we resampled starting records, except for the pairs of Extended Data Fig. 4d. Unless stated otherwise here or in the figure legends, intervals are 95% percentile intervals from 2,000 bootstrap resamples. In GENIE-BPC, resampling was stratified by cancer (seed 20260928), and the censoring weights were estimated again in each resample. Paired differences between settings carry two-sided Wilcoxon signed-rank tests[40]. In the registry experiment, two families of tests were adjusted by the Holm method[41] on unrounded P values: the joint score (four contrasts between settings at four horizons, 16 tests) and the comparison with the generic and matched example patients (two comparisons at four horizons, 8 tests). Other analyses are descriptive or sensitivity analyses, without adjusted P values. All analyses are exploratory.

### Software and reproducibility

Analyses used Python 3.12, with dependencies pinned in a lockfile. Each run retains the raw responses, rejected attempts, returned model and provider identifiers, rendered prompts with the output schema, and generation settings. Run manifests record the code commit and the SHA-256 hashes of the inputs of the generation runs (Code availability). Each analysis output records the hashes of its code and inputs. The registry prompts contained no patient identifiers, institution names or calendar dates. They kept the year of diagnosis and of the treatment start, and intervals in days. Patient-level registry and trial records, and patient-level files derived from them, are not distributed.

### Use of large language model assistants

Claude Code (Anthropic) and Codex (OpenAI) assisted with writing and reviewing code, checking sources and numbers, and editing the text. The authors specified the analyses, reviewed all code and outputs, made the scientific decisions and take responsibility for the content.

Reporting follows the TRIPOD-LLM guideline[42] (Supplementary Table 7).

## Data availability

The trajectories of the cancer atlas and of the controlled experiments were generated from cancer labels and constructed scenarios and contain no patient information. They are deposited on Zenodo under a CC BY 4.0 license (https://doi.org/10.5281/zenodo.24137512), with the raw model responses, rejected attempts, the rendered prompt of each attempt (which contains the output schema) and SHA-256 checksums. Run manifests record the code commit, model identifiers, settings and input checksums of the runs, and each run folder holds its configuration file. The futures generated for GENIE-BPC patients are not deposited, because each was conditioned on a registry record. GENIE-BPC data (NSCLC v2.0-public, CRC v2.0-public and BrCa v1.0-public) are available to registered users who accept the AACR Project GENIE terms of access, through the GENIE Biopharma Collaborative public page on Synapse (syn27056172). Patient-level PRIME data (NCT00364013; dataset Colorec_Amgen_2006_264) are available from Project Data Sphere (https://www.projectdatasphere.org) after registration and acceptance of its data sharing agreement. We do not share patient-level GENIE-BPC or PRIME data, or patient-level files derived from them. Source data are provided with this paper. They give the plotted values for panels built from generated data and, for panels built from GENIE-BPC or PRIME data, aggregate tables without direct identifiers, in which values showing or revealing counts of 1 to 4 patients are suppressed. Kaplan-Meier curves of registry and trial patients (Fig. 4b; Extended Data Figs 6a, 8c and 9a; Supplementary Figs 10a and 14a) and panels that show individual real patients (Fig. 6b; Extended Data Fig. 9c; Supplementary Figs 11a, 12, 13 and 14c) can be rebuilt only with access to the GENIE-BPC or PRIME data.

## Code availability

The code to generate trajectories, analyze them and reproduce the figures is available on GitHub (https://github.com/julienvibert/llm-oncology-world-models, release v1.0.1) and archived on Zenodo (https://doi.org/10.5281/zenodo.24137512) under the Non-Commercial Software License – Heidelberg University Version 1.0 (NC-SA-UHDV1.0). With the deposited model responses, the analyses of the synthetic experiments can be rerun without access to the model APIs. Analyses of GENIE-BPC or PRIME data require access to these data, and the registry forecasting comparisons also require the generated futures, which are not deposited.

## Acknowledgements

The authors would like to acknowledge the American Association for Cancer Research and its financial and material support in the development of the AACR Project GENIE registry, as well as members of the consortium for their commitment to data sharing. Interpretations are the responsibility of study authors. This publication is based on research using information obtained from www.projectdatasphere.org, which is maintained by Project Data Sphere. Neither Project Data Sphere nor the owner(s) of any information from the web site have contributed to, approved or are in any way responsible for the contents of this publication. J.V. is supported by the Ligue contre le Cancer and by the Fondation Nuovo-Soldati. J.N.K. is supported by the Dietmar Hopp Stiftung through the Hopp Foundation Professorship for Artificial Intelligence in Medicine, the German Cancer Aid DKH (DECADE, 70115166), the German Federal Ministry of Research, Technology and Space BMFTR (PEARL, 01KD2104C; TRANSFORM LIVER, 031L0312A; DEEP-HCC, 031L0315A; DECIPHER-M, 01KD2420A; PROSURV, 01KD2509C; IMPACT-CRC, 01KU2605), the German Research Foundation DFG (SFB 1709/1 2025, 533056198), the European Union (ODELIA, 101057091; GENIAL, 101096312; IMPACT-OC, 101306266), the European Research Council ERC (NADIR, 101114631), the Breast Cancer Research Foundation (BELLADONNA, BCRF-25-225) and the National Institute for Health and Care Research NIHR (Leeds Biomedical Research Centre, NIHR203331). The views expressed are those of the author(s) and not necessarily those of the NHS, the NIHR or the Department of Health and Social Care. This work was funded by the European Union. Views and opinions expressed are, however, those of the author(s) only and do not necessarily reflect those of the European Union. Neither the European Union nor the granting authority can be held responsible for them.

## Author contributions

J.V. conceived the study, developed the methods and software, performed the analyses and wrote the manuscript. J.N.K. contributed to the conception of the study, supervised the work and revised the manuscript. Both authors approved the final version.

## Competing interests

J.N.K. holds shares in and serves as scientific advisor of StratifAI, Synagen, Spira Labs, Tremont AI and Saterra AI; is co-principal investigator on institutional research grants from GSK and AstraZeneca; and declares honoraria or consulting fees from AstraZeneca, Bayer, Bioptimus, Daiichi Sankyo, Eisai, Janssen, Merck, MSD, Novartis, BMS, Roche and Pfizer. J.V. declares no competing interests.

# Figure 1 | Six LLMs generate cancer-specific patients built on model-specific archetypes

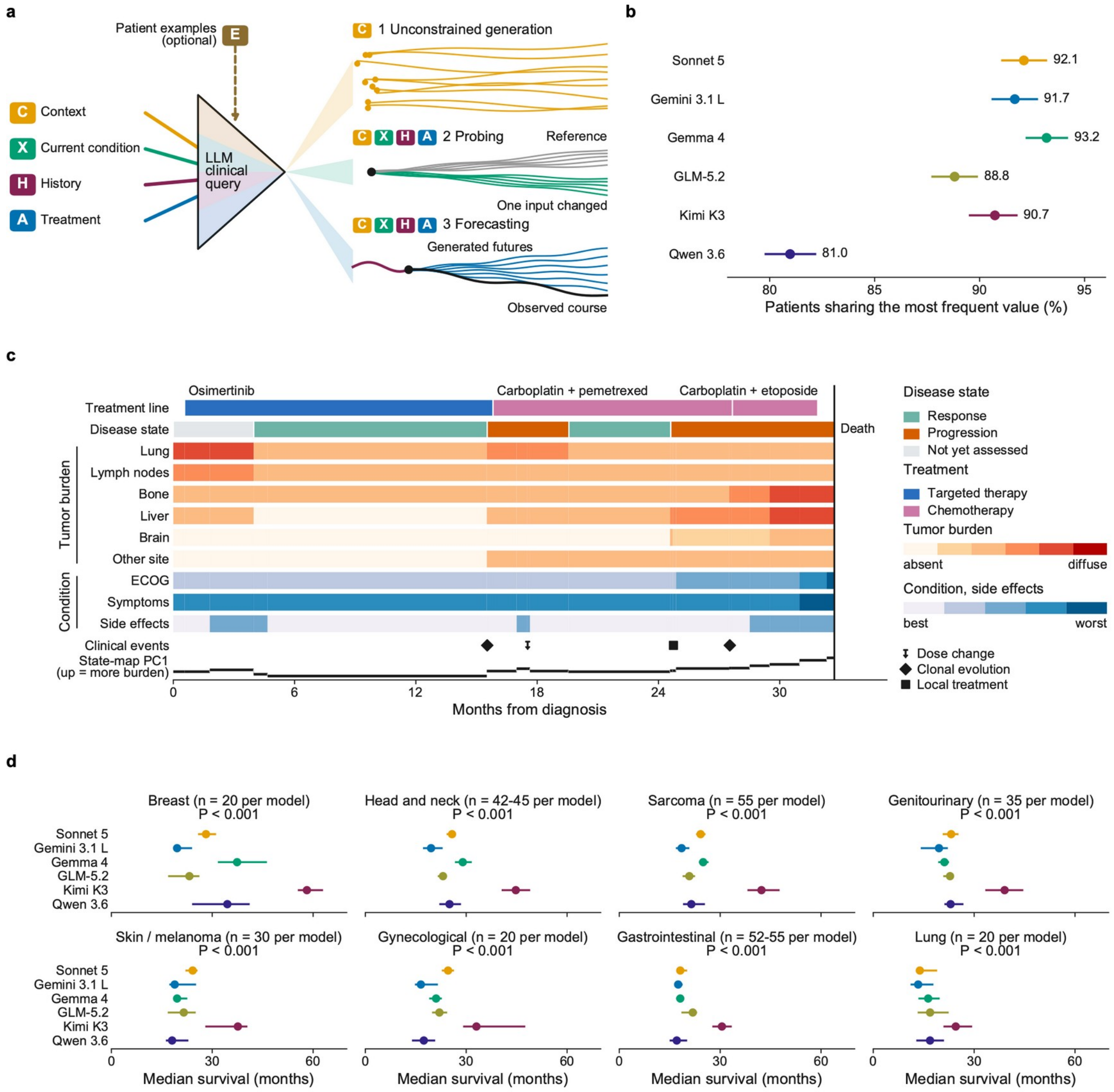

**a**, Study design. A clinical query fixes some of a patient's context (C; such as cancer type, stage, molecular profile), current condition (X), history (H) and treatment (A); patient examples (E) are an optional input. Three experiments: unconstrained generation from the cancer type alone, probing with one input changed at a time, and forecasting real patients of GENIE-BPC under their recorded treatment. **b**, Share of generated patients with the most frequent value of five baseline features (primary driver, sex, stage, smoking status, frailty) within a cancer type, averaged over the five features and 148 cancer types. Dots, mean; lines, 95% bootstrap intervals over cancer types. **c**, One generated patient with metastatic EGFR-mutant lung adenocarcinoma (Claude Sonnet 5; selection in Methods): treatment lines, disease state, tumor burden by organ, physical condition, side effects, events, and position on the first axis of the state map. Vertical line, death. **d**, Median survival from diagnosis in eight cancer families (56 of the 148 cancer types; n, generated patients per model). Dots, median; lines, 95% bootstrap intervals (2,000 resamples). P, Kruskal-Wallis test between models, Holm-adjusted across the eight families.

## Figure 2 | The Clinical State Space represents a patient's current state, history and possible future courses

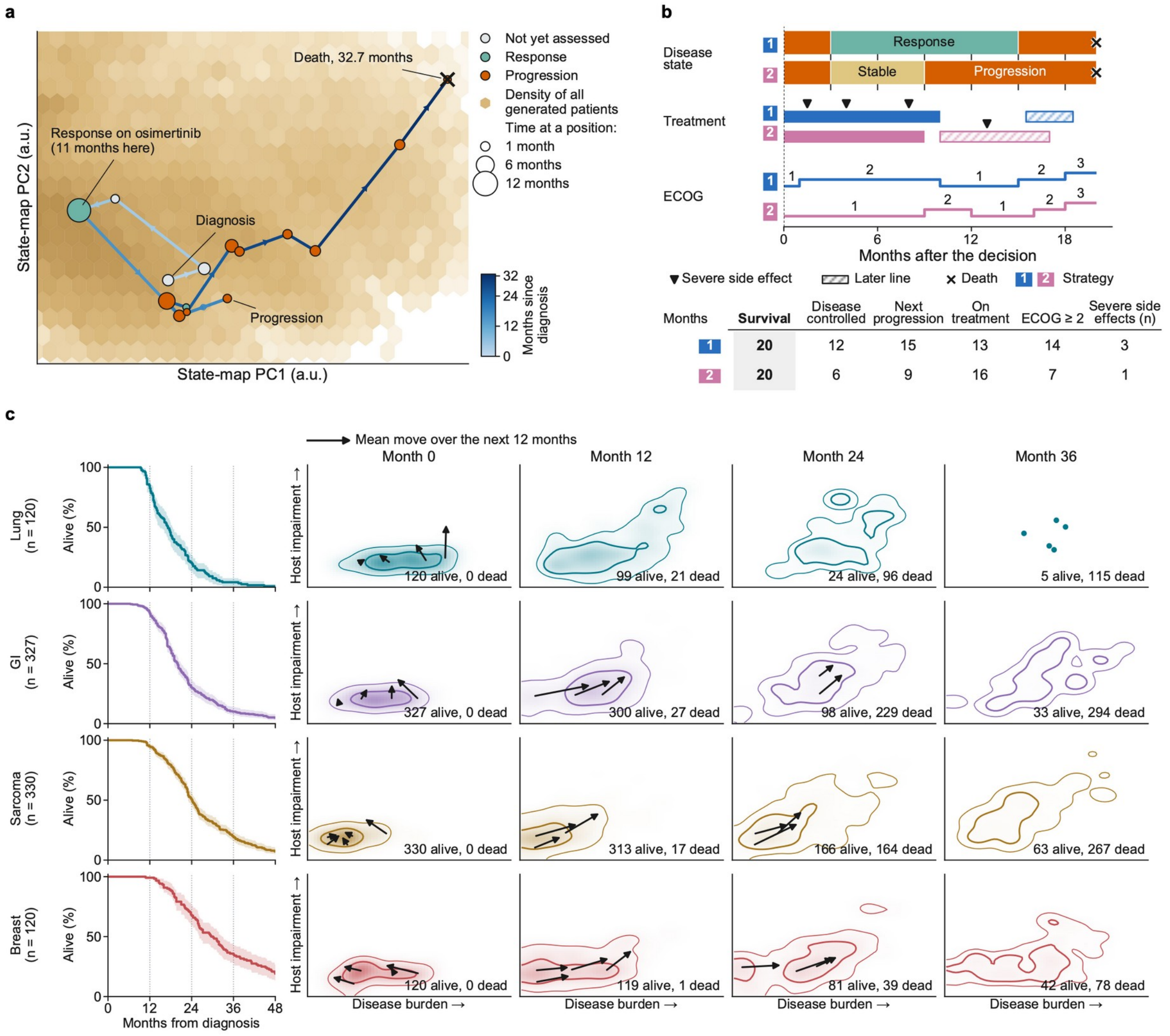

**a**, The patient of Fig. 1c (Claude Sonnet 5, metastatic EGFR-mutant lung adenocarcinoma) on the state map, the first two principal components of the 20 state variables: points, positions colored by disease state, with area proportional to the time spent; line colored by months since diagnosis; cross, death. Shading, density of all 105,839 generated positions (log scale). **b**, Two constructed courses with the same survival under two strategies, from progression to death 20 months later: disease state, treatment (hatched, later lines; triangles, severe side effects) and ECOG performance status; table, readouts of each course. A constructed illustration, not a generated or observed patient. **c**, Generated patients of four cancer families (six models pooled; 897 patients). Left, Kaplan-Meier survival (band, 95% interval; dotted lines, frame times). Right, the same patients on the disease and host axes (first principal component of the 10 tumor and of the 10 host variables) at 0, 12, 24 and 36 months from diagnosis, with numbers alive and dead. Contours, 50% and 90% of the positions of living patients; shading, density relative to month 0; arrows, mean move over the next 12 months of the patients still alive 12 months later, drawn from their positions in the frame shown (a summary of survivors, not a forecast), in each of 36 regions with at least 8 such patients (none at month 36, when too few patients were alive at 48 months; dots, moves too short to draw); with fewer than 20 living patients, points show the patients. Outcome by starting position and clinical course in Extended Data Fig. 3.

# Figure 3 | Generated trajectories under controlled changes in patient features and treatment

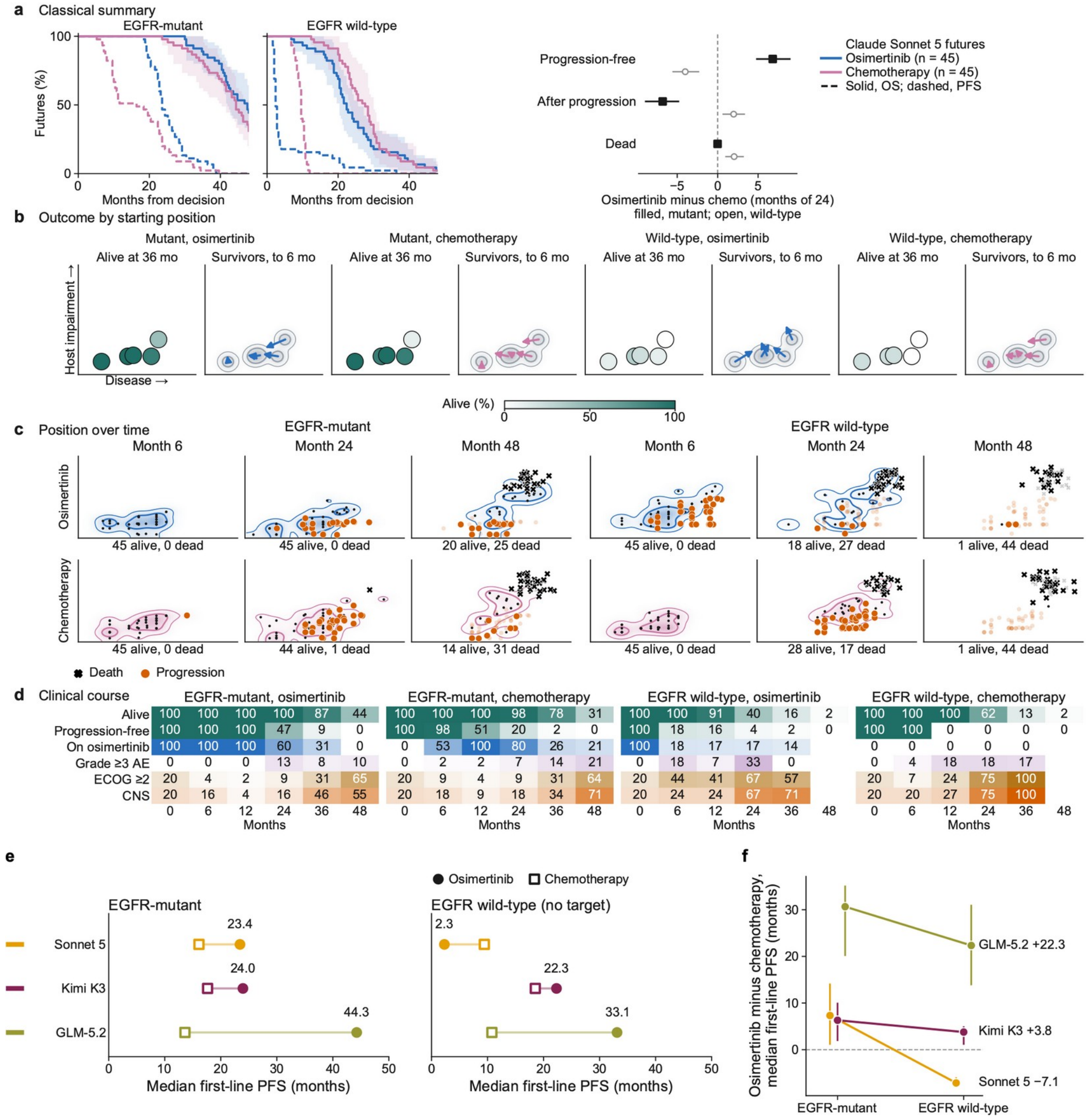

**a**, Claude Sonnet 5 futures of EGFR-mutant and EGFR wild-type lung adenocarcinoma with osimertinib or carboplatin-pemetrexed (45 futures per arm and status, from 15 starting records). Left, Kaplan-Meier overall survival (solid) and first-line progression-free survival (dashed); bands, 95% intervals. Right, the first 24 months divided into time without progression, after progression and dead, osimertinib minus chemotherapy (filled, EGFR-mutant; open, wild-type; 95% bootstrap intervals over starting records). **b**, Overall survival at 36 months by starting state (blank, too few futures) and mean move to month 6 of survivors. **c**, Positions at 6, 24 and 48 months. **d**, Clinical course: Kaplan-Meier survival and, among living futures with a known value, the share in each state; blank, fewer than 5 futures. **e**, Median first-line progression-free survival with osimertinib (filled) and chemotherapy (open) generated by Claude Sonnet 5, Kimi K3 and GLM-5.2. **f**, Osimertinib minus chemotherapy, median first-line progression-free survival; 95% bootstrap intervals over starting records (4,000 resamples). In e and f, 45 futures per arm and status, except GLM-5.2 chemotherapy (35 EGFR-mutant, 42 wild-type).

# Figure 4 | Forecasting survival in real patients

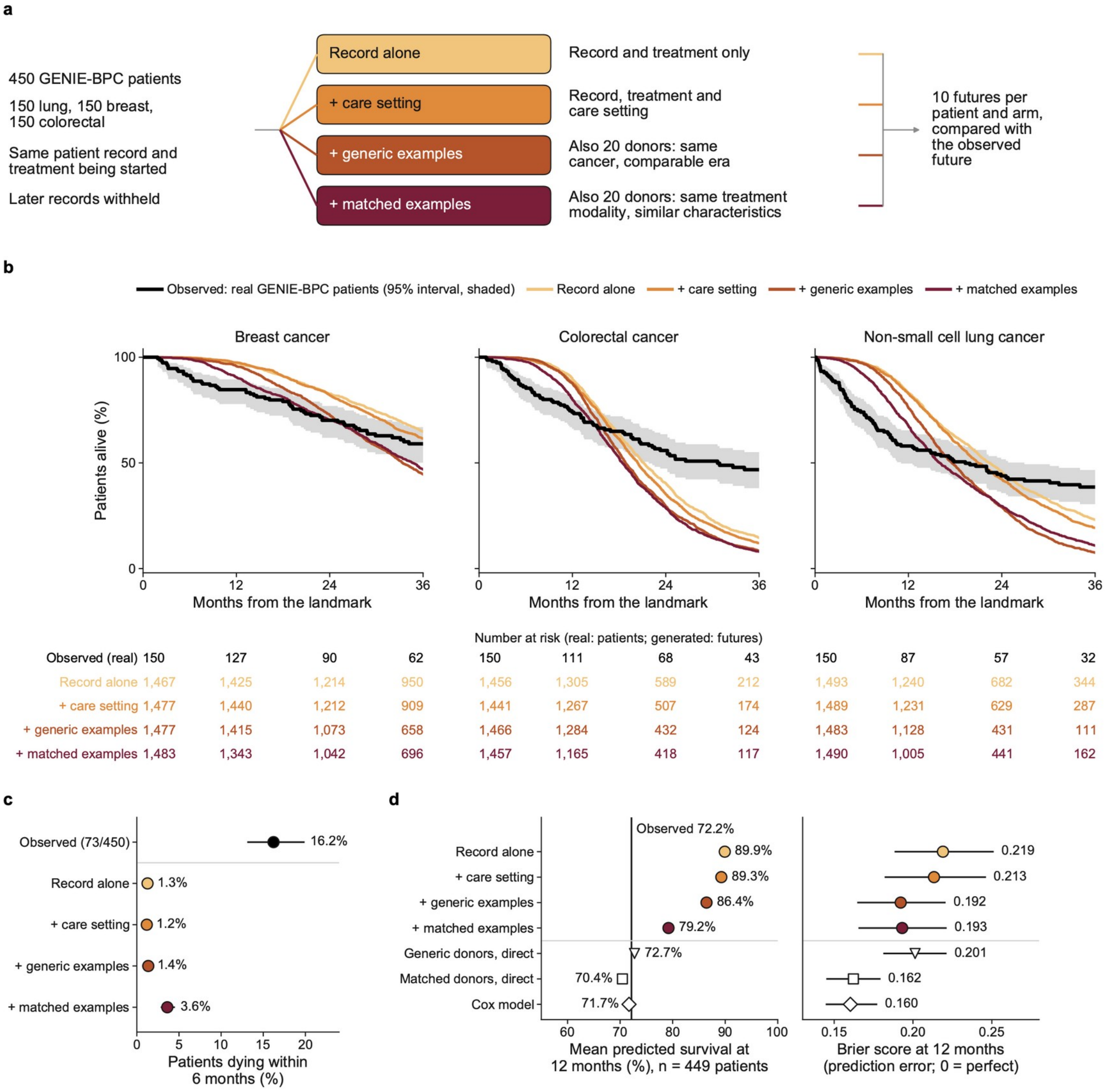

**a**, Design. DeepSeek V4 Flash received the record of 450 GENIE-BPC patients (150 each with lung, breast and colorectal cancer) up to the start of a recorded systemic treatment, and generated the rest of the course, with ten futures per patient and setting. Four settings: record alone; plus care setting; plus 20 generic examples (same cancer type and era, chosen at random); plus 20 matched examples (same treatment modality and similar recorded characteristics). **b**, Survival from the start of the forecast treatment, by cancer. Black, Kaplan-Meier survival of the observed patients (band, 95% confidence interval); colored, generated futures. Table, numbers at risk. **c**, Deaths within 6 months: observed patients (Wilson 95% interval) and generated futures by setting (95% bootstrap intervals over patients). **d**, Survival at 12 months in the 449 patients with all seven forecasters. Left, mean predicted probability of being alive (line, observed survival, censoring-weighted). Right, Brier score (0, perfect). Open symbols, survival of the generic or matched example patients used directly, and a Cox model cross-validated on the other patients. Lines, 95% bootstrap intervals over patients; paired tests in Methods.

# Figure 5 | Whole courses expose discrepancies between generated and observed trajectories that survival misses

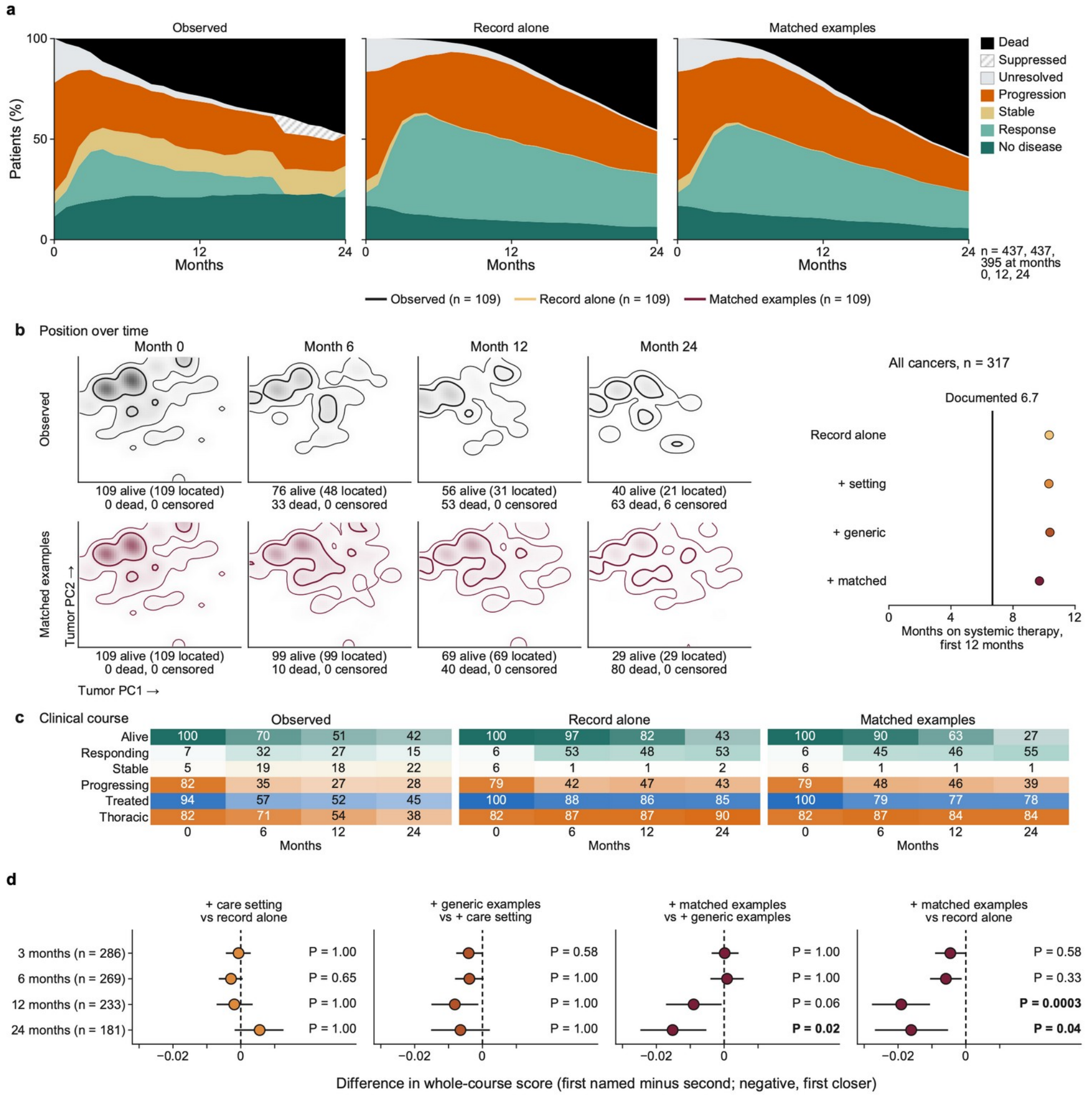

**a**, Disease state month by month over 24 months from the start of the forecast treatment, in the registry and in the futures generated by DeepSeek V4 Flash with the record alone and with matched examples, for 437 GENIE-BPC patients of all three cancers (all settings in Extended Data Fig. 7f). Shares among patients who had died or remained under follow-up at each month; censored patients leave the denominator (437, 437 and 395 patients at months 0, 12 and 24). Hatched, registry states with 1 to 4 patients drawn together with one other state, to prevent disclosure; not an unknown clinical state. At months 19 to 23, the response and unresolved states of the registry are pooled in the hatched band, so the interruption of the response band does not indicate an absence of response. **b**, Lung cancer, 109 patients located at the start: positions at 0, 6, 12 and 24 months, observed and with matched examples, read on each patient's assessment dates; generated counts are patient equivalents (each future weighted by one over the number of the patient's futures) (all groups, survival and starting positions in Supplementary Fig. 10; colorectal cancer in Extended Data Fig. 6). Right, months on systemic therapy in the first 12 months, mean forecast against documented mean, in 317 evaluable patients of all three cancers. **c**, Clinical course of the same 109 patients, observed and in the two settings. **d**, Whole-course score (an energy score on 17 trajectory summaries; lower is better): the setting named first minus the setting named after “vs” (negative, the first is closer to the observed course), at four horizons, all three cancers (286, 269, 233 and 181 patients). Points, mean paired difference; lines, 95% bootstrap intervals (2,000 paired resamples stratified by cancer); P, two-sided Wilcoxon signed-rank test, Holm-adjusted across 16 contrasts. Colorectal cancer, Extended Data Fig. 6.

## Figure 6 | The Clinical State Space applied to an observed randomized trial, and the support of comparable patients

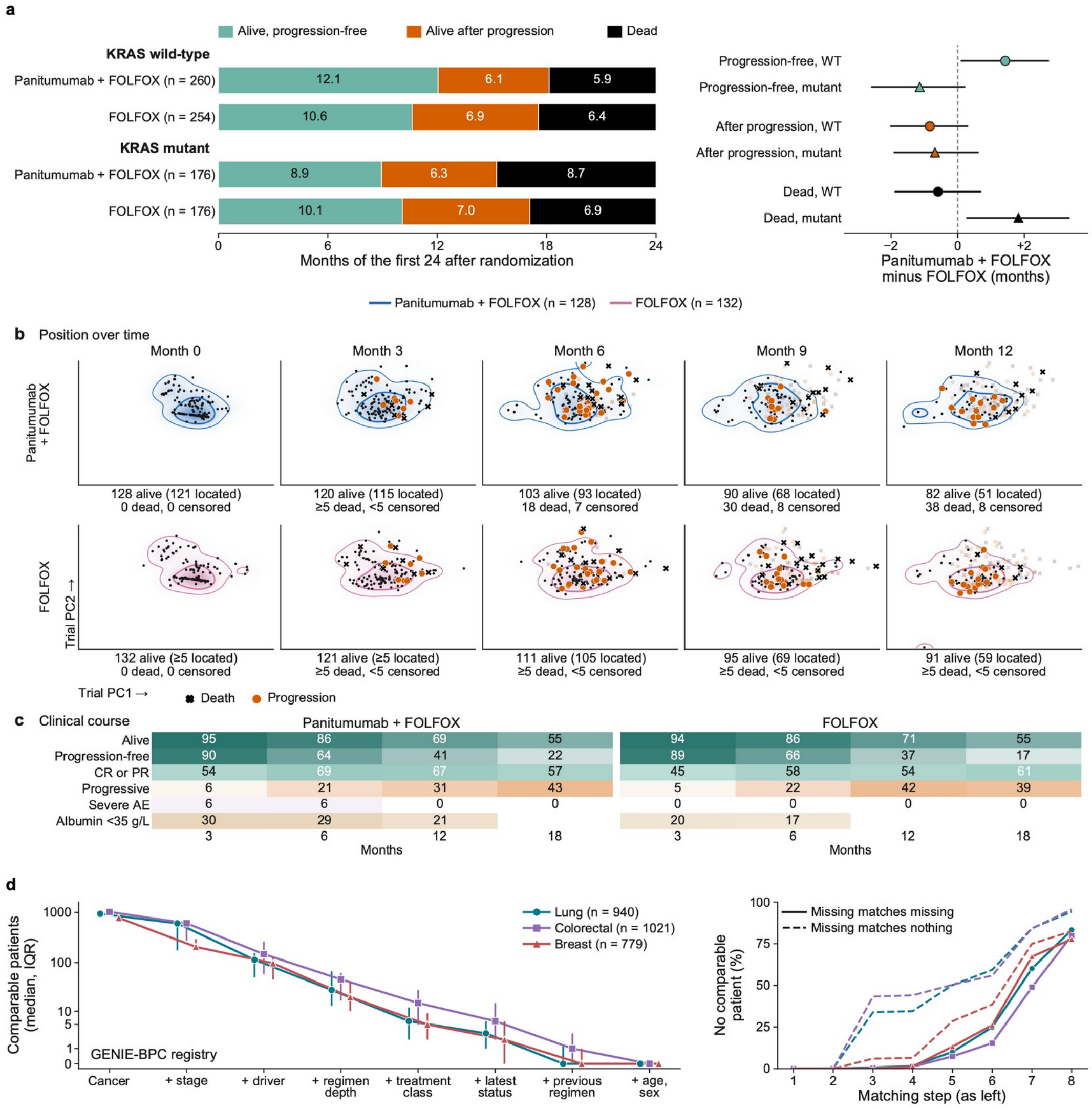

**a**, PRIME (panitumumab + FOLFOX versus FOLFOX in first-line metastatic colorectal cancer): restricted mean months alive without progression, alive after progression and dead over the first 24 months after randomization, by arm and KRAS exon 2 status (left), and panitumumab + FOLFOX minus FOLFOX for each part (right; 95% bootstrap intervals, 2,000 resamples within arm). **b**, KRAS wild-type patients with a lesion assessment by day 14 (128 and 132): positions at 0, 3, 6, 9 and 12 months on the first two components of 14 trial features, among patients assessed in the preceding three months; progressions and deaths at the last assessment. Starting positions and survival, Supplementary Fig. 14; mutant disease, Extended Data Fig. 9. **c**, Clinical course of the same patients: survival and, among patients assessed in the interval, RECIST response, severe adverse events and albumin below 35 g/L. In b and c, counts of 1 to 4 patients are shown as <5 (and one other count as ≥5 where needed to prevent recovery by subtraction) and the matching cells are blank. **d**, Patients comparable to a treatment decision in GENIE-BPC as criteria are added from left to right (one decision per patient: lung 940, colorectal 1,021, breast 779). Left, number of other patients matching all criteria so far (median and interquartile range); right, share of decisions with no comparable patient (dashed, missing values match nothing).

## Extended Data Figure 1 | Clinical phenotypes generated by six models

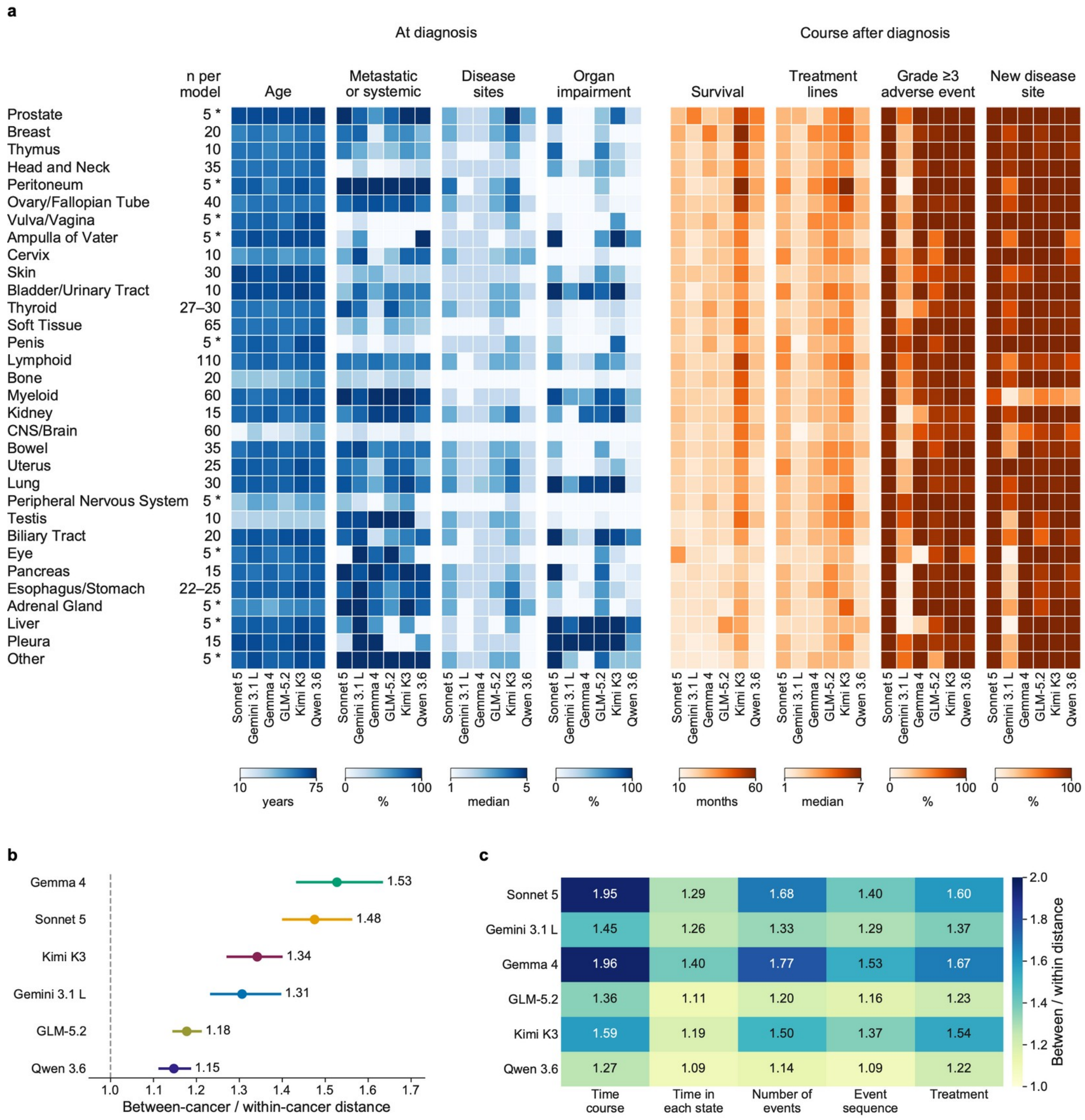

**a**, Clinical characteristics generated by each model, by tissue (4,434 trajectories, 148 cancer types, 32 tissues). Left, at diagnosis: median age, share with metastatic or systemic stage, median number of disease sites, share with any organ impairment. Right, after diagnosis: median survival, median number of treatment lines, share with an adverse event of grade 3 or higher, share with a new disease site. Asterisk, fewer than 10 patients per model. **b**, Distance between courses of different cancers divided by distance between courses of the same cancer (dashed line, no cancer-specific structure). Dots, estimate; bars, 95% bootstrap intervals over 146 cancers. **c**, The same ratio for each of five components of the distance.

# Extended Data Figure 2 | History and possible future courses

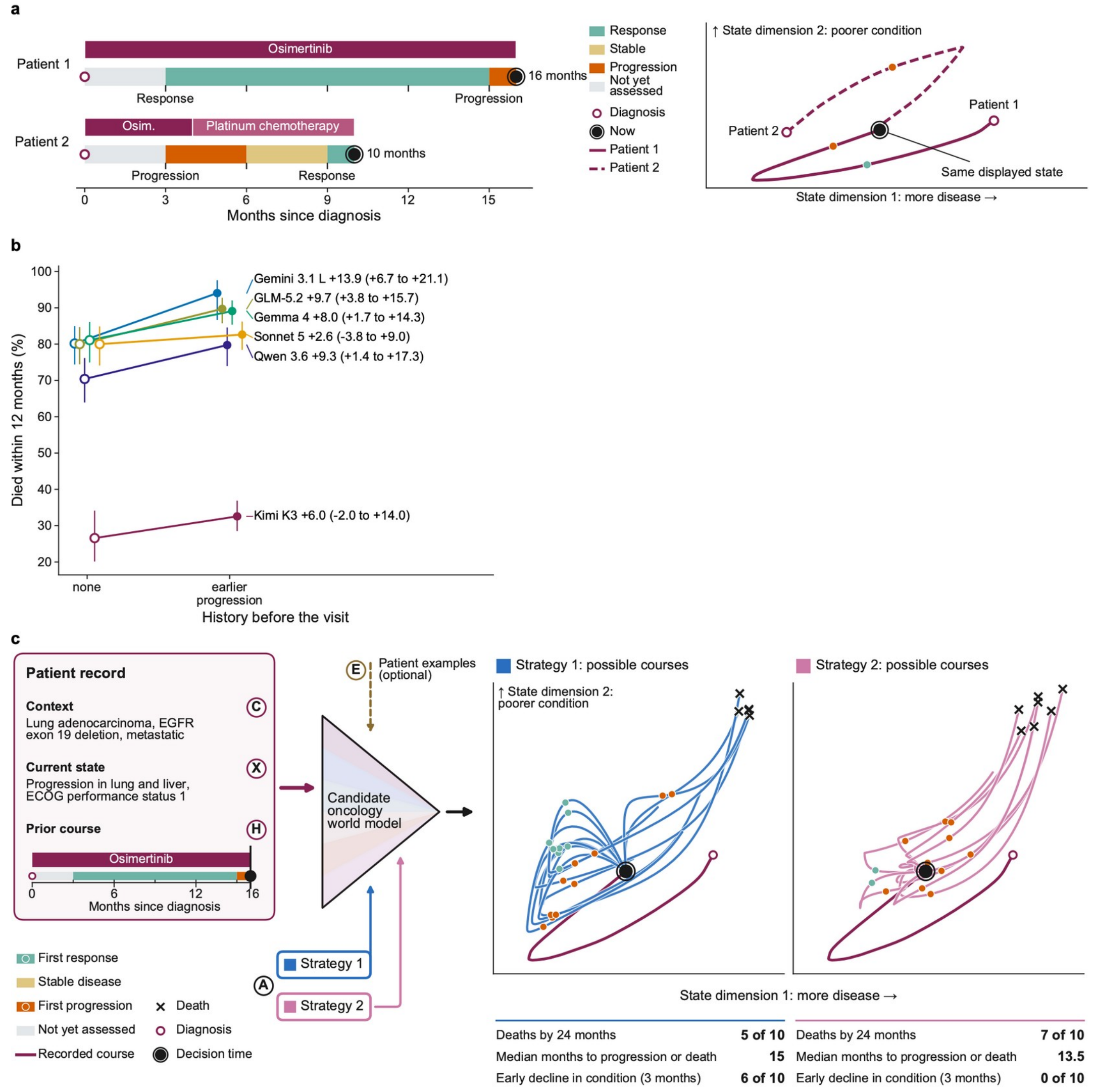

**a**, Two constructed patients with metastatic EGFR-mutant lung adenocarcinoma, as timelines and as paths: patient 1 responded to osimertinib and progressed at month 15; patient 2 progressed early and then responded to platinum chemotherapy. Both reach the same state (black ring) with different histories. Axes, disease burden and physical condition. **b**, Share of generated patients who died within 12 months after a visit at similar tumor burden and condition, without (open) and with (filled) progression before the visit, by model (158-250 and 84-519 visits per model). Bars, Wilson 95% intervals; labels, difference in percentage points with its 95% interval. **c**, A candidate world model forecasting courses: the record of constructed patient 1 (a) at month 16, with its context (C), current state (X) and history (H), one treatment strategy (A) and optional patient examples (E), enters the model. Right, 10 constructed courses per strategy over 24 months (crosses, deaths), on axes of disease burden and physical condition. Table, deaths by 24 months, median months to progression or death, and courses with an early decline in condition at 3 months: strategy 1 delays progression and death (5 against 7 deaths of 10 by 24 months; median 15 against 13.5 months) but worsens physical condition early (6 against 0 of 10 at 3 months). In a and c, constructed courses, not generated or observed patients.

## Extended Data Figure 3 | Outcome by starting position and clinical course of four generated cancer families

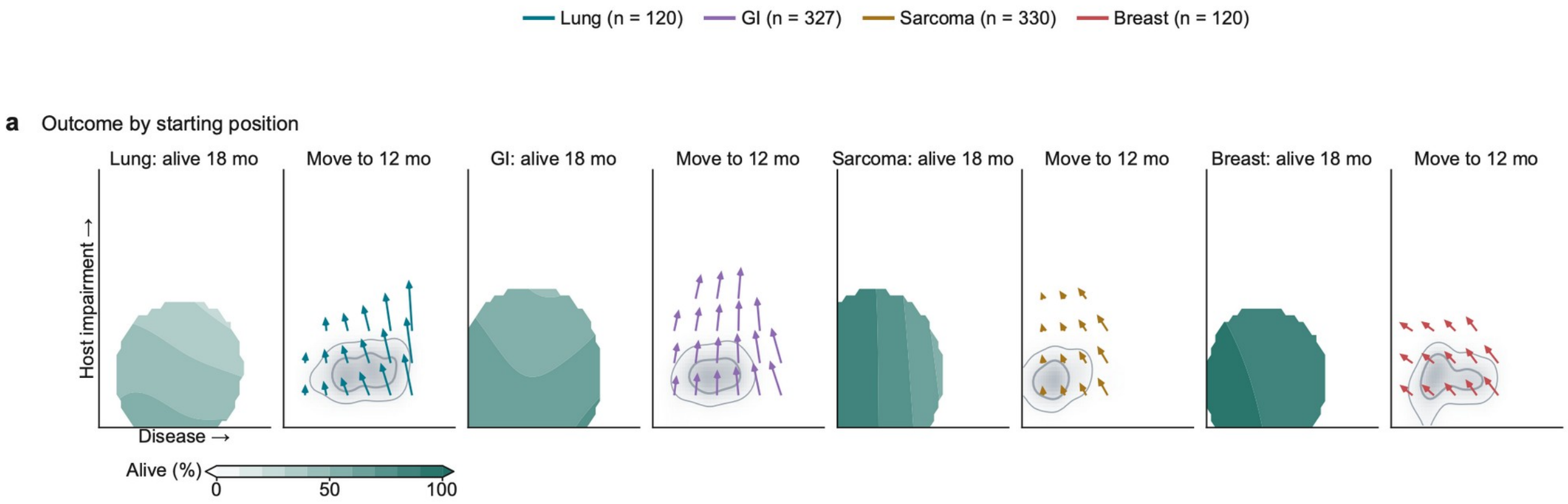

**b** Clinical course

| | Lung | | | | GI | | | | Sarcoma | | | | Breast | | | |
|---|---|---|---|---|---|---|---|---|---|---|---|---|---|---|---|---|
| Months | 0 | 12 | 24 | 36 | 0 | 12 | 24 | 36 | 0 | 12 | 24 | 36 | 0 | 12 | 24 | 36 |
| Alive | 100 | 82 | 20 | 4 | 100 | 92 | 30 | 10 | 100 | 95 | 50 | 19 | 100 | 99 | 68 | 35 |
| ECOG ≥2 | 0 | 27 | 50 | 60 | 0 | 18 | 32 | 39 | 0 | 13 | 33 | 48 | 0 | 8 | 26 | 19 |
| Grade ≥3 AE | 0 | 14 | 4 | 0 | 0 | 15 | 16 | 9 | 0 | 8 | 13 | 11 | 0 | 8 | 14 | 14 |
| Liver | 52 | 57 | 96 | 100 | 80 | 85 | 97 | 97 | 3 | 11 | 33 | 37 | 41 | 45 | 73 | 86 |
| Thoracic | 100 | 99 | 100 | 100 | 9 | 21 | 39 | 33 | 24 | 42 | 66 | 79 | 4 | 5 | 25 | 31 |
| Bone | 35 | 47 | 62 | 80 | 1 | 9 | 20 | 33 | 36 | 15 | 28 | 44 | 42 | 46 | 65 | 74 |
| CNS | 3 | 23 | 29 | 40 | 0 | 1 | 0 | 0 | 0 | 1 | 2 | 3 | 0 | 4 | 12 | 12 |

Generated patients of six models pooled, from diagnosis (lung 120, gastrointestinal 327, sarcoma 330, breast 120); their Kaplan-Meier survival and positions over time are in Fig. 2c. **a**, Survival at 18 months by position at diagnosis (color; kernel-weighted Kaplan-Meier estimate at each starting position, so color bands are smoothed estimates, not separate groups; blank, too few patients) and mean move to month 12 of survivors (arrows). **b**, Clinical course: percentage of patients alive and, among living patients, with ECOG performance status 2 or more, a grade 3 or higher adverse event, or liver, thoracic, bone or central nervous system involvement, at 0, 12, 24 and 36 months.

# Extended Data Figure 4 | Clinical scenarios, the factorial experiment, and where the EGFR treatment difference was spent

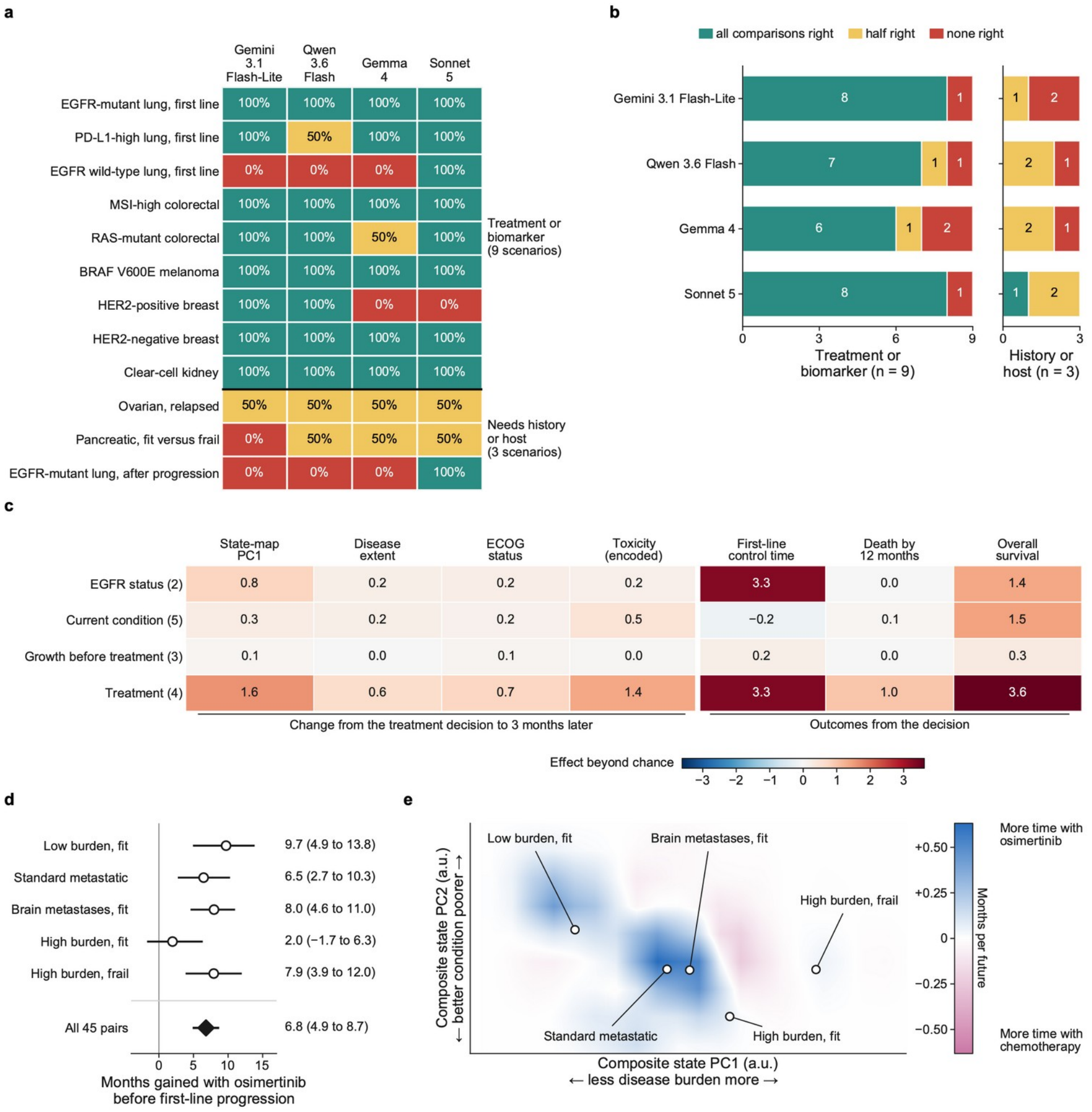

**a**, Share of treatment comparisons ranked correctly by Sonnet 5, Gemini 3.1 Flash-Lite, Gemma 4 and Qwen 3.6 Flash in 12 constructed scenarios. The first nine depend on a biomarker or treatment class; the last three also on previous treatment or physical condition. **b**, Number of scenarios in which all, half or none of the comparisons were correct, per model. **c**, Factorial experiment (Claude Sonnet 5; 360 futures from 120 combinations of EGFR status, 2 levels; current condition, 5; growth before treatment, 3: decision at diagnosis, or untreated progression 1 or 4 months after diagnosis; and treatment, 4). Each cell shows how much changing one input (row) changes one outcome (column), the other inputs being fixed, relative to the variation between futures of the same patient; 0, no effect beyond chance (400 shuffles of the input labels). Left, changes in the 3 months after the decision; right, outcomes from the decision. For growth before treatment, the change along state-map PC1 was not larger than with shuffled labels ($P = 0.28$). **d**, Restricted mean time before first-line progression over 24 months, osimertinib minus chemotherapy (Sonnet 5, EGFR-mutant), for all 45 pairs (diamond) and per starting state (circles, 9 pairs). Each pair joins the k-th futures of the two arms from the same starting record, an arbitrary pairing. Lines, 95% bootstrap intervals over pairs (5,000 resamples); resampling the 15 starting records gives 4.9 to 8.9 (Fig. 3a). **e**, For each region of the map, months per future spent there before first-line progression, osimertinib minus chemotherapy: blue, osimertinib futures spent more months in that region (they stayed longer without progression, mostly near their starting states); pink, chemotherapy futures spent more months there. The colors show where the progression-free time was spent, not which treatment is better in a region. Circles, starting states. Regions add to +12.9 and -6.1 months (net +6.8).

## Extended Data Figure 5 | Survival forecasts: bias, calibration, coverage and ranking

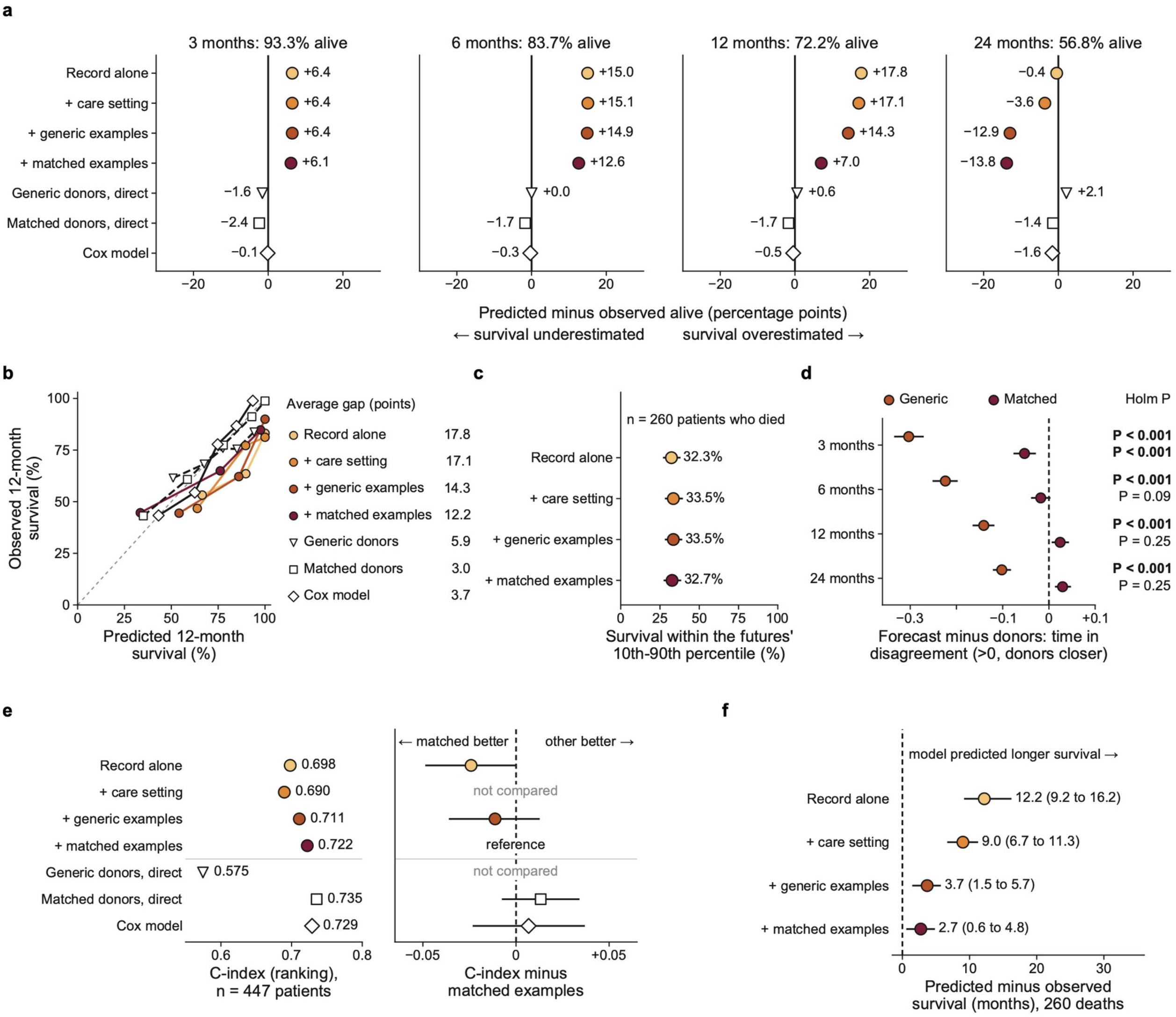

**a**, Mean predicted probability of being alive minus observed survival (censoring-weighted) at 3, 6, 12 and 24 months (449 patients; 447 at 24 months); right of the line, survival overestimated. Symbols as in Fig. 4d. **b**, Calibration of 12-month survival: mean predicted against observed survival in groups of predicted probability; key, average gap between predicted and observed survival across groups (expected calibration error, percentage points). **c**, Among the 260 patients who died, share whose survival fell between the 10th and 90th percentiles of their generated futures. **d**, Time-aligned disagreement of the forecast minus that of the example patients used directly (positive, example patients closer), at four horizons (381-437 patients); P, Wilcoxon signed-rank test, Holm-adjusted over eight comparisons. **e**, Harrell's C-index of predicted restricted mean survival to 24 months (447 patients); right, comparator minus matched examples. **f**, Predicted minus observed survival among the 260 patients who died (mean of each patient's futures). Lines, 95% bootstrap intervals over patients. GENIE-BPC; DeepSeek V4 Flash.

# Extended Data Figure 6 | GENIE-BPC colorectal cancer: observed and generated courses

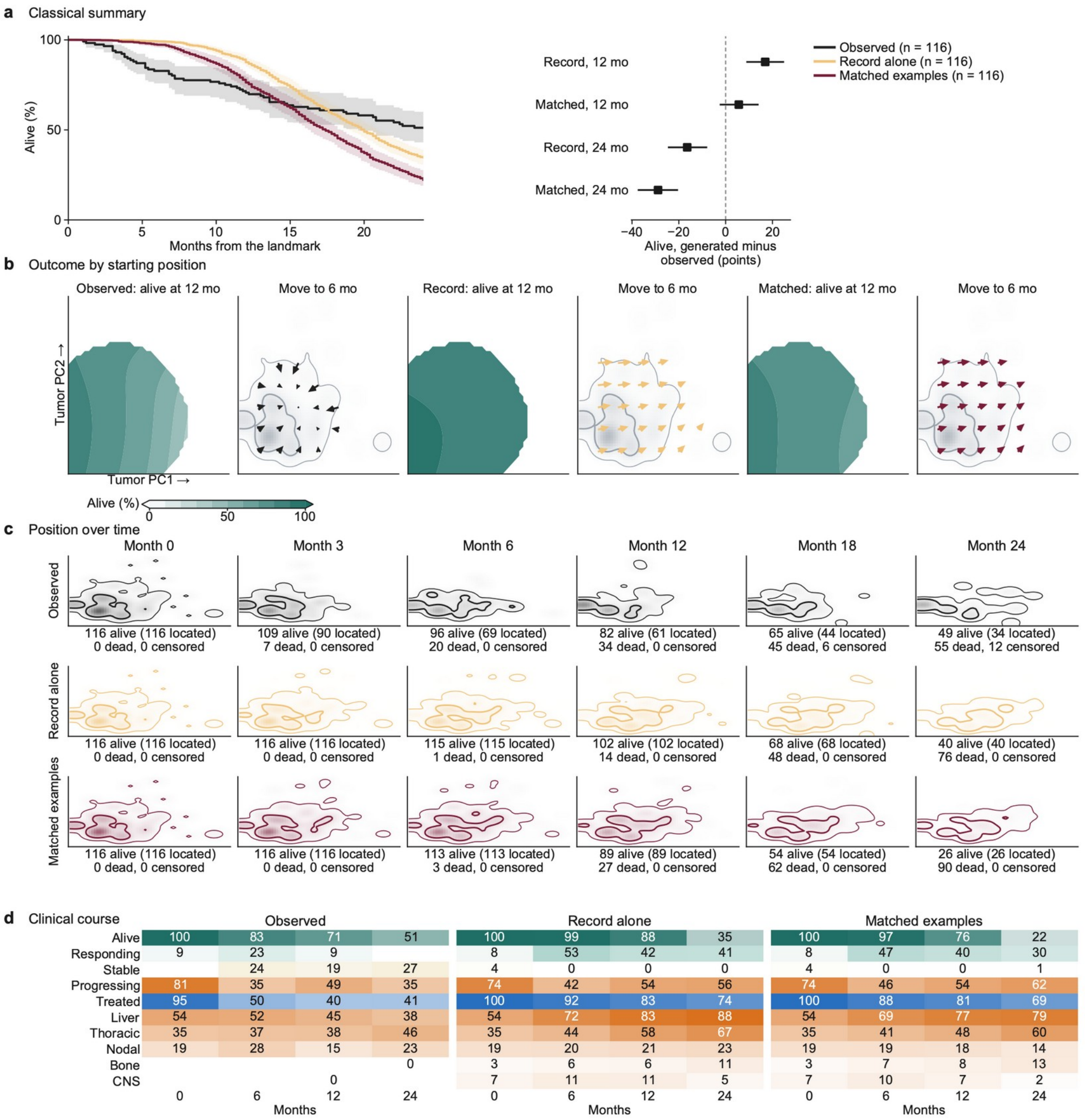

GENIE-BPC colorectal cancer, 116 patients located at the landmark: observed courses and futures generated by DeepSeek V4 Flash with the record alone and with matched examples, read on each patient's assessment dates. **a**, Kaplan-Meier survival from the landmark (generated curves pool each patient's futures with equal patient weight; bands, 95% intervals, 400 resamples of patients), and survival at 12 and 24 months, generated minus observed in percentage points (95% bootstrap intervals, 500 resamples of patients). **b**, Survival at 12 months by position at the landmark (blank where too few patients have a known status) and mean move to month 6 among survivors located then. **c**, Positions at successive months among patients located then; generated counts are patient equivalents (each future weighted by one over the number of the patient's futures). **d**, Clinical course: Kaplan-Meier survival and, among living patients with a known value, disease state, treatment and sites. Axes, first two components of the tumor variables, fitted on observed positions. In d, blank observed cells hide counts of 1 to 4 patients, or a second count that would reveal them by subtraction.

## Extended Data Figure 7 | Whole-course comparisons: disagreement, treatment and disease state

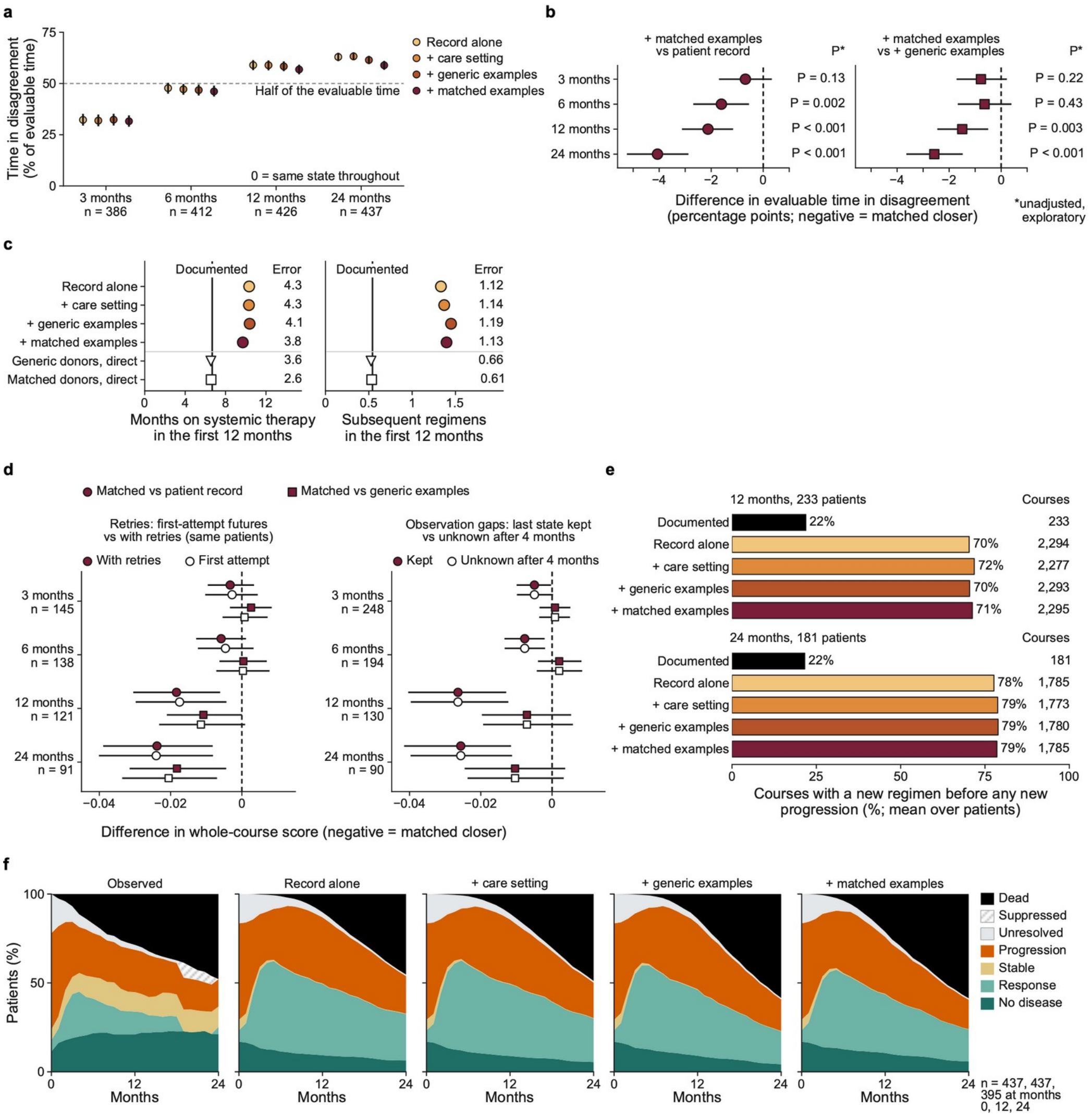

**a**, Share of evaluable time in which generated and recorded courses are in different states (death counted as a state), at four horizons (386, 412, 426 and 437 patients). **b**, Paired differences in a: matched examples minus record alone (circles) and minus generic examples (squares); P, two-sided Wilcoxon signed-rank test, unadjusted. **c**, Months on systemic therapy and number of subsequent regimens in the first 12 months (317 patients): mean forecast (points), documented mean (line) and mean absolute error per patient. **d**, Two robustness checks of the joint score, matched examples minus record alone (circles) and minus generic examples (squares). Left, retries: only the futures accepted at the first generation attempt, against all accepted futures including retries, on the same patients (145, 138, 121 and 91 patients). Right, observation gaps: documented states becoming unknown after four months, against the primary reading in which the last documented state persists until the next update (248, 194, 130 and 90 patients). Points, mean paired difference; P, two-sided Wilcoxon signed-rank test, unadjusted. **e**, Share of courses in which a new regimen started before any new progression, documented and generated, at 12 months (233 patients) and 24 months (181 patients). **f**, Disease state month by month in the registry and in the futures of the four settings, among patients who had died or remained under follow-up at each month (437 patients; 395 at month 24); hatched as in Fig. 5a. Lines in a, b and d, 95% bootstrap intervals over patients. DeepSeek V4 Flash.

# Extended Data Figure 8 | PRIME: subgroups, hazard ratios and progression-free survival

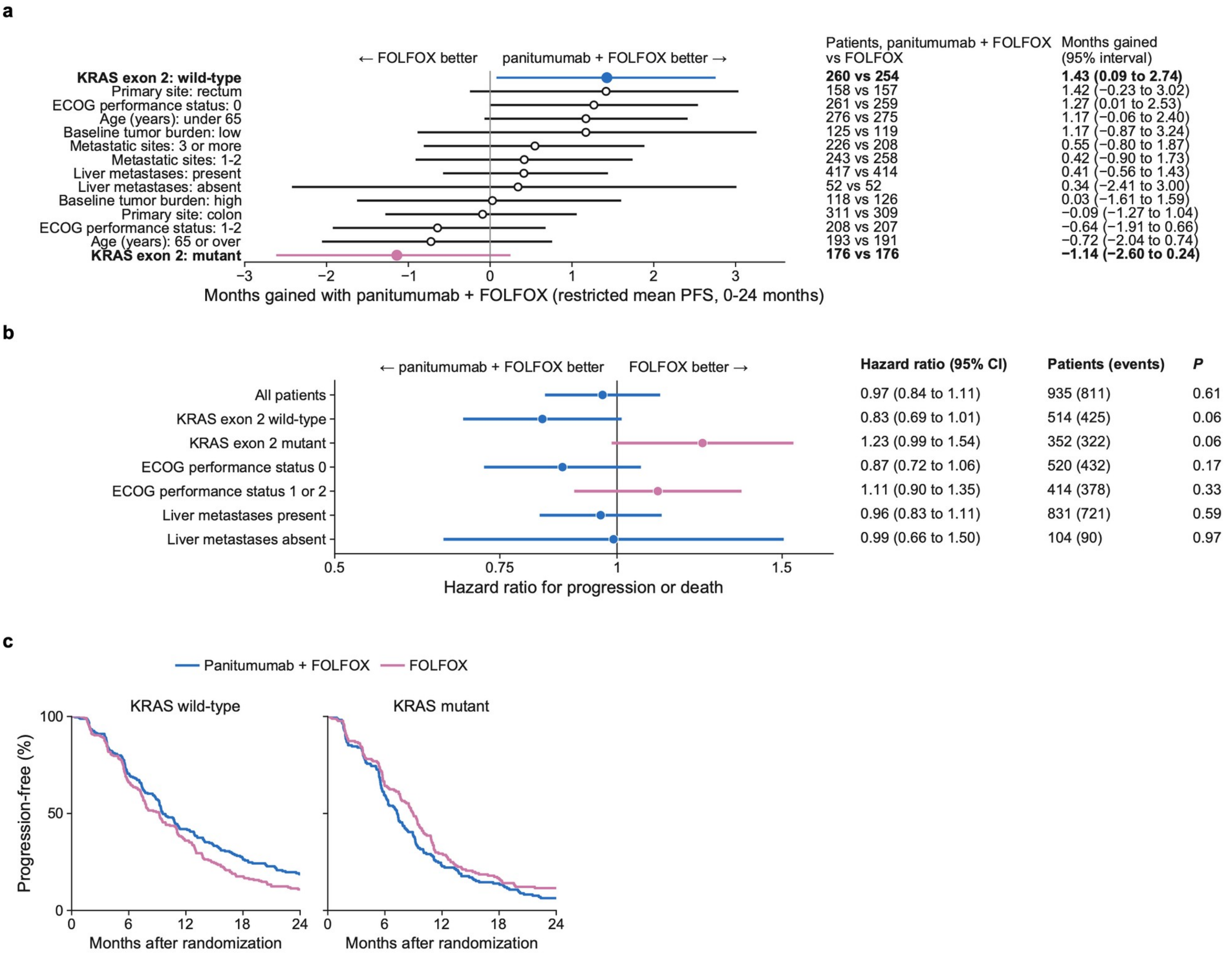

**a**, Difference in restricted mean progression-free survival over 24 months (central review) in each predefined subgroup. The patient with missing ECOG performance status is counted with ECOG 1-2. **b**, Hazard ratios for progression-free survival in all 935 randomized patients (469 and 466) and in each subgroup, from separate Cox models; P, Wald test. **c**, Kaplan-Meier progression-free survival by arm and KRAS exon 2 status (wild-type, 260 and 254 patients; mutant, 176 and 176). Lines, 95% bootstrap intervals (2,000 resamples within arm; a) and 95% confidence intervals (b).

# Extended Data Figure 9 | PRIME, KRAS-mutant disease in the Clinical State Space

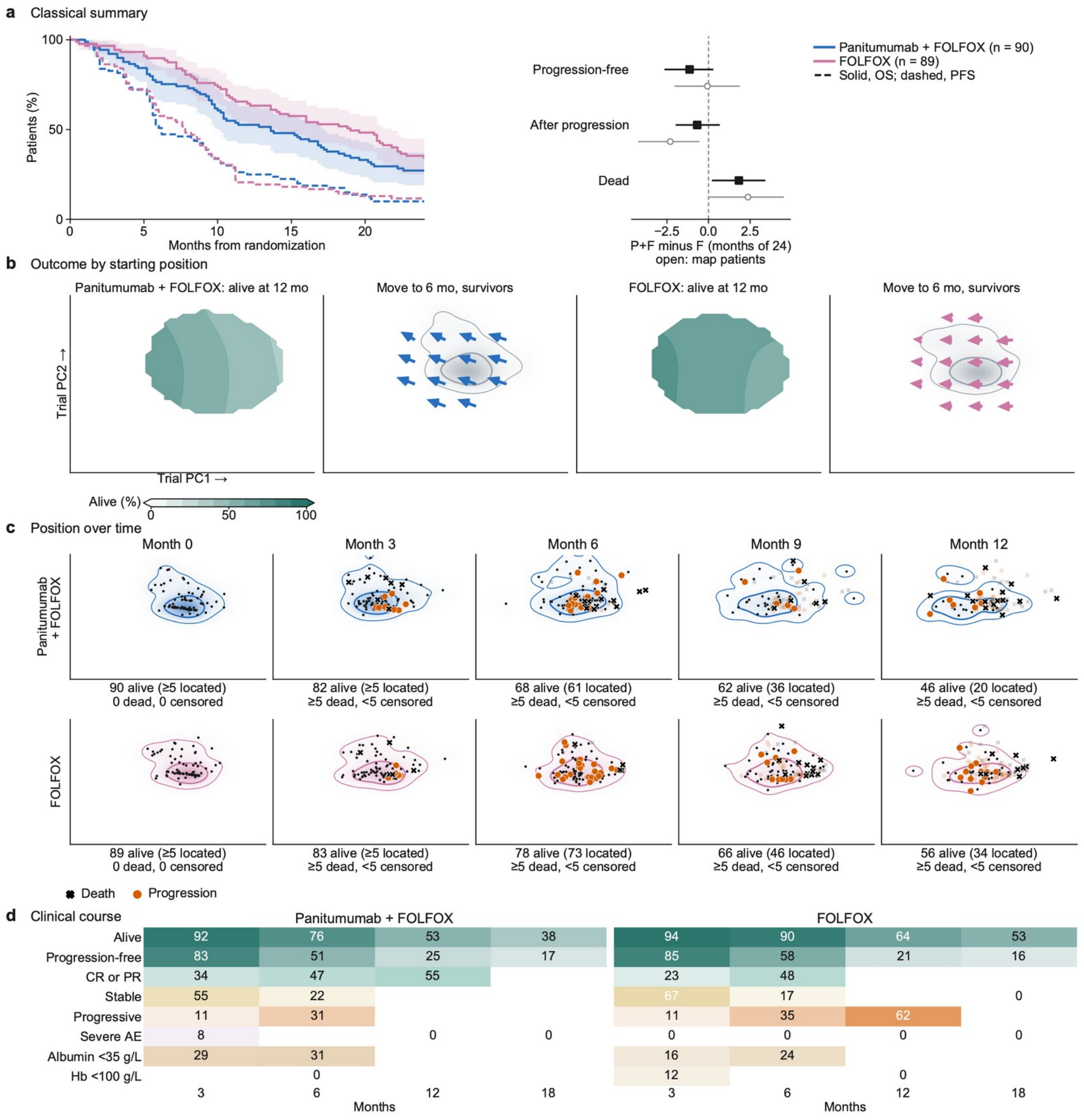

KRAS-mutant patients with a lesion assessment by day 14 (90 and 89), from randomization. **a**, Kaplan-Meier overall survival (solid) and progression-free survival (dashed) (bands, 95% intervals), and the first 24 months divided into time without progression, after progression and dead, panitumumab + FOLFOX minus FOLFOX (P+F minus F; filled, all randomized patients with known KRAS status; open, the mapped patients; curves, mapped patients). **b**, Survival at 12 months by starting position (blank, too few patients) and mean move to month 6 of survivors. **c**, Positions over time, with progressions and deaths at the last assessment. **d**, Clinical course. In c and d, counts of 1 to 4 patients are shown as <5 (and one other count as ≥5 where needed to prevent recovery by subtraction) and the matching cells are blank.

# Supplementary Information

Large language models as oncology world models

## Contents

## Supplementary Figure 1 | Six generated lung adenocarcinoma trajectories, one per model

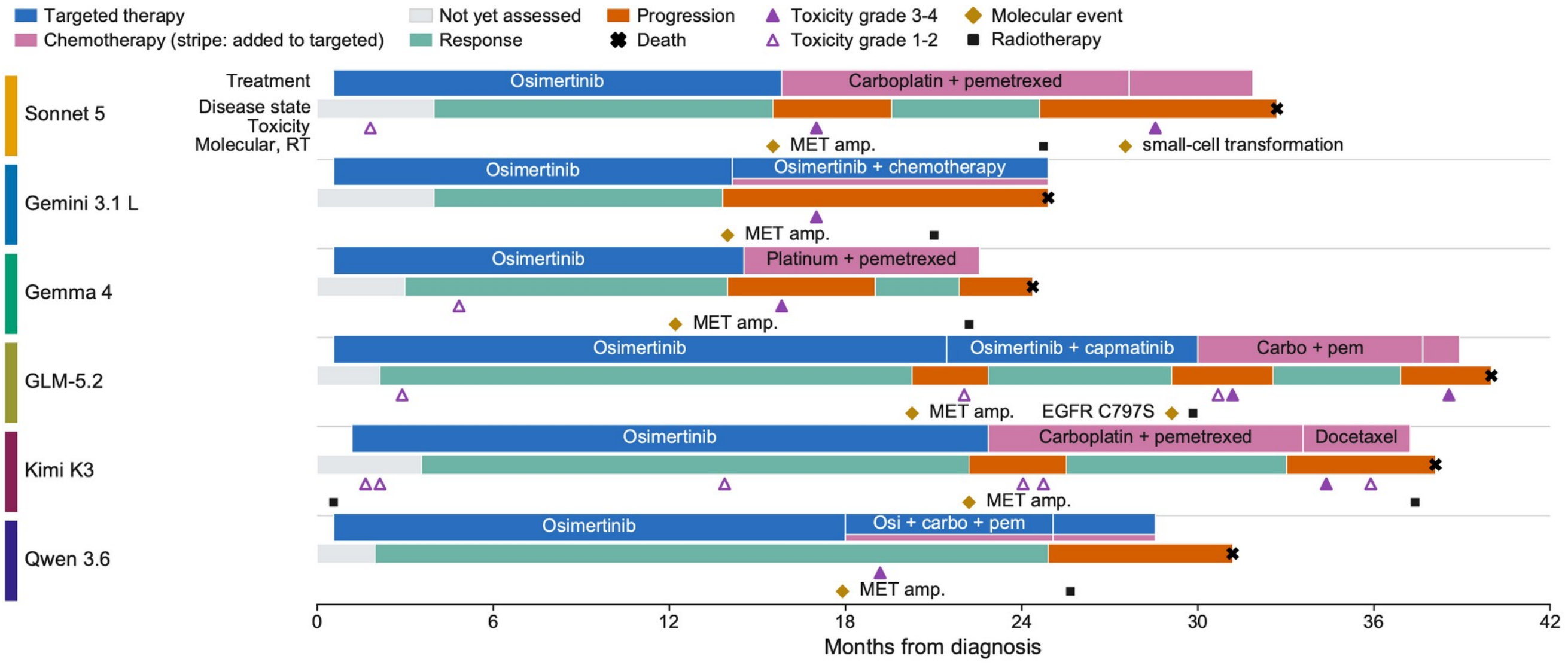

One generated lung adenocarcinoma per model (selection in Methods), against months from diagnosis. Tracks: treatment lines (blue, targeted therapy; pink, chemotherapy; striped, both) with regimen names; disease state (gray, not yet assessed; cross, death); adverse events (filled triangle, grade 3-4; open, grade 1-2); molecular changes (diamonds) and radiotherapy (squares). Color bar, model.

# Supplementary Figure 2 | One generated patient: record, summaries and path on the state map

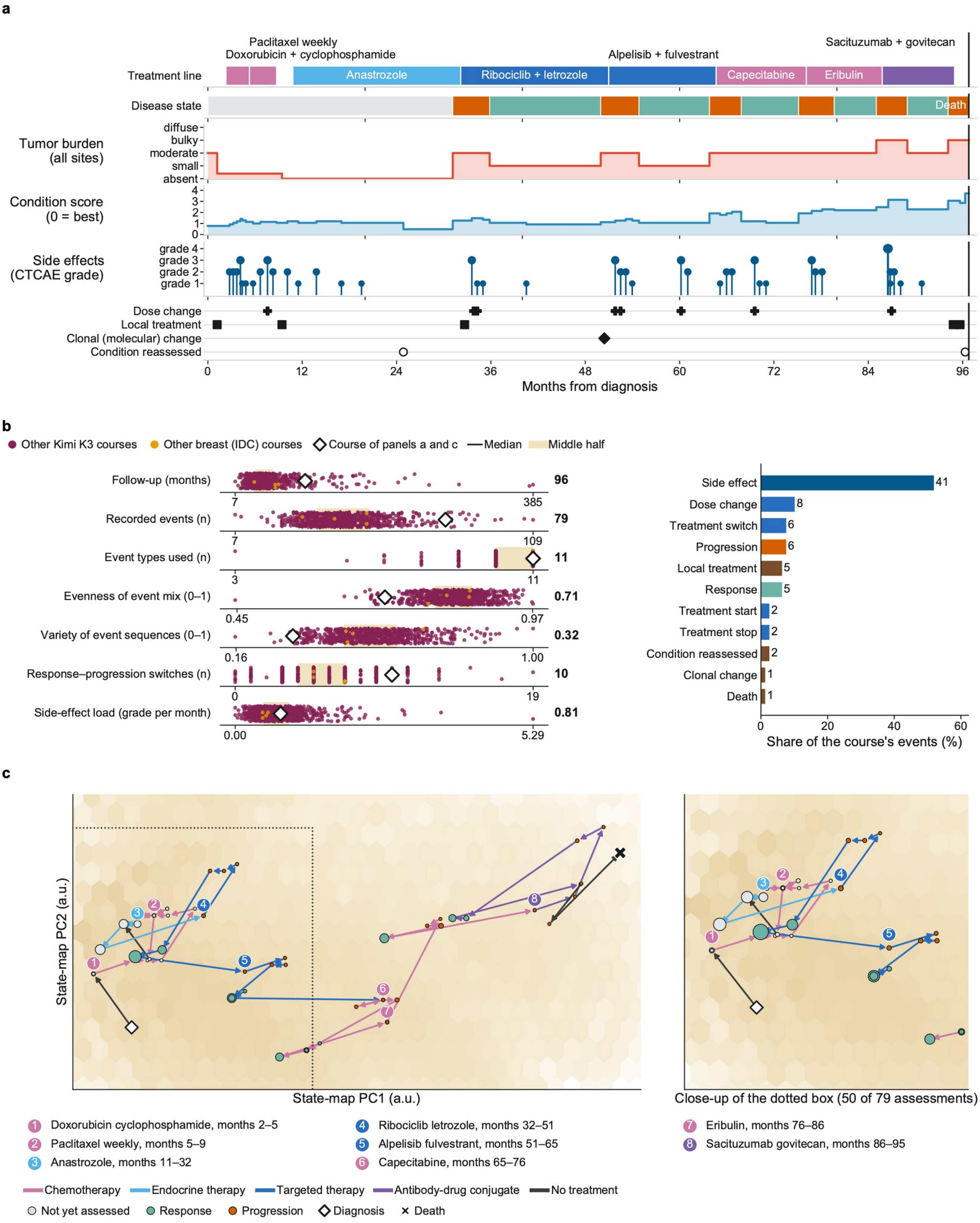

One generated invasive ductal carcinoma (Kimi K3; selection in Methods). **a**, The course as tracks: treatment line, disease state, overall tumor burden, condition score (a weighted composite of ECOG, symptoms, side effects and organ function), side effects (stem height, CTCAE grade) and events. **b**, Seven summaries of this course (diamond) among the 736 other courses of the same Kimi K3 run (dots; orange, other invasive ductal carcinomas); line, median; band, middle half. Right, its 79 events by type. **c**, The same course on the state map defined in Fig. 2 (Supplementary Fig. 4b,c): dots, generated assessments colored by disease state (area, time to the next); arrows colored by treatment class; numbered discs, start of each line; diamond, diagnosis; cross, death; right, close-up of the dotted box.

# Supplementary Figure 3 | Treatment sequences and adverse events in the generated atlas

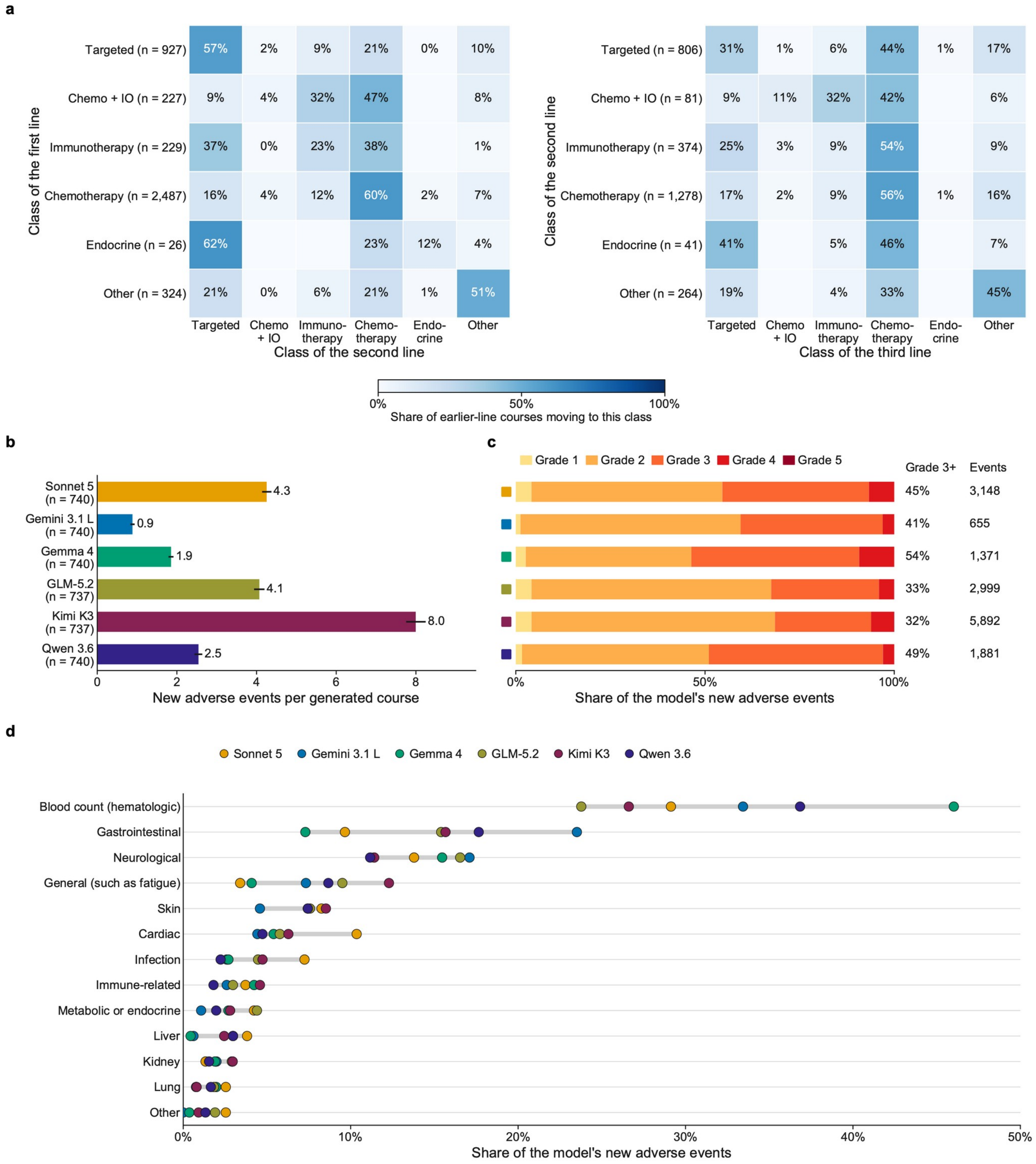

Courses generated by six models. **a**, Class of the next line given the class of the earlier line (left, first to second line, 4,220 transitions; right, second to third, 2,844); rows sum to 100%. **b**, New adverse events per course, by model; bars, mean; lines, 95% bootstrap intervals over courses (2,000 resamples). **c**, Grade at onset of these events (CTCAE v5); right, share of grade 3 or higher and number of events. **d**, Organ system of the events, as assigned by each model from the 13 categories of the output schema (not recoded or reviewed): dots, share of each model's events; gray bar, range across models (15,946 events).

## Supplementary Figure 4 | The clinical state map: layouts, loadings and event checks

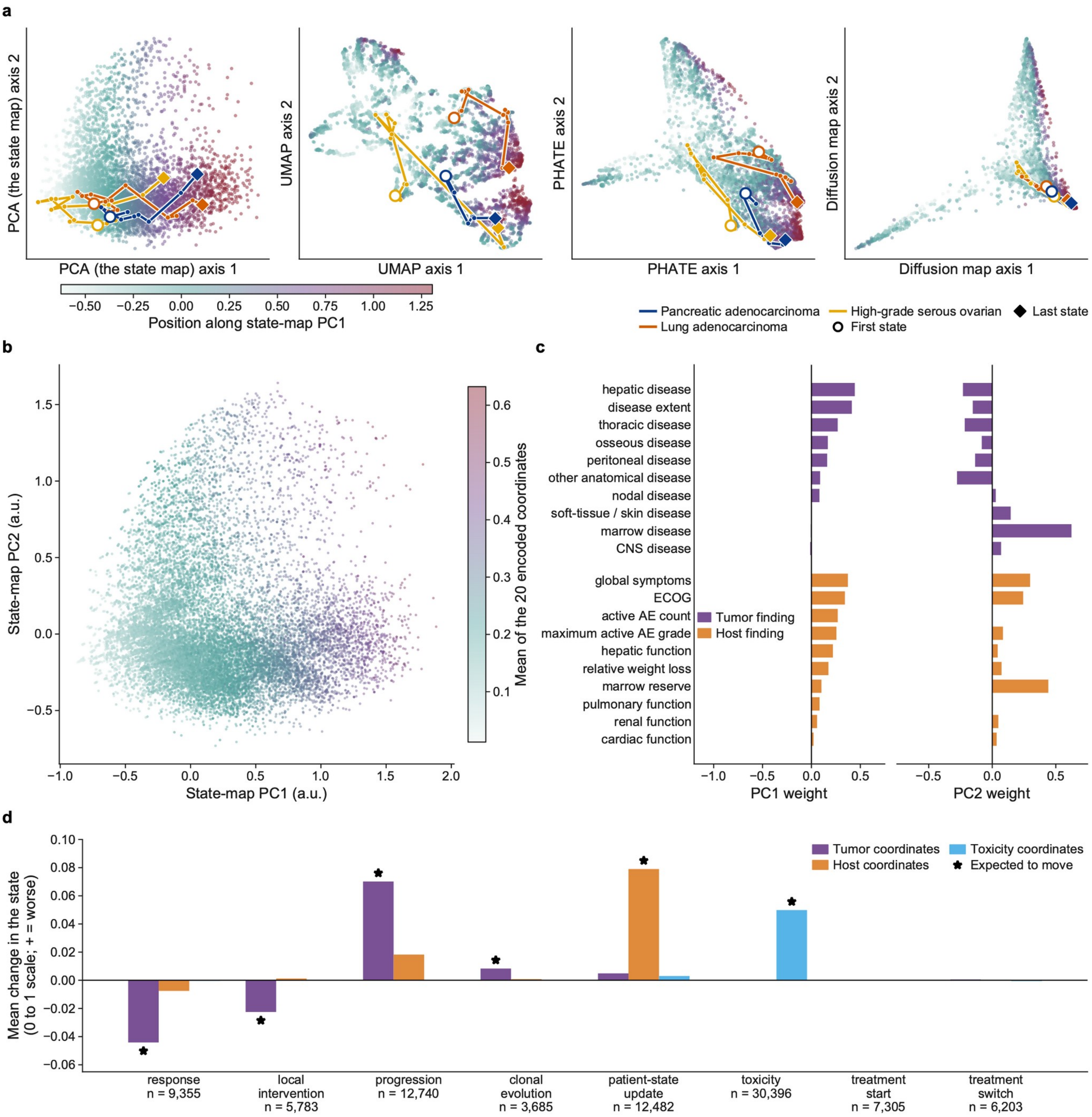

**a**, The same 6,000 generated states in four layouts: the linear map used in the paper (PCA), UMAP, PHATE and a diffusion map, colored by state-map PC1; lines, one trajectory for each of three cancers (open circle, first state; diamond, last). **b**, Generated living states (24,000 of 105,839) on the state map: first two principal components (25.6% and 13.9% of variance) of 10 tumor and 10 host variables; color, mean of the 20 encoded variables. **c**, Loadings of each variable on PC1 (left) and PC2 (right); purple, tumor; orange, host. **d**, Mean change of the tumor (purple), host (orange) and toxicity (blue) coordinates at each event type (n, events); stars, the coordinate expected to move. Treatment start and switch are negative controls. Generated atlas, 4,434 trajectories.

# Supplementary Figure 5 | Where the six models place their patients on the state map

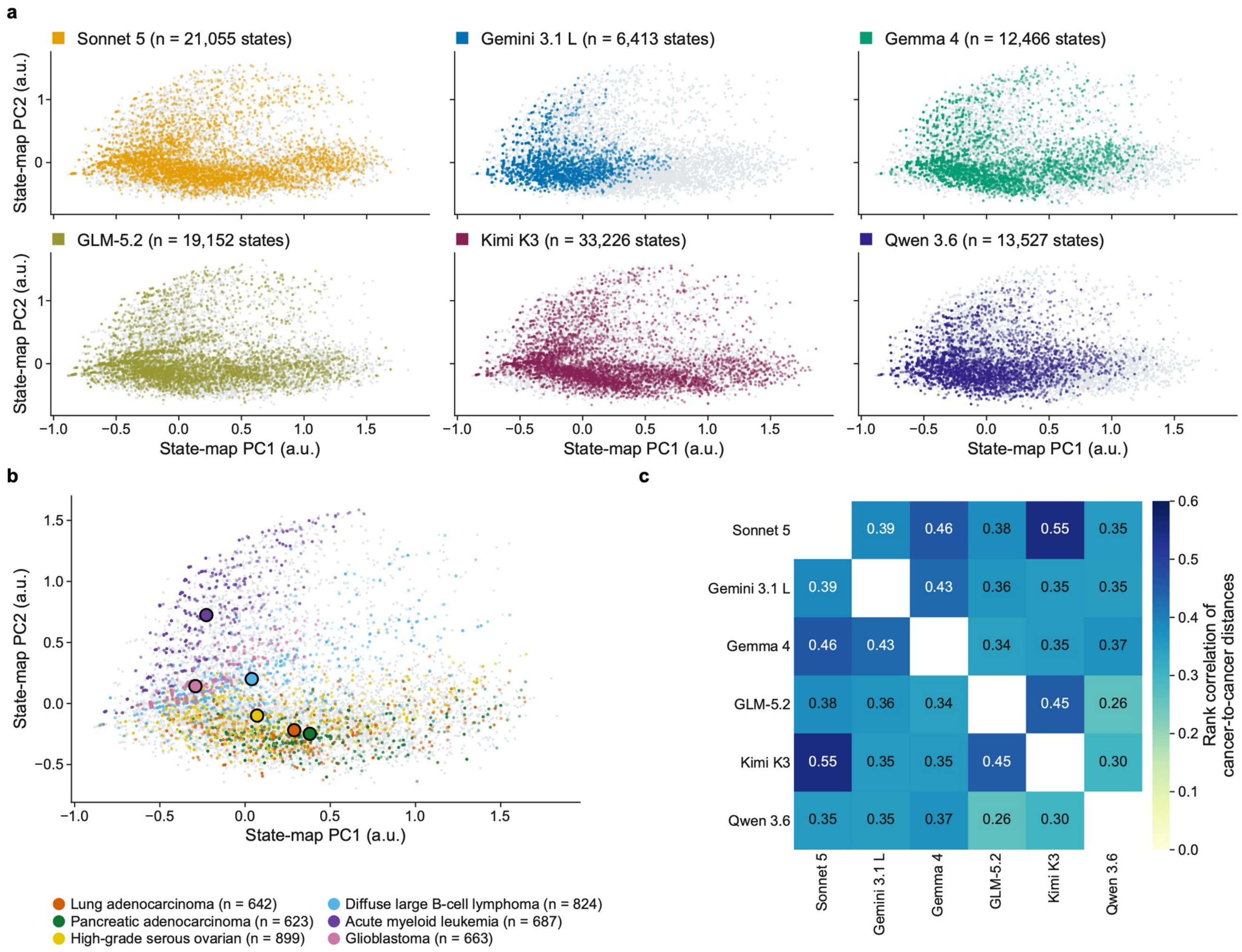

**a**, Living states of each model on the state map (5,000 per model; gray, 8,000 states of all models). **b**, States of six cancer types, all models pooled (n in the key); circles, mean position. **c**, Spearman correlation between models of their distances between 146 cancers.

## Supplementary Figure 6 | Event-conditioned landscapes: a conceptual model and generated trajectories

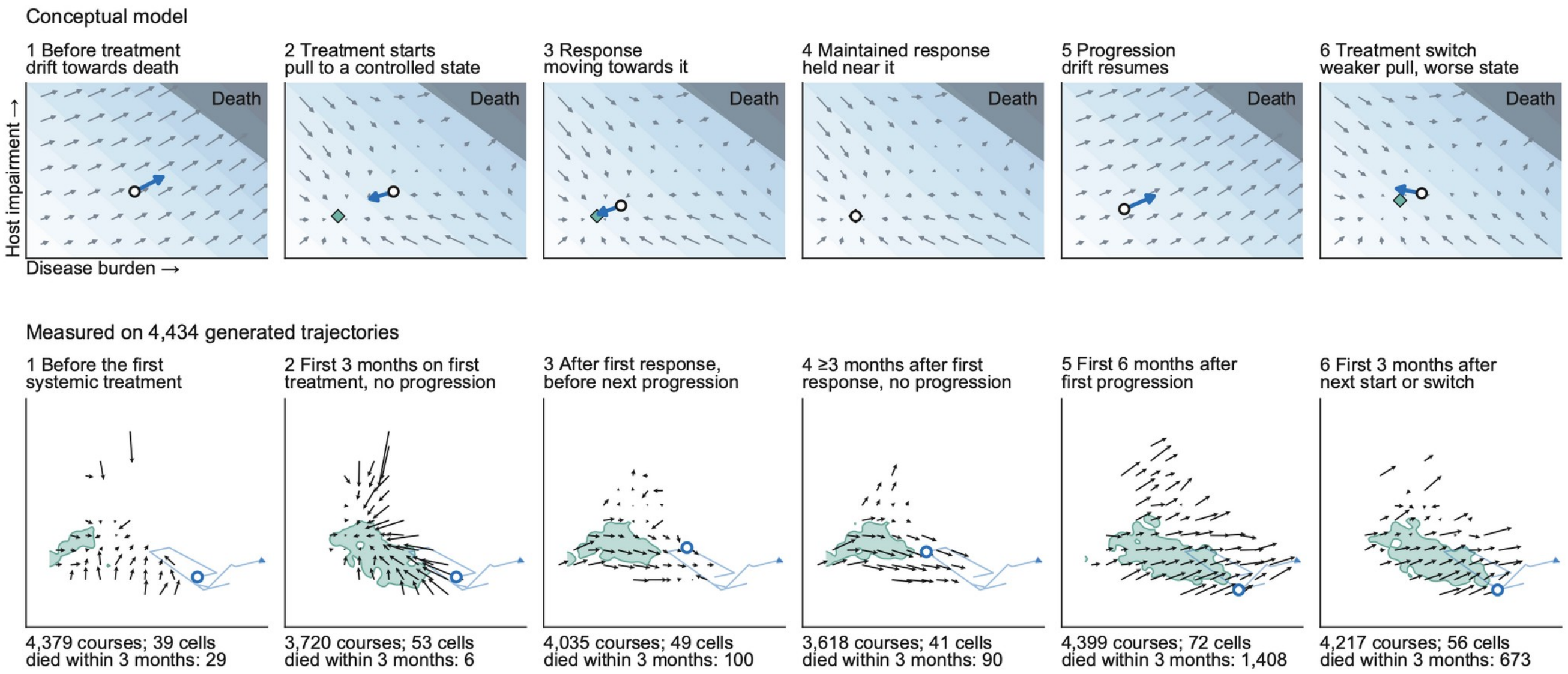

Top, a conceptual model drawn by hand, not estimated from data, at six moments of one course: the patient (open circle) on axes of disease burden and host impairment. Without active treatment, gray arrows drift towards more disease and poorer condition, faster as both increase, and end in death, an absorbing outcome (dark corner, drawn as a region for display; death is not a position on the map). An active treatment pulls the patient towards a controlled state (diamond) with less disease and better condition, but not cure; the pull weakens near death, and a later treatment pulls less, towards a worse state. Blue arrow, net movement of the patient. The controlled state is a conceptual hypothesis, not an attractor shown in data. Bottom, generated trajectories (4,434, six models) in six overlapping groups of visits that correspond to the moments above, each defined from a trajectory's own events (Methods); these are groups of visits, not successive snapshots of one population: black arrows, mean 3-month move of survivors in each cell with at least 30 courses; green, the 50% highest-density region of the phase; counts below, deaths within 3 months. Blue line, the patient of Fig. 1c.

## Supplementary Figure 7 | Vector fields of generated trajectories, and when averaging misleads

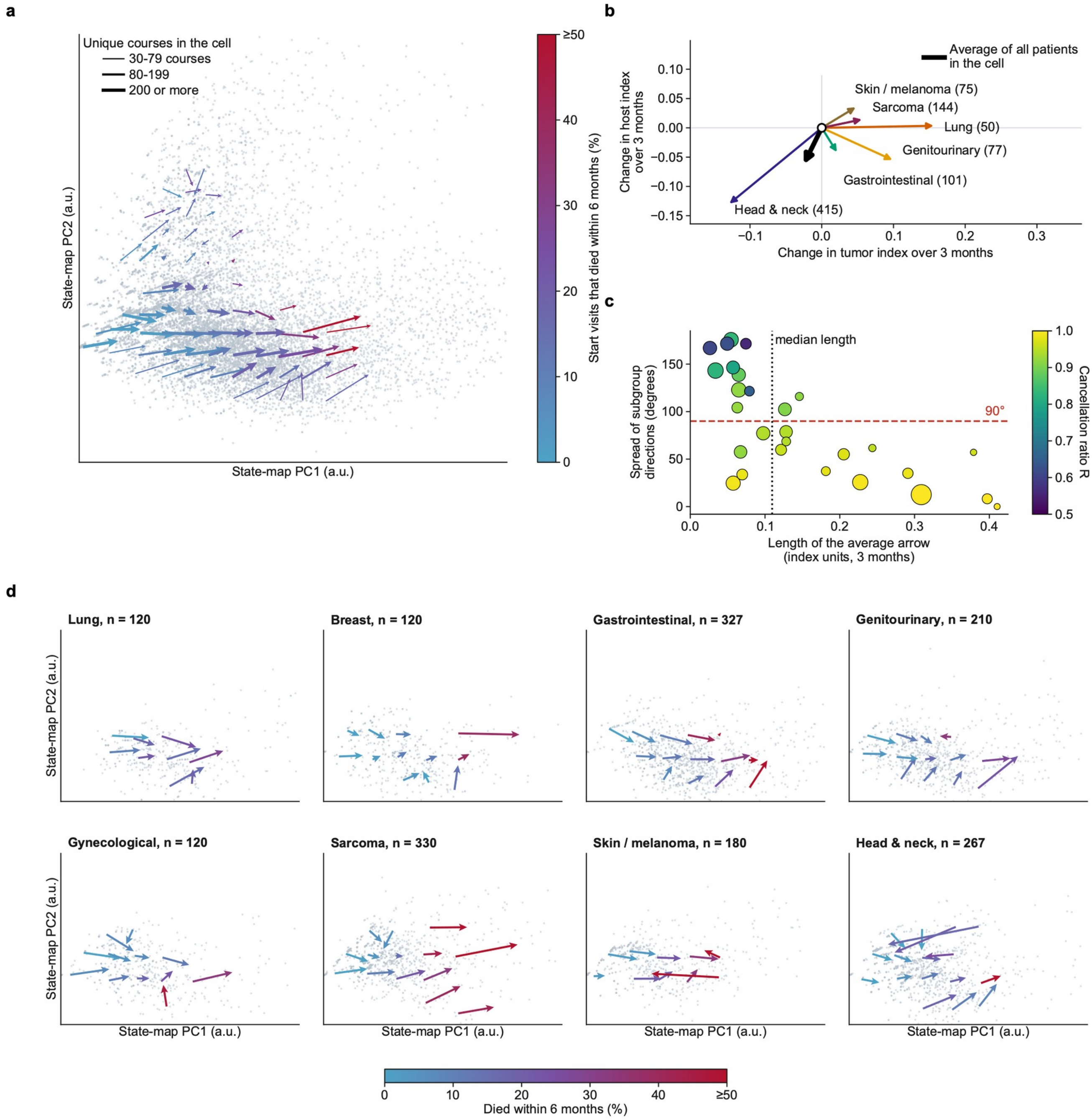

**a**, Mean 6-month move on the state map in the generated atlas (4,432 trajectories; 18,213 start visits at 3, 6, 12, 18 and 24 months, gray). Arrows, mean move of survivors in cells with at least 30 courses and 30 survivors (69 cells); thickness, number of courses; color, share dying within 6 months. Panels b and c use a tumor index (mean of eight standardized tumor features: overall, high-risk, maximum-site, liver and visceral burden, number of active sites, active clones and acquired alterations) and a host index (mean of standardized ECOG performance status, symptoms and worst organ function), over 3 months on a 7 × 7 grid, instead of the state map of a and d. **b**, One cell (896 visits) in which tumor families move in different directions: colored arrows, mean 3-month move per family (visits in brackets); black arrow, their weighted mean, 44% shorter than their mean length (cancellation ratio R = 0.56; directions span 171 degrees). **c**, Cells with at least 80 visits (n = 28): length of the pooled arrow against the angle spanned by subgroup arrows; color, R; size, visits. **d**, Fields as in a for eight tumor families.

## Supplementary Figure 8 | Reading an arrow of the state map, and how events change the next move

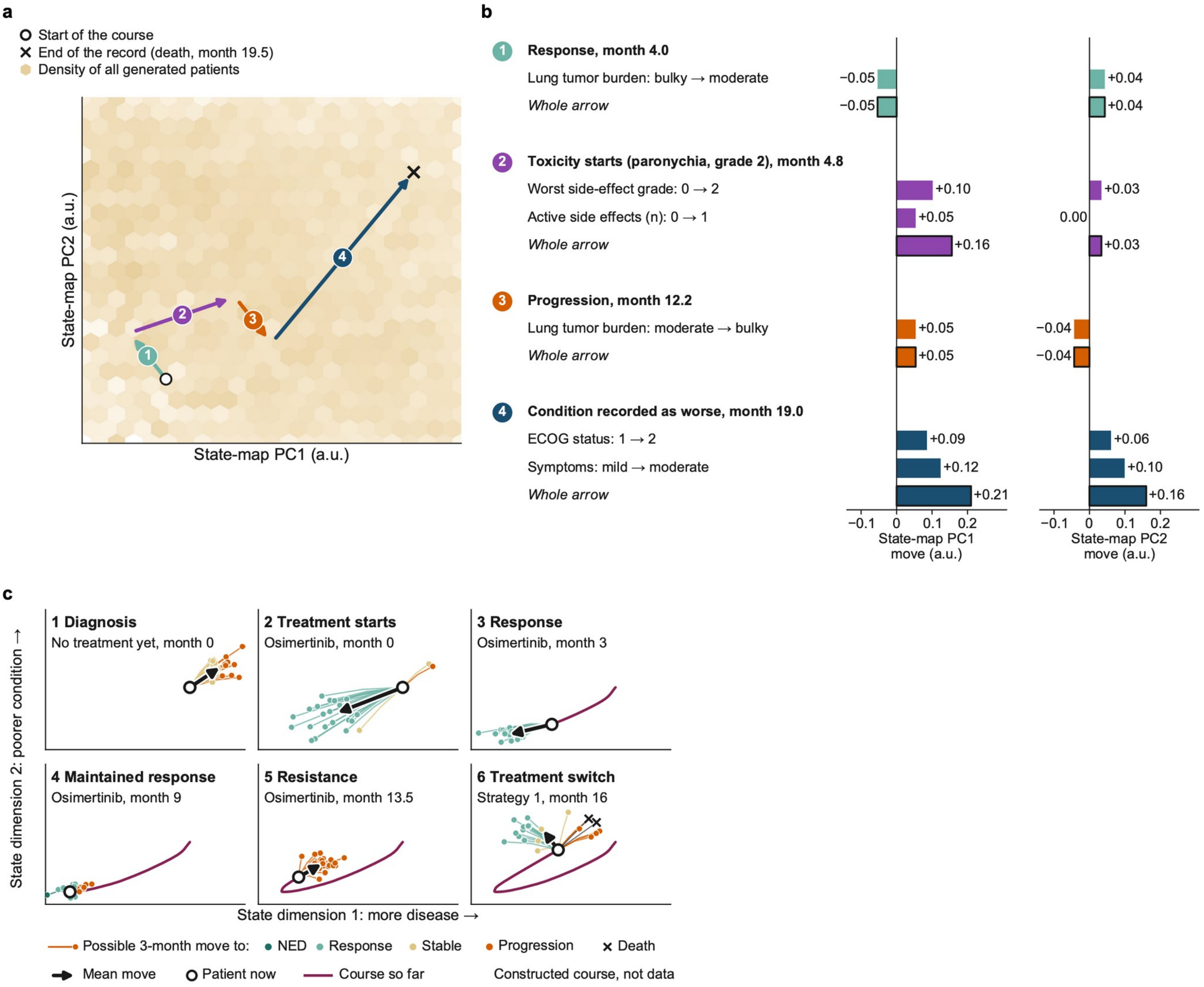

**a**, One generated lung adenocarcinoma (Gemini 3.1 Flash-Lite; selection in Methods) as arrows on the state map: each arrow, one event that changed the encoded state, numbered as in b and colored by event type; circle, start; cross, death. **b**, Clinical changes behind each arrow (value before → after) and their contribution to the move along PC1 (left) and PC2 (right); outlined bar, the whole arrow. **c**, Six moments of the constructed course of patient 1 of Extended Data Fig. 2a (a constructed illustration, not generated or observed data): thin lines, 24 possible 3-month moves from the current state (open circle), ends colored by disease state at the next assessment; crosses, deaths within 3 months; black arrow, mean move of survivors.

## Supplementary Figure 9 | Four scored scenario comparisons from the treatment decision

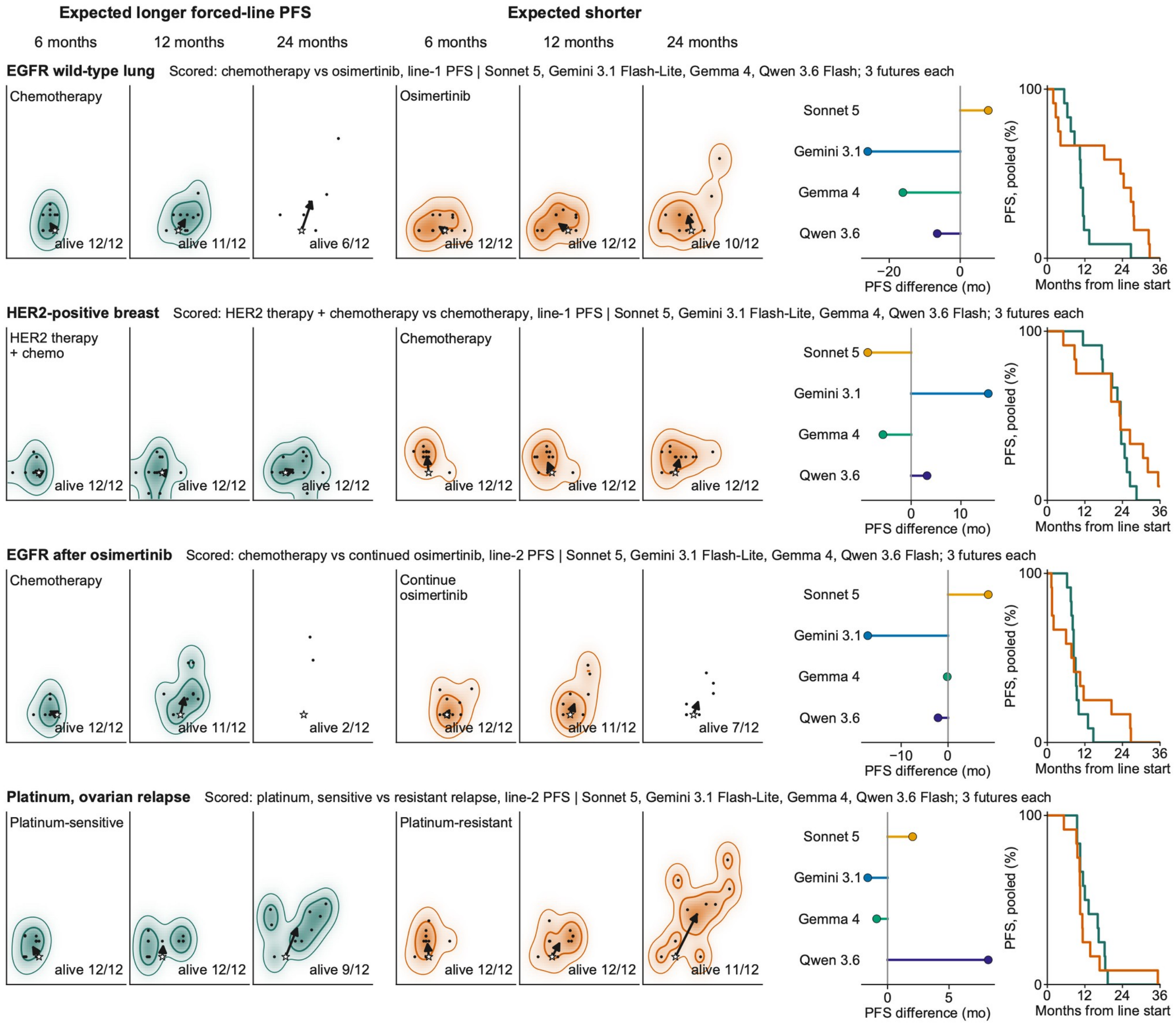

Four scored comparisons of the scenario benchmark, four models pooled (Sonnet 5, Gemini 3.1 Flash-Lite, Gemma 4, Qwen 3.6 Flash). Left and right, the two situations of each comparison on the disease and host axes (shading and contours, density when at least 8 futures are alive; points, futures; star, mean state at the decision; arrow, mean move). Each row names its scored comparison (3 futures per model and situation, 12 per situation). Center right, median progression-free survival (PFS) of the forced line in the situation expected to do better minus the other, per model; this is the scored endpoint. Right, PFS of the forced line in the two situations, four models pooled, from the start of the line to the first progression or death (line 1 in first-line scenarios, line 2 when the supplied history includes line 1); every future ended in death, so none was censored. Pooled median PFS, situation expected to do better against the other: 10.6 and 23.9 months (EGFR wild-type lung, chemotherapy against osimertinib), 23.4 and 23.2 (HER2-positive breast), 8.6 and 8.0 (EGFR after osimertinib) and 12.6 and 10.4 (platinum-sensitive against platinum-resistant ovarian relapse). No P value is given for pooled curves, because futures are grouped within model.

## Supplementary Figure 10 | GENIE-BPC lung cancer: observed and generated courses, all three groups

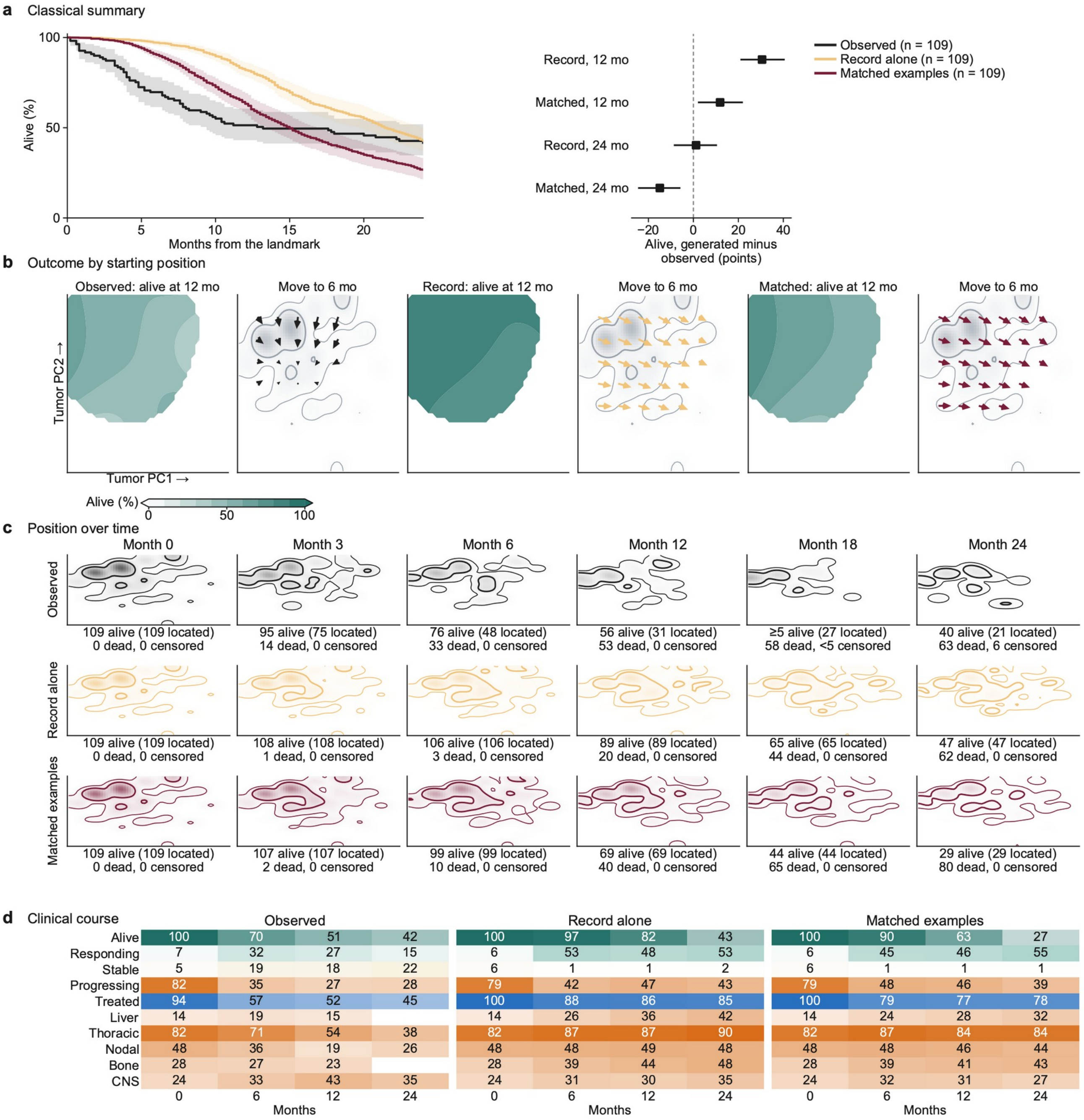

GENIE-BPC lung cancer, 109 patients located at the landmark; panels as in Extended Data Fig. 6 (survival and its differences, outcome by starting position, positions over time and clinical course), with frames for all three groups and more months. Observed counts of 1 to 4 are shown as <5 (and one other count as ≥5 where needed to prevent recovery by subtraction), and observed shares are blank when their count or its complement is 1 to 4.

## Supplementary Figure 11 | One real patient with generated futures, and the whole-course score by block and by cancer

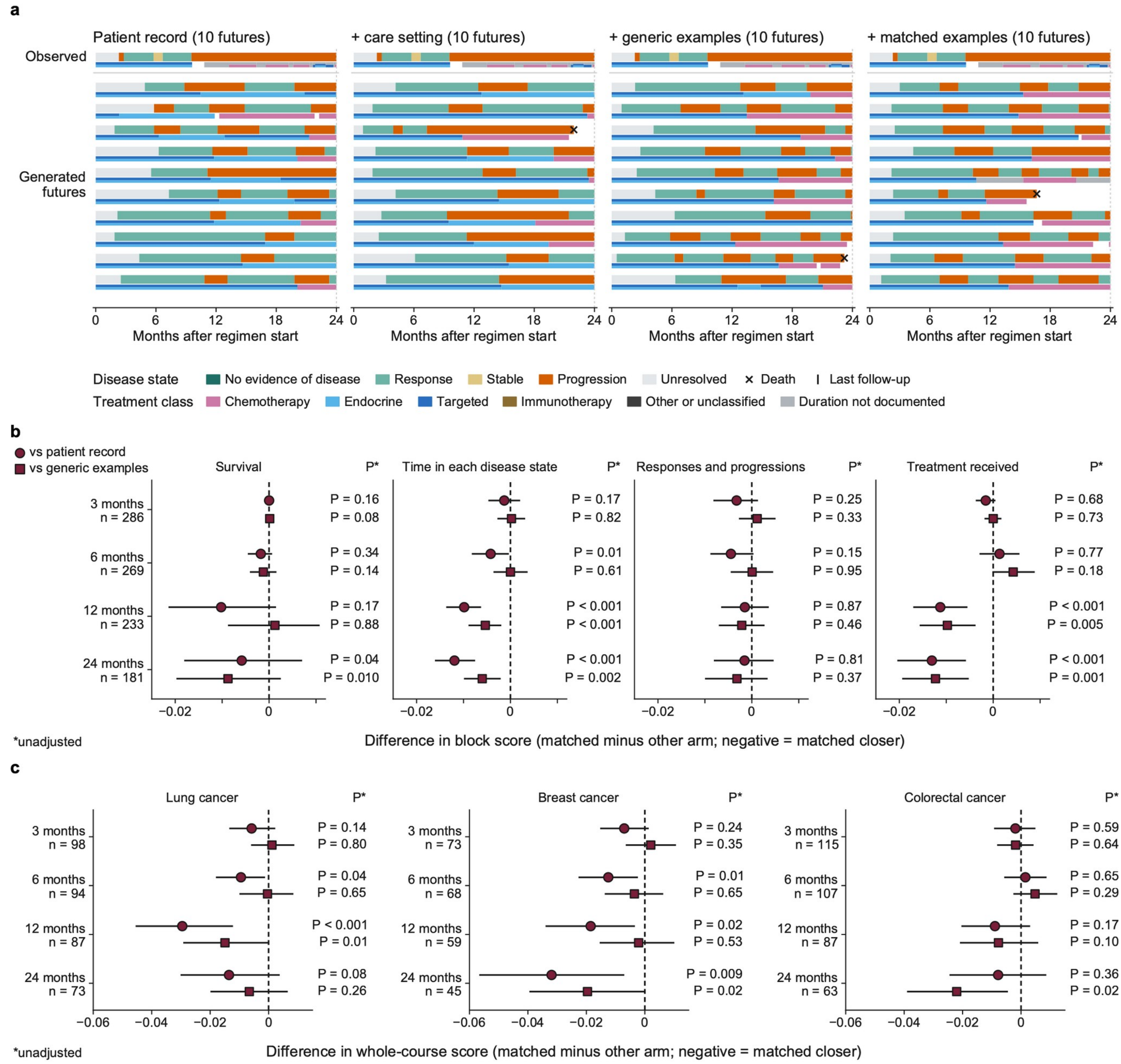

**a**, One observed patient (breast cancer, second regimen, letrozole + palbociclib; selection in Methods) and the futures generated in the four settings: top, observed course; below, every accepted future. Thick bars, disease state; thin strips, systemic treatment class (gray, duration not documented); ×, death; |, last follow-up. **b**, Whole-course score of Fig. 5d by block (survival, time in each disease state, new responses and progressions, treatment): matched examples minus record alone (circles) and minus generic examples (squares). **c**, Whole-course score by cancer, same contrasts. In b and c, points, mean paired difference; lines, 95% bootstrap intervals over patients; P, two-sided Wilcoxon signed-rank test, unadjusted. DeepSeek V4 Flash.

## Supplementary Figure 12 | A lung cancer case: one observed patient and generated futures

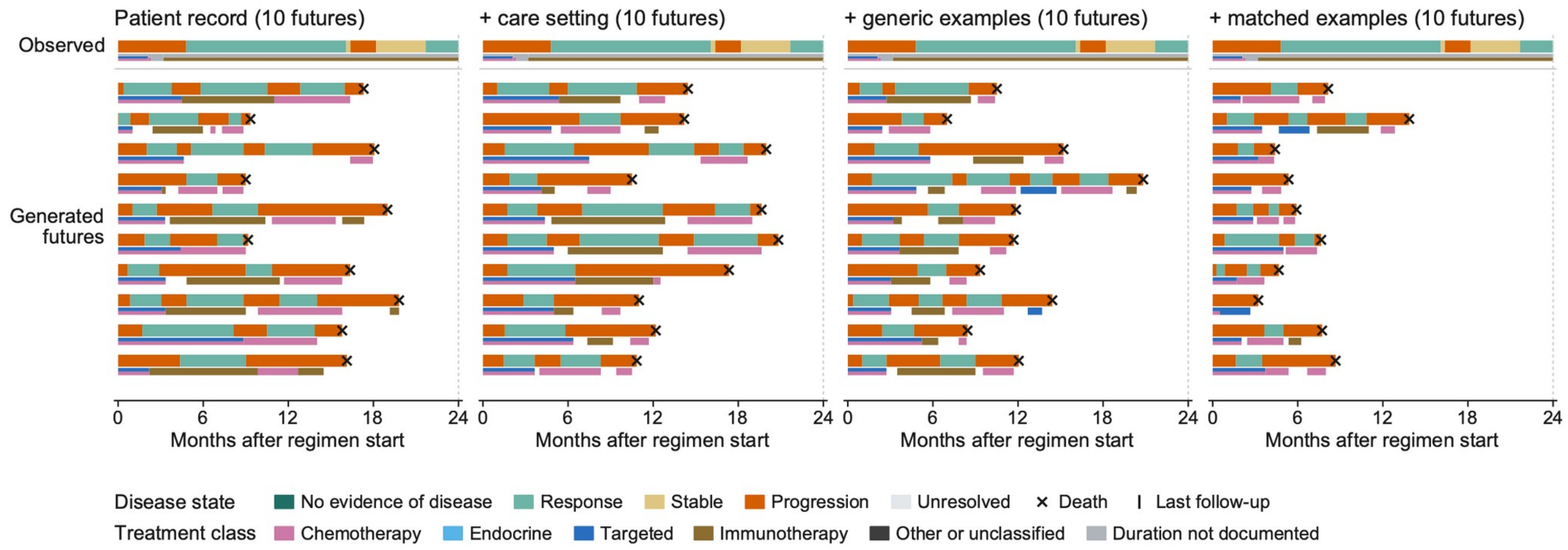

One observed patient with non-small-cell lung cancer (GENIE-BPC; third or later regimen, docetaxel + ramucirumab; selection in Methods) and the futures generated by DeepSeek V4 Flash in the four settings. Each column, one setting: observed course at the top, then every accepted future. Thick bars, disease state; thin strips, systemic treatment class (gray, duration not documented); ×, death; |, last follow-up before 24 months; dotted line, end of the observed window.

## Supplementary Figure 13 | A colorectal cancer case: one observed patient and generated futures

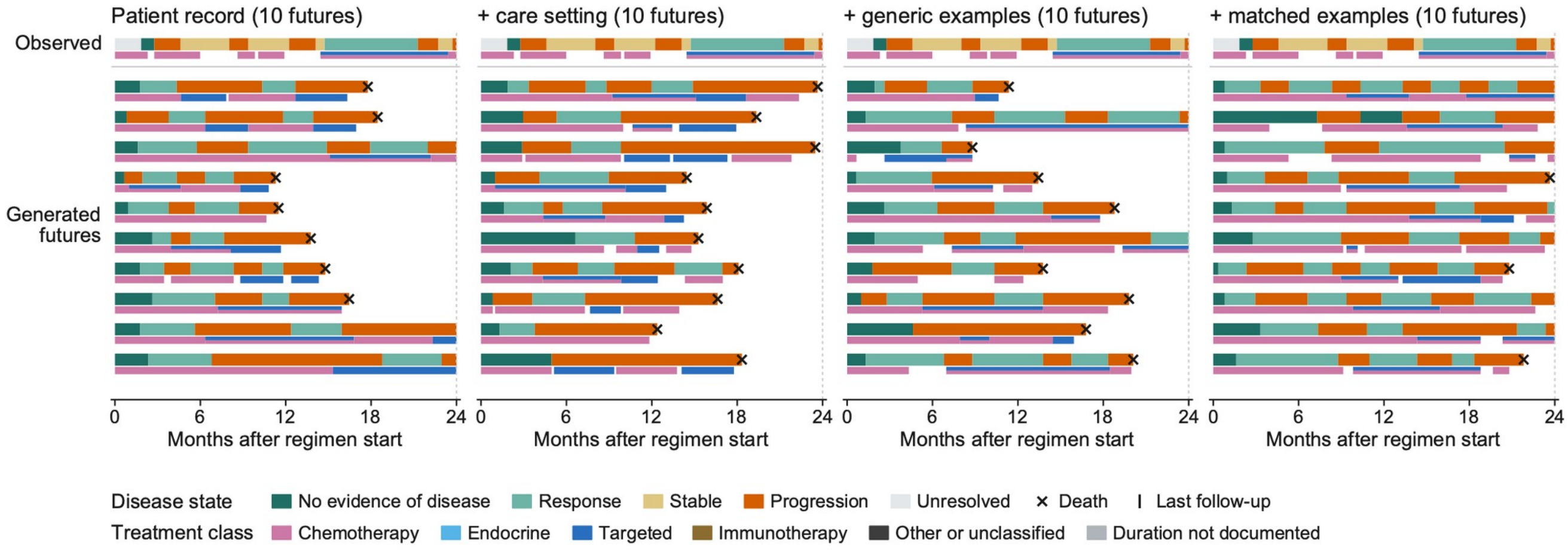

As Supplementary Fig. 12, for one observed patient with colorectal cancer (second regimen, floxuridine).

## Supplementary Figure 14 | PRIME, KRAS wild-type disease: starting positions and full course

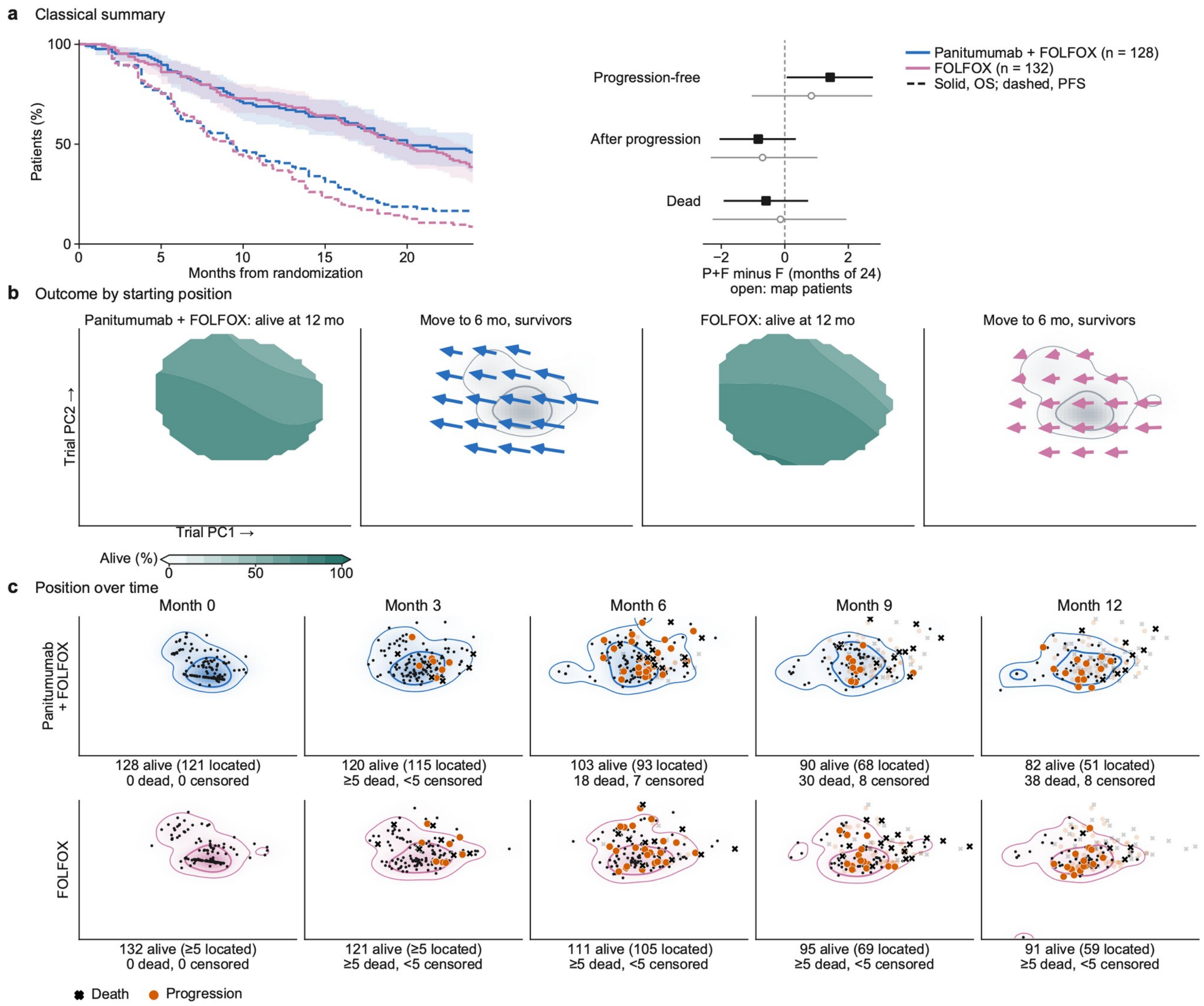

**d** Clinical course

| | Panitumumab + FOLFOX | | | | FOLFOX | | | |
|---|---|---|---|---|---|---|---|---|
| Months | 3 | 6 | 12 | 18 | 3 | 6 | 12 | 18 |
| Alive | 95 | 86 | 69 | 55 | 94 | 86 | 71 | 55 |
| Progression-free | 90 | 64 | 41 | 22 | 89 | 66 | 37 | 17 |
| CR or PR | 54 | 69 | 67 | 57 | 45 | 58 | 54 | 61 |
| Stable | 41 | 10 | | 0 | 50 | 20 | | 0 |
| Progressive | 6 | 21 | | 43 | 5 | 22 | | 39 |
| Severe AE | 6 | 6 | 0 | 0 | 0 | 0 | 0 | 0 |
| Albumin <35 g/L | 30 | 29 | 21 | | 20 | 17 | | |
| Hb <100 g/L | | | 0 | 0 | 5 | 8 | 0 | 0 |

As Extended Data Fig. 9, for KRAS wild-type patients with a lesion assessment by day 14 (128 and 132).

# Supplementary Figure 15 | The precision ladder in the generated atlas

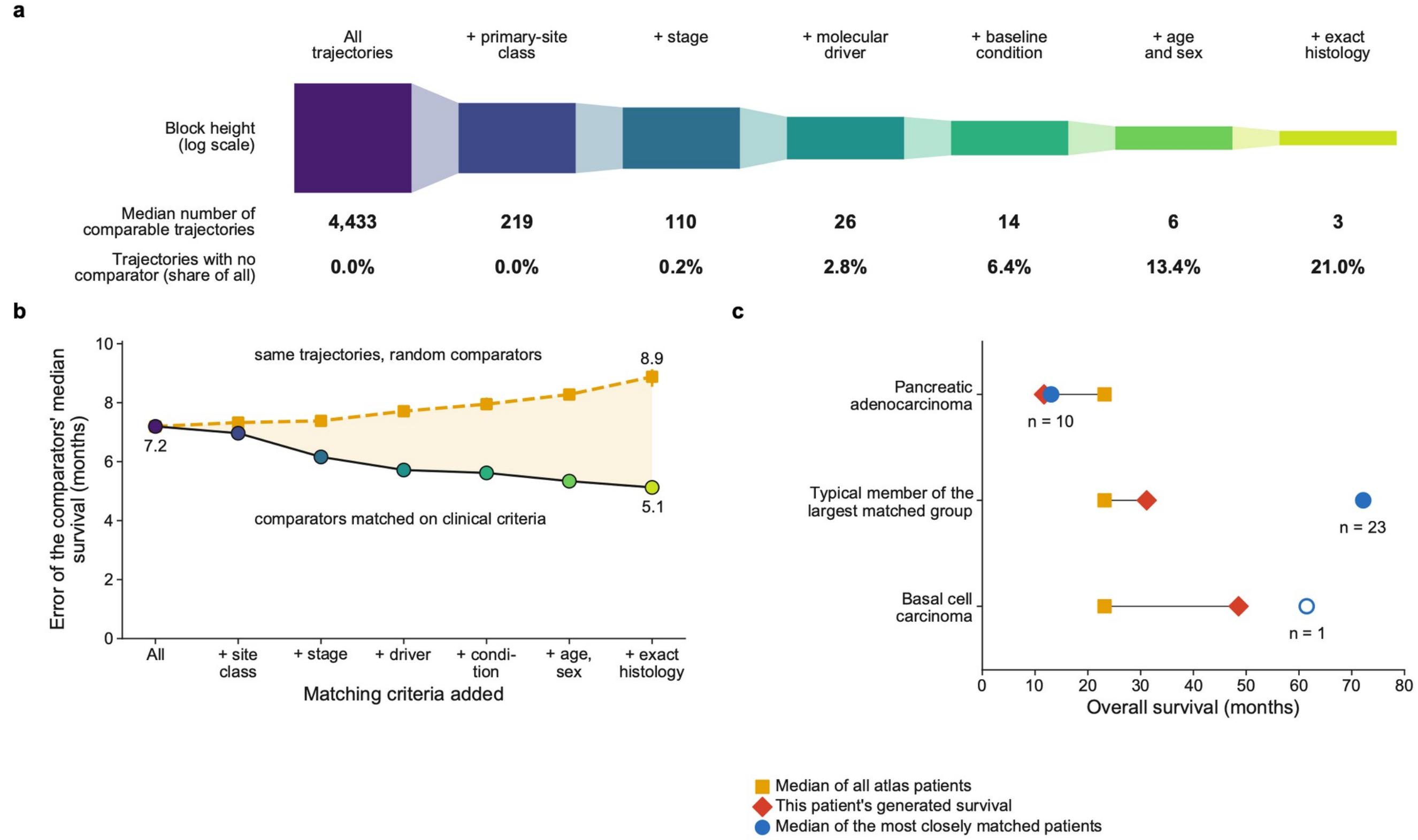

**a**, Median number of other generated trajectories matching an index trajectory as baseline criteria are added from left to right (4,434 trajectories; bottom row, share with no comparator). **b**, Median absolute error of the comparators' median survival as a forecast of the trajectory's survival, as criteria are added (3,503 trajectories with a comparator at every step; black), and with the same number of comparators drawn at random (orange dashed; median and central 95% of 10 draws). **c**, Three index trajectories (top, pancreatic adenocarcinoma; middle, a hairy cell leukemia generated by GLM-5.2, the member nearest the center of the largest group matched on all criteria; bottom, basal cell carcinoma): median survival of the atlas (orange square), the trajectory's own survival (red diamond) and the median of its matched comparators from the six models (blue circle; n, comparators; open, fewer than 3). For the hairy cell leukemia, the median generated survival ranged from 31 months (GLM-5.2) to 183 months (Kimi K3) across models.

## Supplementary Figure 16 | Gain from matched examples by agreement of the matched example patients, GENIE-BPC registry experiment

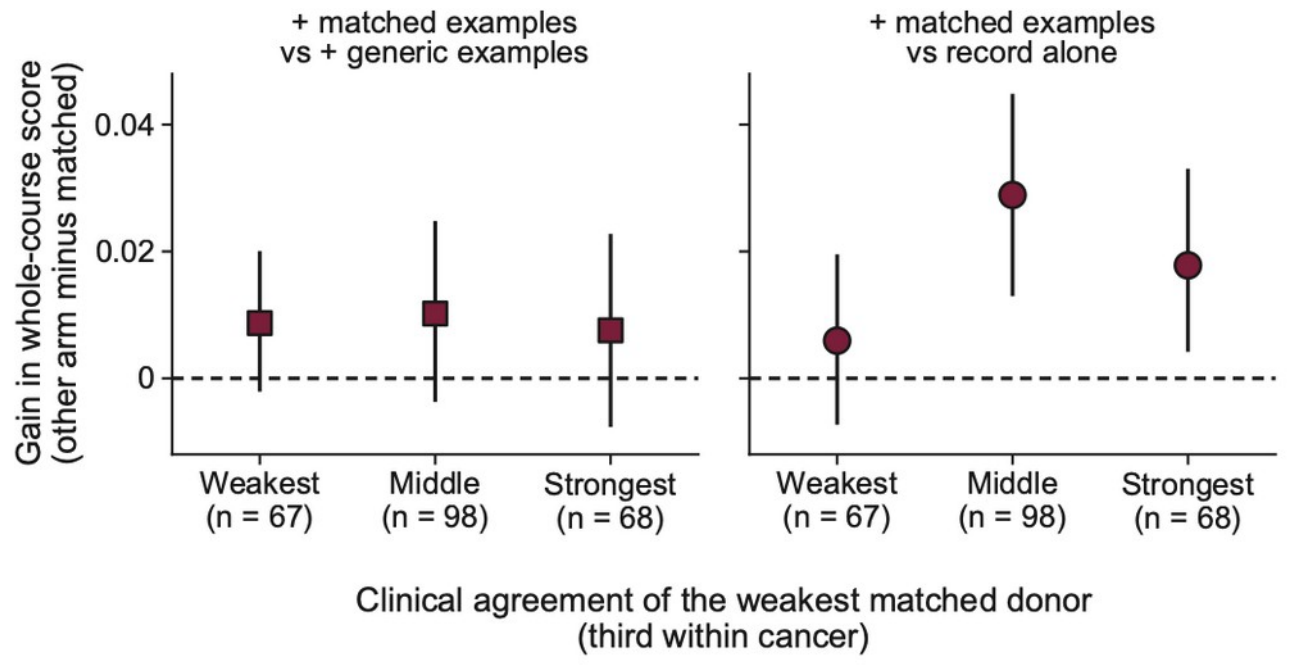

Patients evaluable at 12 months (n = 233), grouped in thirds of the clinical agreement of the weakest of their 20 matched example patients, ranked within cancer (weakest, middle and strongest thirds: 67, 98 and 68 patients; patients with the same agreement are kept in the same third). Left, gain from matched over generic examples (squares); right, gain from matched examples over the record alone (circles), in the whole-course joint score (energy score on 17 summaries, lower is better; gain = other setting minus matched, positive when matched examples are closer to the observed course). Points, means; lines, 95% intervals from 2,000 bootstrap resamples within cancer and third. We detected no association between agreement and the gain over generic examples (Spearman $\rho = 0.01$, 95% interval −0.13 to 0.14; $P = 0.91$, 2,000 permutations within cancer) or over the record alone ($\rho = 0.09$, 95% interval −0.05 to 0.20; $P = 0.18$). Every patient had at least 43 comparable candidates (same cancer, treatment modality and era), so this analysis concerns weaker clinical agreement, not the absence of comparable patients. The share of matching criteria unknown for the patient was 10%, 7% and 6% from the weakest to the strongest third, so weaker agreement partly reflects sparser documentation.

## Supplementary Table 1 | Clinical questions restated within the trajectory framework

| Clinical question | Formulation within the framework |
|---|---|
| Diagnosis and current state: what is happening to this patient? | Infer the current disease and patient state from the available observations. |
| Prognosis: what will happen under the current strategy? | Describe the possible future trajectories, with their timing and uncertainty, under a stated care strategy. |
| Prediction of treatment benefit: what would another treatment change? | Compare the futures under alternative actions: disease control, toxicity, functional decline and survival. Establishing a real effect requires causal evidence. |
| Monitoring: what does a new assessment tell us? | Update the estimated state and the distribution of subsequent trajectories. |
| Patient similarity: which patients are comparable? | Compare current condition and relevant history for the outcome, treatment and horizon considered. |
| Outcome definition: what constitutes benefit? | Evaluate the whole trajectory, including time spent well, treatment burden and adverse outcomes, rather than one endpoint. |
| Precision: how precise can the description be? | Condition on more of the state and history while recording how many comparable patients remain. |

The table states what each question is within the framework. It does not claim that any model, including the LLMs studied here, answers these questions correctly.

## Supplementary Table 2 | The 12 constructed clinical scenarios

Each scenario is a constructed patient for whom a model generated futures under several assigned treatments (3 futures per treatment and model). Treatments are compared on the median generated progression-free time of the assigned line (line 1, or line 2 where the scenario includes prior treatment). The score of a scenario is the share of its scored checks that the model passed; observed values are 0%, 50% and 100%. Comparisons with supportive care are not in the score (supportive care has no assigned line; it is a separate control on overall survival), and neither is the separate check that a wrong-target treatment does not rank first. Scores describe these scenarios, not clinical error rates (Extended Data Fig. 4a,b). Models: Gemini 3.1 Flash-Lite, Qwen 3.6 Flash, Gemma 4, Claude Sonnet 5.

| # | Scenario | Tests | Assigned treatments | Scored checks | Score (%): Gemini / Qwen / Gemma / Sonnet |
|---|---|---|---|---|---|
| 1 | EGFR-mutant lung adenocarcinoma, first line | direction | osimertinib; pembrolizumab; supportive care | osimertinib > pembrolizumab | 100 / 100 / 100 / 100 |
| 2 | Driver-negative, PD-L1-high lung adenocarcinoma, first line | direction (two acceptable standards) | pembrolizumab; chemotherapy-immunotherapy; chemotherapy; supportive care | pembrolizumab > chemotherapy; chemotherapy-immunotherapy > chemotherapy | 100 / 50 / 100 / 100 |
| 3 | EGFR wild-type lung adenocarcinoma, first line (pair of 1) | biomarker change | osimertinib (no target); chemotherapy; supportive care | chemotherapy > osimertinib | 0 / 0 / 0 / 100 |
| 4 | MSI-high metastatic colorectal cancer | direction | pembrolizumab; chemotherapy; cetuximab; supportive care | pembrolizumab > chemotherapy | 100 / 100 / 100 / 100 |
| 5 | MSS, RAS-mutant metastatic colorectal cancer (pair of 4) | biomarker change | chemotherapy + bevacizumab; pembrolizumab; cetuximab; supportive care | chemotherapy + bevacizumab > pembrolizumab; chemotherapy + bevacizumab > cetuximab | 100 / 100 / 50 / 100 |
| 6 | BRAF V600E melanoma | direction (two acceptable standards) | BRAF + MEK inhibitor; combined immunotherapy; chemotherapy; supportive care | BRAF + MEK > chemotherapy; immunotherapy > chemotherapy | 100 / 100 / 100 / 100 |
| 7 | HER2-positive breast cancer | direction | anti-HER2 therapy + chemotherapy; chemotherapy; endocrine therapy; supportive care | anti-HER2 + chemotherapy > chemotherapy | 100 / 100 / 0 / 0 |
| 8 | HER2-negative, hormone-receptor-positive breast cancer (pair of 7) | biomarker change | anti-HER2 therapy + chemotherapy (no target); endocrine therapy; chemotherapy; supportive care | endocrine therapy > anti-HER2 + chemotherapy | 100 / 100 / 100 / 100 |
| 9 | Clear-cell renal cell carcinoma | direction | immunotherapy + TKI; TKI; cytotoxic chemotherapy; supportive care | immunotherapy + TKI > cytotoxic chemotherapy | 100 / 100 / 100 / 100 |
| 10 | High-grade serous ovarian cancer at relapse, platinum-sensitive versus platinum-resistant | history | platinum; non-platinum chemotherapy; PARP inhibitor; supportive care | platinum progression-free time longer after sensitive than after resistant relapse, by more than 1.5 months; platinum not first after resistant relapse | 50 / 50 / 50 / 50 |
| 11 | Pancreatic adenocarcinoma, fit versus frail | patient condition | FOLFIRINOX; gemcitabine + nab-paclitaxel; gemcitabine; supportive care | FOLFIRINOX first in the fit patient; FOLFIRINOX not first in the frail patient | 0 / 50 / 50 / 50 |
| 12 | EGFR-mutant lung adenocarcinoma after progression on osimertinib | history | switch to chemotherapy; continue osimertinib; supportive care | chemotherapy > continue osimertinib | 0 / 0 / 0 / 100 |

Scenarios 1-9 depend on biomarker or treatment class; 10-12 also need prior treatment or the patient's condition. Mean scores over 1-9 and 10-12: Gemini 88.9% and 16.7%; Qwen 83.3% and 33.3%; Gemma 72.2% and 33.3%; Sonnet 88.9% and 66.7%.

## Supplementary Table 3 | Characteristics of the real patients analyzed

GENIE-BPC: the 450 patients of the forecasting experiment (Figs 4 and 5), by cancer; counts of 1 to 4, and counts within 4 of the total, are masked. PRIME: patients with known KRAS exon 2 status and the map subset (baseline lesion assessment by day 14; Fig. 6), by arm. Descriptive; no test.

| Cohort | n | Missing (age / sex / stage) | Age at diagnosis, median (IQR), years | Female | Stage IV at diagnosis | Year of diagnosis, range | Year of forecast treatment start, range | Regimen 1 / 2 / 3 or later |
|---|---|---|---|---|---|---|---|---|
| GENIE-BPC NSCLC | 150 | 0 / 0 / 0 | 64 (54-71) | 73 (49%) | 77 (51%) | 2006-2017 | 2012-2019 | 50 (33%) / 50 (33%) / 50 (33%) |
| GENIE-BPC BrCa | 150 | 0 / <5 / 0 | 46 (39-50) | >145 | 40 (27%) | 1999-2018 | 2006-2020 | 50 (33%) / 50 (33%) / 50 (33%) |
| GENIE-BPC CRC | 150 | 0 / 0 / 0 | 52 (45-63) | 64 (43%) | 83 (55%) | 2006-2018 | 2011-2019 | 50 (33%) / 50 (33%) / 50 (33%) |

| Cohort | n | Missing (age / sex / ECOG / liver) | Age, median (IQR), years | Female | KRAS exon 2 wild-type | ECOG 0, of known | Liver metastases |
|---|---|---|---|---|---|---|---|
| PRIME, known KRAS exon 2 status: Panitumumab + FOLFOX | 436 | 0 / 0 / 0 / 0 | 62 (55-70) | 154 (35%) | 260 (60%) | 241 (55%) | 388 (89%) |
| PRIME, known KRAS exon 2 status: FOLFOX | 430 | 0 / 0 / 1 / 0 | 62 (54-68) | 174 (40%) | 254 (59%) | 240 (56%) | 382 (89%) |
| PRIME, map patients: Panitumumab + FOLFOX | 218 | 0 / 0 / 0 / 0 | 62 (55-69) | 75 (34%) | 128 (59%) | 116 (53%) | 200 (92%) |
| PRIME, map patients: FOLFOX | 221 | 0 / 0 / 1 / 0 | 61 (52-68) | 91 (41%) | 132 (60%) | 116 (53%) | 209 (95%) |

## Supplementary Table 4 | Generating models and settings

**a, Cancer atlas (six models; 148 cancer types; target 5 accepted trajectories per cancer type and model).** All calls: temperature 1.0; reasoning off; output cap 16,384 tokens; no seed sent; JSON-object response format; at most 3 attempts per requested trajectory and at most 24 requested trajectories per cell; prompt prompts/oncollm_v1/experimental_lean_ledger_v2_rich_neutralized.md (SHA-256 3109486f...0bac6), enumeration order shuffled with one fixed seed for every model. Costs are the per-attempt costs reported by the provider and stored in the run directory; they include attempts reused when a run was resumed.

| Model (paper name) | Model identifier returned | Run | Dates (UTC) | Accepted | Rejected after 3 attempts | Provider attempts | Cost (USD) |
|---|---|---|---|---|---|---|---|
| Claude Sonnet 5 | anthropic/claude-sonnet-5-20260630 | sonnet5_mini_atlas_n100_2026_07_08 + sonnet5_full_atlas_topup_remaining128_n5_2026_07_09 | 8-9 Jul 2026 | 740 | 93 | 1,296 | 246.72 |
| Gemini 3.1 Flash-Lite | google/gemini-3.1-flash-lite-20260507 | pillar1_anchored_atlas_tranche1_v4_n5_live01 | 3 Jun 2026 | 740 | 36 | 1,084 | 3.31 |
| Gemma 4 | google/gemma-4-31b-it-20260402 | pillar1_anchored_atlas_gemma4_n5_live01 | 6 Jun 2026 | 740 | 57 | 1,040 | 2.62 |
| Kimi K3 | moonshotai/kimi-k3 | kimi_k3_off_full_atlas_n5_complete | 27 Jul 2026 | 737 | 39 | 980 | 77.32 |
| GLM-5.2 | GLM-5.2-FP8 | glm52_full_atlas_n5 | 25 Jul 2026 | 737 | 43 | 1,098 | not recorded |
| Qwen 3.6 Flash | qwen/qwen3.6-flash | pillar1_anchored_atlas_tranche1_v4_n5_live01 | 3 Jun 2026 | 740 | 16 | 1,079 | 18.16 |
| Total | | | | 4,434 | 284 | 6,577 | 348.13 (five models) |

Kimi K3 and GLM-5.2 each have one cell that reached the 24-request cap with 2 accepted trajectories (THFO and ESCA, respectively); every other cell reached the target of 5. One further Sonnet slot was interrupted before completion (one saved attempt) and is counted neither as accepted nor as rejected. Sonnet, Gemini, Gemma, Kimi and Qwen were called through OpenRouter. GLM-5.2 (FP8 weights) was called through an institutional OpenAI-compatible endpoint with temperature 1.0 and top-p 1.0 sent explicitly; this endpoint reports no cost. A few attempts (2 Sonnet, 9 Gemma, 17 Kimi) failed before a provider response and carry no usage record. Reasoning was requested off in every call. The provider nevertheless reported reasoning tokens for Qwen 3.6 Flash in every attempt with a usage record, in the atlas (1,079 of 1,079 attempts; median 9,305 tokens) and in the scenario benchmark of b; the provider also reported reasoning tokens for one Gemma 4 attempt of the scenario benchmark (field usage.reasoning_tokens of each result.json).

**b, Controlled-input experiments.** Same prompt template and settings as in a, except the number of attempts: at most 4 per requested future (trajectory_max_attempts 4 in every run manifest). Each design cell had a fixed number of requested futures (3); rejected futures were not replaced, except for Kimi K3 (see below).

| Experiment | Model | Run(s) | Date (UTC) | Accepted / requested | Provider attempts | Cost (USD) |
|---|---|---|---|---|---|---|
| Factorial experiment and EGFR assay (2 EGFR x 5 states x 3 histories x 4 treatments x 3 futures) | Claude Sonnet 5 (anthropic/claude-sonnet-5) | interventional_phase1_sonnet_full_v1c | 23 Jul 2026 | 360 / 360 | 397 | 57.16 |
| EGFR checks | Kimi K3 | interventional_phase1_kimi_k3_off_full_v1c + interventional_phase1_kimi_k3_off_topup75_v1c_retry2 | 28 Jul 2026 | 360 / 360 | 679 | 29.14 |
| EGFR checks | GLM-5.2 (FP8) | interventional_phase1_glm52_full_v1c | 26 Jul 2026 | 341 / 360 | 587 | not recorded |
| 12 constructed clinical scenarios (Supplementary Table 2; 3 futures per arm and model) | Gemini 3.1 Flash-Lite, Qwen 3.6 Flash, Gemma 4, Claude Sonnet 5 | sentinel_usecase_*_v5_2026_07_13 (nine runs; for the post-osimertinib scenario, the explicit-resistance run ..._v5_v2) | 13 Jul 2026 | 159 per model (636) | 719 | 31.47 |

EGFR checks, accepted futures per arm and EGFR status (45 requested each): Sonnet and Kimi K3, 45 in every arm; GLM-5.2, 45 except EGFR-mutant chemotherapy 35, EGFR-mutant pembrolizumab 39 and EGFR-wild-type chemotherapy 42 (19 futures rejected after 4 attempts, for failed observed-prefix audit, event validation or replay; not replaced). Kimi K3: in the main run, 75 requested futures failed with transport or runtime errors, without a usable model response; these 75 slots were requested again in a top-up run (..._topup75_v1c_retry2), and all 75 were accepted. Attempt and cost totals cover the analyzed runs.

**c, Registry (GENIE-BPC) four-arm experiment.**

| Item | Value |
|---|---|
| Model | DeepSeek V4 Flash, identifier deepseek/deepseek-v4-flash-0731 (every attempt returned this identifier) |
| Run | runs/four_arm_v1_definitive_p40b20_20260826 |
| Date (UTC) | 26 Aug 2026 |
| Providers | OpenRouter, restricted to Novita, SiliconFlow and DeepInfra; data collection denied; zero data retention requested. Attempts served: Novita 15,553, SiliconFlow 6,672, DeepInfra 898 |
| Sampling | temperature 1.0; top-p not sent; reasoning off; output cap 24,576 tokens |
| Design | 450 patients x 4 arms x 10 requested futures = 18,000 requested futures (4,500 per arm) |
| Attempts | at most 3 per requested future; rejected slots not refilled; 23,123 provider attempts in total |
| Accepted futures | 17,679 of 18,000: patient record only 4,416; + care setting 4,407; + generic examples 4,426; + matched examples 4,430 |
| Prompt | continuation template prompts/anchored_generation_v1/lean_ledger_v2_rich_neutralized_continuation.md (SHA-256 e150ea14...a32c5), derived from the atlas template; setting sentence "Care is provided in a North American academic oncology practice in {year}, where clinical trials may be available."; knowledge cutoff stated in the prompt, 31 Dec 2024 |
| Cost | 103.28 USD (sum of provider-reported cost over all attempts) |

## Supplementary Table 5 | Generation contract: output schema and prompts

**a, Output schema.** Every generated record is one JSON object: a baseline at diagnosis and a chronological event ledger. Deterministic replay applies the events in order to reconstruct the patient's state at every time. Both parts are validated against Pydantic models; the JSON schemas of these models are injected verbatim into every prompt.

| Part | Content | Source |
|---|---|---|
| Baseline (BaselineObject) | identity card (demographics: age, sex, ECOG at diagnosis; diagnosis: date, OncoTree code, histology and primary-site classes; staging; molecular profile: alterations and biomarkers; baseline health: comorbidities, organ function, contextual factors); clone registry; disease burden by site at baseline; baseline patient state; medical history | src/synthonccases/oncollm/ replay.py, classes BaselineObject, IdentityCard and their members |
| Event ledger (LedgerEvent) | one entry per change, with date and type. Eleven event types used in this study: systemic treatment start, switch, stop and modification; response; progression; clonal evolution; toxicity (onset, grade change or resolution, CTCAE v5 grade 1-5, adverse-event category); local intervention; patient-state update (ECOG, weight, symptom burden, organ function); death (exactly one, terminal). A twelfth type, last follow-up, exists in the engine and was not allowed in prompts | replay.py, EventType and LedgerEvent |
| Acceptance rules | parseable JSON with only allowed keys; every event valid against the schema; dates in order and, for continuation, strictly after the cutoff; a regimen for every treatment event; exactly one terminal death; no truncated output. A record failing any rule is rejected, not repaired | src/synthonccases/oncollm/ poc_runner.py; rejection reasons in runs/four_arm_v1_definitive_p40b20_20260826/summary.json |

**b, Prompts** used for the runs of Supplementary Table 4.

| Prompt | Used for | File (SHA-256) | Summary |
|---|---|---|---|
| Atlas template ("neutralized framing control") | cancer atlas (six models), factorial experiment, EGFR checks, 12 constructed scenarios | prompts/oncollm_v1/ experimental_lean_ledger_v2_rich_neutralized.md (3109486f7e2f96f94c61685bd000a15fdef52e6759b13b11517af15407d0bac6) | Asks for one JSON object, {baseline_patch, ledger}, describing the patient's physical course from diagnosis to death: disease changes, clonal evolution, host-state changes, treatments, toxicities and death. Gives the output contract, a completeness rule (every clinically meaningful transition, no routine stable visits), event-by-event rules, the injected event and baseline schemas, the baseline seed with pinned fields (for the atlas, only the OncoTree code, its name and a diagnosis date), the fields the model must fill, and run-specific instructions (none for the atlas; the specified state, history and assigned treatment for controlled inputs). Knowledge is limited to a stated cutoff. Schema-control only; no clinical coaching. Allowed values are listed in a shuffled order fixed by one seed |
| Continuation template | registry four-arm experiment | prompts/ anchored_generation_v1/ lean_ledger_v2_rich_neutralized_continuation.md (e150ea142d4219b3e9cb5d5e414521081c35882e030d62180805a339800a32c5) | Same template, changed only where continuation requires it: output is {ledger} only; the patient's recorded baseline and history up to the landmark are given as a fixed seed and fixed prefix; fields not supplied are stated to be unobserved, not normal; every event must fall strictly after the landmark (start of the index regimen); one terminal death. Knowledge cutoff 31 Dec 2024 |
| Arm-specific block (four-arm experiment) | arms 2-4 | scripts/explore/ emit_four_arm_prompts_v1.py (SETTING, card renderer); provenance outputs/exploratory_four_arm_prompts/provenance.json | Arm 1, nothing added. Arm 2, one setting sentence: "Care is provided in a North American academic oncology practice in {year}, where clinical trials may be available." Arm 3, the setting sentence and 20 generic same-cancer, same-era example patients. Arm 4, the setting sentence and 20 matched example patients. Arms 3 and 4 use the same renderer, without arm labels or similarity scores. The patient record and index regimen are byte-identical across the four arms (asserted at emission; 450 quartets, no problems recorded) |

## Supplementary Table 6 | Cohorts, exclusions and analysis populations

Counts only. No patient identifiers are listed.

**a, Generated cancer atlas (no patients).**

| Population | n |
|---|---|
| Cancer types (OncoTree codes) requested | 148 |
| Accepted trajectories, six models (target 5 per cancer type and model) | 4,434 (740 per model; 737 for Kimi K3 and GLM-5.2) |
| Cancer types with all five trajectories in every model (between- and within-cancer variation; all 148 codes are present in every model) | 146 |
| Trajectories in the six-month displacement field | 4,432 |
| Trajectories with a comparator at every matching step | 3,503 |

**b, GENIE-BPC four-arm experiment (DeepSeek V4 Flash).**

| Population | n |
|---|---|
| Patients (queries), one landmark per patient (start of a named systemic regimen), stratified by regimen depth | 450 |
| By cancer: non-small-cell lung, breast, colorectal | 150 each |
| Patients drawn at random before selection and excluded from the patients and example patients of this experiment | 100 per cancer |
| Requested futures (450 patients x 4 arms x 10) | 18,000 |
| Accepted futures after at most 3 attempts | 17,679 (patient record only 4,416; + care setting 4,407; + generic examples 4,426; + matched examples 4,430) |
| Rejected requested futures (slot not refilled) | 321 |
| Observed deaths within 6 months | 73 of 450 |

Horizon-specific analysis populations, primary convention (last documented state carried forward) unless stated. Each analysis and each horizon has its own population, so trends across horizons are not inferred from the same patients. "Composition" means outputs/four_arm_preprint_20260928/population_composition_scores_20261012.csv, with the value of its population column; "Precheck" means outputs/four_arm_preprint_20260928/eligibility_precheck_20261012.csv. Rules are those of Manuscript, Methods (Forecasting registry patients) and the Amendment.

| Population | Eligibility rule | 3 mo | 6 mo | 12 mo | 24 mo | Used in |
|---|---|---|---|---|---|---|
| Survival, all seven forecasters available | each arm has at least seven accepted futures that all resolve survival at H; both donor cards give a Kaplan-Meier estimate at H; cross-validated Cox prediction available | 449 | 449 | 449 | 447 | Fig. 4d; Extended Data Fig. 5a,b,e |
| Treatment summaries | documented months on systemic therapy and number of subsequent regimens resolved in the real patient at H; at least seven accepted futures per arm, all resolving both; at least seven members of each donor card with resolved values | 386 | 365 | 317 | 244 | Fig. 5b (inset) and Extended Data Fig. 7c (12 months) |
| Real target resolves all 17 summaries | observed course documented to H for every summary | 288 | 278 | 260 | 216 | eligibility step of the joint score |
| Joint score (four-arm intersection) | real target and every accepted future resolve all 17 summaries; at least seven accepted futures in each of the four arms | 286 | 269 | 233 | 181 | Fig. 5d; Supplementary Fig. 11b,c |
| Joint score, first responses only | joint-score patients whose first attempts alone give at least seven accepted futures per arm, all resolving the 17 summaries | 145 | 138 | 121 | 91 | Extended Data Fig. 7d |
| Joint score, eligible under both conventions | joint-score rule met under the primary and the four-month expiry reading | 248 | 194 | 130 | 90 | Extended Data Fig. 7d |
| Documented-time disagreement (course diagnostic) | at least one month of evaluable query time within H; at least seven accepted futures per arm, all projectable at H; generated time without a resolved state is charged as disagreement | 386 | 412 | 426 | 437 | Extended Data Fig. 7a,b |
| Donor diagnostic, generic card | at least seven usable donors (shared documented window of at least one month); generic-examples arm evaluable as above | 386 | 412 | 426 | 437 | Extended Data Fig. 5d |
| Donor diagnostic, matched card | at least seven usable donors; matched-examples arm evaluable as above | 381 | 411 | 426 | 437 | Extended Data Fig. 5d |

Deaths by 24 months make up 51.9% of the primary 24-month joint-score set (94 of 181 patients), 80.0% of the both-convention set (72 of 90) and 42.2% of the full cohort (190 of 450) (Amendment, section 9; Composition, deaths_by_h). These sets are selected populations; the eligibility rule required the outcome to be evaluable under the minimum follow-up or death rule (Manuscript, Methods, Forecasting registry patients). At 24 months, 42 to 43 patients of the survival, course and donor populations remain unresolved (alive, followed less than 24 months); the joint-score sets contain none.

**c, PRIME trial (observed data; no generated forecasts analysed).**

| Population | n |
|---|---|
| Randomised patients, progression-free survival analysis (panitumumab + FOLFOX versus FOLFOX alone) | 935 (469 versus 466) |
| KRAS exon 2 test failed or missing (all-patient row only) | 69 |
| Baseline lesion assessment by day 14 and known KRAS exon 2 status (map population) | 439 |
| Map population by arm: KRAS wild-type, panitumumab + FOLFOX / FOLFOX alone | 128 / 132 |
| Map population by arm: KRAS mutant, panitumumab + FOLFOX / FOLFOX alone | 90 / 89 |

Per-subgroup patient numbers for the restricted mean progression-free survival differences are given in Extended Data Fig. 8a.

## Supplementary Table 7 | TRIPOD-LLM checklist

Reporting checklist for studies using large language models (TRIPOD-LLM; Gallifant et al., Nature Medicine 2025), for a study that evaluates existing LLMs (design E, evaluation; H, evaluation in healthcare settings; task OF, outcome forecasting). Items are abbreviated from the published checklist. NA, not applicable.

| Section | Item | Checklist item (abbreviated) | Where reported |
|---|---|---|---|
| Title | 1 | Study as development, tuning or evaluation of an LLM; task, population, outcome | Title; Abstract |
| Abstract | 2 | TRIPOD-LLM for Abstracts | Abstract |
| Background | 3a | Healthcare context and rationale, with existing approaches | Introduction |
| Background | 3b | Target population and intended use in the care pathway | Introduction; Discussion |
| Objectives | 4 | Objectives: development, tuning or validation | Introduction, last paragraph |
| Data | 5a | Sources of the evaluation data and rationale | Methods, Data sources and ethics; Forecasting registry patients; PRIME trial |
| Data | 5b | Description of the data points and their distribution | Methods; Supplementary Tables 6 and 3 |
| Data | 5c | Dates of the oldest and newest items in the evaluation data | GENIE-BPC public releases (Methods, Forecasting registry patients) and PRIME (Methods, PRIME trial); prompts carry no calendar date. Years of diagnosis and of the forecast treatment start, Supplementary Table 3 |
| Data | 5d | Pre-processing and quality checking | Methods, Generation of patient trajectories; Forecasting registry patients |
| Data | 5e | Missing and imbalanced data; reasons for omitting data | Methods, Forecasting registry patients; Supplementary Table 6 |
| Analytical methods | 6a | LLM name, version and last date of training | Methods, Generation of patient trajectories; Supplementary Table 4. Training cut-off dates are not reported by the developers |
| Analytical methods | 6b | Development process (architecture, training, alignment) | NA (existing models, not developed here) |
| Analytical methods | 6c | How text was generated: prompts, consistency, inference settings | Methods, Generation of patient trajectories; Supplementary Tables 4 and 5 |
| Analytical methods | 6d | Initial and post-processed output | Methods, Generation of patient trajectories |
| Analytical methods | 6e | How probabilities were determined and thresholds identified | Methods, Forecasting registry patients: the predicted probability of being alive at a horizon is the share of a patient's generated futures alive at that horizon; no classification threshold was used |
| LLM output | 7a | Metrics of the quality of generative outputs against gold standards | Methods, Forecasting registry patients |
| LLM output | 7b | Relevance of the metrics to the downstream task | Methods, Forecasting registry patients; Discussion |
| LLM output | 7c | Outcome definition, how predictions were calculated, inference dates | Methods, Forecasting registry patients; Generation of patient trajectories (call dates) |
| LLM output | 7d | Assessors for subjective outcome assessment | No subjective assessment of outputs: the expected orderings of the scenarios were defined by the authors (Supplementary Table 2), and scoring was automated |
| LLM output | 7e | Comparison with other LLMs, humans or benchmarks | Results; Methods, Forecasting registry patients (Cox model and example patients as comparators) |
| Annotation | 8a-c | Annotation guidelines, annotators, agreement | NA (no annotation) |
| Prompting | 9a | Prompt design, curation and selection | Methods, Generation of patient trajectories; Supplementary Table 5 |
| Prompting | 9b | Data used to develop the prompts | Methods, Generation of patient trajectories; Supplementary Table 5 |
| Summarization | 10 | Pre-processing before summarization | NA |
| Instruction tuning | 11 | Instruction tuning or alignment | NA |
| Compute | 12 | Compute or proxies (cost, time, machines) | Supplementary Table 4: requested and accepted outputs, provider attempts and cost per run |
| Ethical approval | 13 | Ethics committee and consent or waiver | Methods, Data sources and ethics: secondary analysis of existing, de-identified data; no patients recruited or contacted |
| Open science | 14a | Funding and role of funders | Acknowledgements |
| Open science | 14b | Conflicts of interest | Competing interests |
| Open science | 14c | Protocol | No prospective protocol; all analyses are exploratory (Methods, Statistical analysis) |
| Open science | 14d | Registration | Not registered |
| Open science | 14e | Availability of data | Data availability |
| Open science | 14f | Availability of code | Code availability |
| Public involvement | 15 | Patient and public involvement | None |
| Participants | 16a | Flow of patient data, with outcomes and follow-up | Methods, Forecasting registry patients; Supplementary Table 6 |
| Participants | 16b | Characteristics overall and by data source | Supplementary Table 3 |
| Participants | 16c | Clinical variables in development versus evaluation data | NA (no development data) |
| Participants | 16d | Participants and outcome events in each analysis | Figure legends; Supplementary Table 6 |
| Performance | 17 | Performance on prespecified metrics | Results; Figs 4 and 5; Extended Data Figs 5 and 7 |
| LLM updating | 18 | Results of LLM updating | NA |
| Interpretation | 19a | Overall interpretation, including fairness | Discussion |
| Limitations | 19b | Limitations, bias, uncertainty, generalizability | Discussion |
| Usability | 19c | Challenges of the data: representation, | Discussion; Methods, Forecasting registry patients |
| | | missingness, harmonization, bias | |
| Usability | 19d | Intended use, input, end-user, human oversight | Discussion (no clinical use without prospective validation) |
| Usability | 19e | Handling of poor-quality or unavailable input | Methods, Generation of patient trajectories |
| Usability | 19f | User interaction and expertise | NA (no deployment) |
| Usability | 19g | Next steps | Discussion |

## Supplementary Note 1 | Supporting analyses

**Survival in metastatic disease at diagnosis.** In the eight cancer families of Fig. 1d, patients with metastatic disease at diagnosis made up 43.9% to 70.4% of each model's patients. Among these 951 of 1,674 patients, the median across families of the difference between the longest and shortest median survival of the six models was 19.1 months, against 19.5 months for all stages. Kimi K3 still generated the longest median survival in every family. Three model-family groups had fewer than five metastatic patients (Gemma 4: breast cancer, 2; head and neck cancer, 3; sarcoma, 4), and excluding them did not change these values.

**Sensitivity of the scenario benchmark.** The scores of Extended Data Fig. 4a,b were unchanged for platinum-sensitivity thresholds from -0.89 to 2.04 months. With overall survival instead of progression-free survival, two additional models satisfied the comparison between platinum-sensitive and platinum-resistant relapse.

**EGFR ordering checks.** Claude Sonnet 5 and Kimi K3 passed all five ordering checks, and GLM-5.2 failed the ranking of treatments in wild-type disease. Kimi K3 passed although it kept osimertinib ahead of chemotherapy in EGFR-wild-type disease (Fig. 3f), because no check required the reverse ordering.

**EGFR wild-type futures before first-line progression.** None of the 135 osimertinib futures generated in EGFR wild-type disease by the three models acquired a sensitizing EGFR alteration before first-line progression, and one GLM-5.2 future had another treatment added. After excluding, in both arms, futures with a sensitizing EGFR alteration or an added treatment before progression, the osimertinib-minus-chemotherapy difference in median time before first-line progression was -7.1 months for Claude Sonnet 5, +3.8 for Kimi K3 and +22.1 for GLM-5.2, against −7.1, +3.8 and +22.3 without exclusions (Fig. 3f). These exclusions removed one osimertinib and one chemotherapy future of GLM-5.2 (44 and 41 futures retained) and no future of the other two models. GLM-5.2 generated an EGFR resistance mutation (C797S or T790M) before progression in 17 of its 45 wild-type osimertinib futures, and Kimi K3 started chemotherapy instead of the assigned osimertinib in two wild-type futures.

**Survival from generic and matched example patients used directly.** At 12 months, the Brier score of the survival estimated directly from the matched example patients was lower than that from the generic example patients (paired difference -0.039, 95% interval -0.060 to -0.019; 449 patients; Fig. 4d). The interval does not include uncertainty in the selection of the example patients, which were fixed.

**Correction for the finite number of futures.** At 12 months, after correcting for the finite number of futures per patient, the Brier scores of the LLM forecasts were 0.212 with the record alone, 0.206 with the care setting, 0.183 with generic examples and 0.183 with matched examples (449 patients), against 0.162 for the matched example patients used directly and 0.160 for the Cox model, which need no correction.